\documentclass{article}

\usepackage[T1]{fontenc}
\usepackage[letterpaper,left=1.5in,right=1.5in,top=1in,bottom=1in]{geometry}
\usepackage{natbib}

\usepackage{amsmath}
\usepackage{amsfonts}
\usepackage{nicefrac}
\usepackage{xcolor}
\usepackage{graphicx}
\usepackage{booktabs}
\usepackage{multirow}
\usepackage{subcaption}
\usepackage{tikz}
\usetikzlibrary{arrows.meta, positioning, shadows, calc}
\usepackage{url}
\usepackage{hyperref}

\title{InCLAD: A Continual Learning Benchmark for Industrial Visual Anomaly Detection}

\author{
  {\L}ukasz Marcjan \\
  AGH University of Krakow \\
  Krakow, Poland \\
  \texttt{lmarcjan@agh.edu.pl} \\
  \and
  Dorota Wilk-Ko{\l}odziejczyk \\
  AGH University of Krakow \\
  Krakow, Poland \\
  \texttt{dwilk@agh.edu.pl} \\
  \and
  Roberto Corizzo \\
  American University \\
  Washington, DC, USA \\
  \texttt{rcorizzo@american.edu}
}
\date{}

\begin{document}

\maketitle

\begin{abstract}

Industrial anomaly detection benchmarks have advanced rapidly, yet they remain largely static and therefore underrepresent deployment settings in which inspection systems must be extended from one industrial component to another over time. In our setting, continual industrial anomaly detection is framed as a task-incremental problem: each component defines a task, while the learning objective remains semi-supervised. The goal extends beyond conventional anomaly detection, since models should preserve their performance on all components as they are sequentially exposed to new components. This framing is especially relevant because many existing industrial pipelines either assume fully supervised defect labels or study only isolated datasets, leaving open the question of how semi-supervised anomaly detectors behave across a sequence of heterogeneous components. We address this gap by organizing the benchmark as a sequence of related component-level tasks and by defining multiple scenarios, including random, easy-to-hard, and hard-to-easy. In addition to single-dataset continual scenarios, we propose challenging scenarios spanning across multiple datasets. We further evaluate continual performance from two complementary perspectives: image-level anomaly detection and pixel-level anomaly localization. By recasting industrial anomaly detection as a continual benchmark rather than a static train/test problem, our work establishes a concrete path toward studying knowledge retention, adaptation, and generalization in evolving industrial environments.
\end{abstract}

{

\section{Introduction}

Industrial anomaly detection has achieved strong results on standard benchmarks,
yet most published settings remain static: one model is trained offline on a
fixed distribution and then evaluated on a held-out split drawn from the same
or closely related conditions. Real deployment is significantly more challenging. New
components are introduced over time, and an inspection system that works well
for one component often has to be extended to additional ones without
retraining from scratch on the full historical data. A practical anomaly
detector must therefore do more than generalize once; it must continually
adapt while preserving previously acquired knowledge.

This requirement exposes a gap between current industrial anomaly detection and
continual learning. In static anomaly detection, the dominant objective is to
separate normal and anomalous patterns for one dataset at one time. In a
continual setting considered here, the objective changes: the model must absorb
new component-specific normality patterns without degrading its performance on
previously learned components. The central drawback is catastrophic
forgetting, which is especially costly in industrial inspection because a
system that forgets how to model earlier components can miss anomalies when
those products reappear in production.

The need for continual anomaly detection is reinforced by the structure of
industrial data. Indeed, in real-world settings, defective samples may be scarce,
incomplete, or too costly to annotate comprehensively.
As a result, many real-world anomaly-detection problems are inherently
semi-supervised, in the sense that they do not assume availability of anomalies as training data, they train models only on normal samples, and detect deviations at test time.  
Under this normal-only setting, the continual problem is not to incrementally learn defect classes, but to adapt the notion of normality across a sequence of
components. 
Methods that work well in a single static benchmark may still be brittle when they must preserve normal representations for earlier components while learning new ones.

Motivated by this gap, we frame continual industrial anomaly detection as a
benchmark problem in which each task corresponds to one component and the
inspection objective remains fixed across the full sequence. 
The goal is then not just merely adaptation to multiple labeled defect categories. Instead, the model has to  maintain strong anomaly detection on all components as it is sequentially exposed to new component distributions under a normal-only training regime. 
This perspective also makes it possible to study benchmark configurations that
have received little attention in continual anomaly detection: curriculum-style
orderings that explicitly control task difficulty, and multi-dataset streams
that combine component-level and dataset-level distribution shifts. In
particular, we consider random orderings, easy-to-hard and hard-to-easy
curricula, and multi-dataset scenarios. In
addition, we evaluate the continual setting from two complementary viewpoints:
image-level anomaly detection and pixel-level anomaly localization.

In summary, the main contributions of this paper are as follows:

\noindent - We formulate continual industrial anomaly detection as a benchmark problem
motivated by realistic production and inspection workflows, where each
component defines a task challenging model adaptation and knowledge retention.

\noindent - We focus on the semi-supervised setting, where models are exclusively trained with normal data, and the challenge is continual adaptation across components rather than incremental learning of multiple supervised defect classes.

\noindent - We introduce principled benchmark scenarios that go beyond prior
continual anomaly-detection studies by explicitly evaluating curriculum-based
orderings (easy-to-hard and hard-to-easy) and multi-dataset streams, enabling a
cohesive analysis across diverse datasets and task sequences. 

\noindent - We devise non-trivial continual adaptations of PatchCore, PaDiM, and CFA and perform an extensive evaluation that considers state-of-the-art continual industrial anomaly detection models across two perspectives (image-level, pixel-level), providing a multi-faceted analysis.
}

\begin{figure}[t]
\centering
\resizebox{\linewidth}{!}{%
\begin{tikzpicture}[font=\sffamily, >=Latex, line cap=round, line join=round]
\tikzset{
  card/.style={rounded corners=7pt, draw=#1!65!black, very thick, fill=#1!4, minimum width=4.35cm, minimum height=4.35cm, drop shadow={shadow xshift=0.8pt, shadow yshift=-0.8pt, opacity=0.18}},
  title/.style={font=\bfseries\large, text=black!88},
  sub/.style={font=\scriptsize, text=black!62},
  badge/.style={circle, draw=#1!75!black, fill=#1!18, very thick, inner sep=1.5pt, minimum size=0.58cm, font=\bfseries\scriptsize, text=#1!45!black},
  chip/.style={rounded corners=3pt, draw=#1!45, fill=#1!12, inner xsep=3pt, inner ysep=2pt, font=\scriptsize, text=#1!55!black},
  arrow/.style={-{Latex[length=2.8mm]}, very thick, draw=black!55}
}

\node[card=blue] (data) at (0,0) {};
\node[badge=blue] at ([xshift=0.42cm,yshift=-0.42cm]data.north west) {1};
\node[title] at ([yshift=-0.55cm]data.north) {Datasets};
\node[sub] at ([yshift=-0.92cm]data.north) {industrial components yield a};
\node[sub] at ([yshift=-1.22cm]data.north) {semi-supervised task pool};
\node[chip=blue] at ([xshift=-1.05cm,yshift=0.38cm]data.center) {MVTec};
\node[chip=blue] at ([xshift=0.10cm,yshift=0.38cm]data.center) {VisA};
\node[chip=blue] at ([xshift=1.18cm,yshift=0.38cm]data.center) {BTAD};
\node[chip=blue] at ([xshift=-0.72cm,yshift=-0.12cm]data.center) {MPDD};
\node[chip=blue] at ([xshift=0.45cm,yshift=-0.12cm]data.center) {DAGM};
\node[chip=blue, draw=blue!65!black, very thick] at ([yshift=-0.78cm]data.center) {Multi-dataset};
\node[sub, text width=3.55cm, align=center] at ([yshift=-1.55cm]data.center) {within-dataset tasks plus a cross-dataset succession};

\node[card=teal] (scenarios) at (5.05,0) {};
\node[badge=teal] at ([xshift=0.42cm,yshift=-0.42cm]scenarios.north west) {2};
\node[title] at ([yshift=-0.55cm]scenarios.north) {Scenarios};
\node[sub] at ([yshift=-1.02cm]scenarios.north) {exploiting curriculum orderings};
\node[chip=teal] at ([yshift=0.42cm]scenarios.center) {Easy $\rightarrow$ Hard};
\node[chip=teal] at ([yshift=-0.12cm]scenarios.center) {Hard $\rightarrow$ Easy};
\node[chip=teal] at ([yshift=-0.66cm]scenarios.center) {Random};
\node[sub, text width=3.45cm, align=center] at ([yshift=-1.48cm]scenarios.center) {task ordering changes \\ adaptation and forgetting \\ dynamics};

\node[card=orange] (eval) at (10.10,0) {};
\node[badge=orange] at ([xshift=0.42cm,yshift=-0.42cm]eval.north west) {3};
\node[title] at ([yshift=-0.55cm]eval.north) {Evaluation};
\node[sub] at ([yshift=-1.05cm]eval.north) {retention and adaptation};
\node[chip=orange] at ([xshift=-0.72cm,yshift=0.42cm]eval.center) {Image-Level};
\node[chip=orange] at ([xshift=0.78cm,yshift=0.42cm]eval.center) {Pixel-Level};
\node[sub] at ([yshift=-0.14cm]eval.center) {NCL $\rightarrow$ CL};
\node[sub] at ([yshift=-0.45cm]eval.center) {ROC-AUC / BWT};
\node[chip=orange, draw=orange!65!black, very thick] at ([yshift=-0.85cm]eval.center) {9 methods};
\node[sub, text width=3.45cm, align=center] at ([yshift=-1.48cm]eval.center) {reconstruction-based, feature and patch distribution, teacher-student distillation, flow-based};

\draw[arrow] ([xshift=0.10cm]data.east) -- ([xshift=-0.10cm]scenarios.west);
\draw[arrow] ([xshift=0.10cm]scenarios.east) -- ([xshift=-0.10cm]eval.west);
\end{tikzpicture}
}
\caption{Our InCLAD benchmark at a glance. Tasks are drawn from five industrial datasets plus a multi-dataset stream, then arranged as curriculum or random continual scenario orderings. We evaluate 9 methods under non-continual (NCL) and continual (CL) protocols from both the image-level and pixel-level perspectives using ROC-AUC and BWT.}
\label{fig:benchmark-overview}
\end{figure}
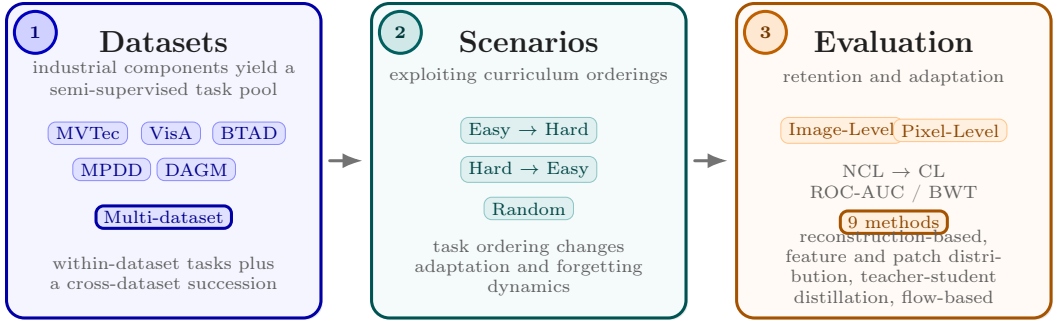

\section{Related Works}


\paragraph{Industrial anomaly detection.}
Industrial anomaly detection research has been driven by datasets such as MVTec
AD, MPDD, and VisA, which stress different aspects of visual defects,
from structural surface anomalies to logical inconsistencies in object
arrangement. These benchmarks established strong baselines for industrial visual
inspection, but they usually assume a static train/test split and independent
evaluation within each dataset. 
ADBench~\citep{han2022adbench} is a broad anomaly-benchmarking
work that systematizes evaluation across supervision regimes and
dataset types, but it is not specific to industrial visual inspection and does
not study sequential learning. In industrial vision, CableInspect-AD~\citep{arodi2024cableinspect}
introduces a challenging expert-annotated benchmark for robotic power-line
inspection, emphasizing realistic anomalies, cross-validation, and few-shot to
zero-shot evaluation. ReinAD~\citep{wang2025reinad}
argues that existing
industrial datasets are too controlled and proposes a contrastive benchmark
with unaligned reference/test pairs from production lines. Although these works are
highly relevant, they remain static benchmarks. In contrast, our benchmark
focuses on task-incremental continual learning, where components arrive
sequentially, models are trained only on normal samples, and evaluation
explicitly measures retention, forgetting, and adaptation under alternative
task orderings, including multi-dataset streams.
From a methodological perspective, recent research is also dominated by
non-continual methods that assume a static deployment setting. 
Reconstruction-based methods model the appearance manifold of normal samples and flag anomalies through
reconstruction mismatch or adversarially regularized feature space. GANomaly is
the representative reconstruction baseline in our experiments, combining an
encoder--decoder--encoder architecture with adversarial training for normality
modeling~\citep{ganomaly}. More recent reconstruction and restoration variants
strengthen this idea with richer architectures and objectives, including
semantic inpainting~\citep{tao2022dualsiamese}, masked-patch
reconstruction~\citep{jiang2023mstunet}, multilevel feature
restoration~\citep{guo2024mldfr}, and reconstruction regularization with memory
or text guidance~\citep{zhang2024dcae,qiang2025tgard}.

Another major family relies on pretrained features, patchwise distributions, or
memory-based matching rather than explicit image reconstruction.
PatchCore~\citep{DBLP:conf/cvpr/RothPZSBG22} stores a coreset of representative
normal patch embeddings and scores test patches by nearest-neighbor distance,
and PaDiM~\citep{DBLP:conf/icpr/DefardSLA20} fits Gaussian models to patchwise
features at each spatial location. CFA~\citep{lee2022cfa} complements these
memory- and distribution-based approaches by learning compact normal feature
representations through a soft-boundary objective.

Teacher--student and distillation-based methods form a further relevant thread.
STFPM~\citep{DBLP:journals/corr/abs-2103-04257} aligns multiscale teacher and
student feature pyramids so that anomalies emerge where the student fails to
reproduce normal teacher activations. RD4AD~\citep{DBLP:conf/cvpr/DengL22}
reconstructs teacher embeddings from compressed one-class representations,
EfficientAD~\citep{DBLP:conf/wacv/BatznerHK24} retains the teacher--student
intuition while targeting millisecond-level inference, and
PaSTe~\citep{barusco2025paste} pushes this direction toward edge deployment.

Finally, density estimation methods such as 
FastFlow~\citep{DBLP:journals/corr/abs-2111-07677} model the density of normal
samples using normalizing flows and provides a complementary probabilistic
perspective to reconstruction, patch-memory, and teacher--student detectors.
Recent work also extends industrial anomaly detection toward multimodal
inspection, more realistic static benchmarks, and weakly supervised
localization~\citep{cheng2025ucf,wang2025reinad,saliency_guided_transformer_2026}.


However, several gaps remain directly relevant to our work. First, even methods trained only on normal
data are almost always evaluated in isolation on a single static benchmark.
Second, this literature rarely studies how image-level detection and pixel-level
localization behave together under sequential learning. Third, static
evaluations do not measure retention after a model encounters new industrial
concepts. As a result, the non-continual state of the art provides strong static
baselines but leaves open the task-incremental, normal-only, multi-dataset
setting that we target.

\paragraph{Continual learning for industrial inspection.}
Continual learning provides a natural framework for industrial settings where
data arrive in stages and the target distribution evolves. Most continual
learning benchmarks, however, focus on closed-set classification. Industrial
anomaly detection differs in three ways: it is anomaly-centric, it is often
trained only on normal samples, and it may require localization in addition to
image-level decisions. These properties motivate a benchmark in which anomaly
sensitivity and forgetting control are evaluated jointly. Direct continual
anomaly-detection work is still rare, but the recent literature already
suggests several distinct continual-learning strategies. A first line focuses
on online updating and
lightweight rehearsal. \cite{gao2023totalonline} proposed
LeMO, a fully online anomaly-detection and localization method that replaces
offline pre-collection with a learnable memory updated jointly with feature
extraction through a contrastive objective. \cite{li2022continualadaptation} explicitly introduced continual anomaly
detection for industrial inspection and proposed the distribution of normal
embeddings, which summarizes previous tasks through normal-feature
distributions and leverages synthetic anomalies so that anomaly detection can
benefit from supervised continual-learning machinery. In a more
deployment-oriented direction, \cite{barusco2025memoryefficient} studied continual anomaly detection on
edge devices, showing that replay and compressed replay can make lightweight
teacher--student models such as STFPM and PaSTe practical under strict memory
budgets.

Other studies pursue parameter-efficient adaptation or stronger generative
mechanisms. \cite{liu2024ucad} proposed UCAD, where a
key--prompt--knowledge memory bank and structure-based contrastive learning
yield a prompt-based continual learner that separates task-specific normal
knowledge from task-invariant anomaly cues. \cite{tang2024iuf} integrated incremental learning into a unified
reconstruction framework for small defect inspection, using object-aware
self-attention and semantic compression to reduce feature conflicts without
storing large feature banks. Diffusion-based methods have become especially
prominent in recent work. \cite{li2025oneformore}
introduced One-for-More, a continual diffusion model that combines gradient
projection, an iterative singular-value-decomposition scheme for memory
reduction, and anomaly-masked conditioning to limit forgetting and hallucinated
reconstructions. \cite{hu2025replaycad} then pushed replay
further with ReplayCAD, where a pretrained diffusion model is used for
generative replay to improve especially anomaly segmentation. Finally,
continual anomaly detection is beginning to expand
beyond 2D RGB inspection: \cite{lu2025c3dad} proposed C3D-AD
for 3D anomaly detection, combining kernel-attention feature normalization, a
learnable advisor, and representation-preserving parameter perturbation to
handle emerging point-cloud categories. A task-aware parameter decoupling
framework  proposed by \cite{zhang2026taskaware} moves from one-model-per-category
assumptions to a one-for-more setting through instance-aware prompt tuning and
gradient-aware parameter decoupling.

The current continual literature remains much narrower than the
broader non-continual anomaly-detection landscape, but it already spans online
memory learning, replay-based and generative-replay strategies, prompt-based
adaptation, and architectural mechanisms for incremental reconstruction.
However, important limitations remain. Several approaches still incorporate
supervision such as synthetic anomalies or explicit defect-oriented
signals, which weakens the match to purely normal-only industrial training.
Moreover, existing continual studies are typically limited to one dataset or
one scenario at a time, so they do not provide a cohesive analysis across
diverse benchmarks, alternative task orderings, and cross-dataset streams.
This leaves curriculum effects and multi-dataset continual anomaly-detection
scenarios largely unexplored, even though both are central to understanding how
task difficulty and broader domain shift affect forgetting.
Finally, the literature does not yet systematically compare image-level and
pixel-level continual anomaly detection within the same benchmark design. These
gaps are precisely what motivate the benchmark proposed in this paper.

\section{The InCLAD Benchmark}

The proposed benchmark is defined within industrial visual inspection. Every task concerns anomaly detection in manufactured products or industrial scenes, but each task corresponds to a different inspected component or concept. In this task-incremental framing, distribution shift is driven
primarily by component-specific normality rather than by supervised defect-class
changes. This matches deployment settings in which inspection systems should
absorb new product variants incrementally without losing competence on previous
components.

InCLAD is therefore a benchmark protocol rather than a raw-data collection
effort: it builds on public industrial datasets and contributes a continual
learning layer of task definitions, orderings, metrics, and evaluation artifacts
over them.

\textbf{Task Definition}: We consider a stream of tasks
$\mathcal{T}_1,\mathcal{T}_2,\ldots,\mathcal{T}_N$. Each task contains normal
training samples, anomalous test samples (only in the test portion of the task), and, when available, pixel-level masks for anomalous regions. The benchmark is evaluated from two complementary
perspectives: image-level anomaly detection, which asks whether an image is
abnormal, and pixel-level anomaly localization, which asks where the anomalous
region appears.
{
The task identity is tied to the inspected component or concept rather than to a
defect label. Across a stream, shifts may arise from new products, materials, or
surface geometries; from different industrial datasets; from acquisition changes
such as illumination, camera configuration, viewpoint, resolution, or background;
or from temporal and batch evolution in production. The key property is that the
inspection objective remains anomaly detection while the definition of normality
changes over time.
}

\textbf{Single-Dataset Scenarios}: We instantiate task streams through an empirical scenario-extraction procedure
based on single-task anomaly-detection difficulty. The first stage starts from
raw dataset concepts and converts each dataset into three continual-learning
scenarios. For each concept, we train Single-Task Experts independently using
three representative state-of-the-art methods: PatchCore~\citep{DBLP:conf/cvpr/RothPZSBG22},
PaDiM~\citep{DBLP:conf/icpr/DefardSLA20}, and CFA~\citep{lee2022cfa}, and
collect concept-level ROC-AUC scores. This verifies that candidate tasks are
learnable, with ROC-AUC values above the random-guessing level of $0.5$, and
provides a task-wise difficulty estimate that is independent of any
continual-learning strategy.
To avoid defining difficulty from a single detector, we rank concepts separately
for PatchCore, PaDiM, and CFA, then average the ranks across the three methods.
These average ranks define the within-dataset curricula. The easy-to-hard
sequence orders tasks from easiest to hardest, the hard-to-easy sequence
reverses this order, and the random sequence provides a non-curriculum
reference. This yields three within-dataset scenarios per dataset. The same
scenario definitions are used for image-level and pixel-level evaluation,
enabling a direct comparison between detection and localization under identical
task streams.

\begin{figure*}[t]
\centering
\begin{minipage}{0.32\textwidth}
\centering
\includegraphics[width=\linewidth]{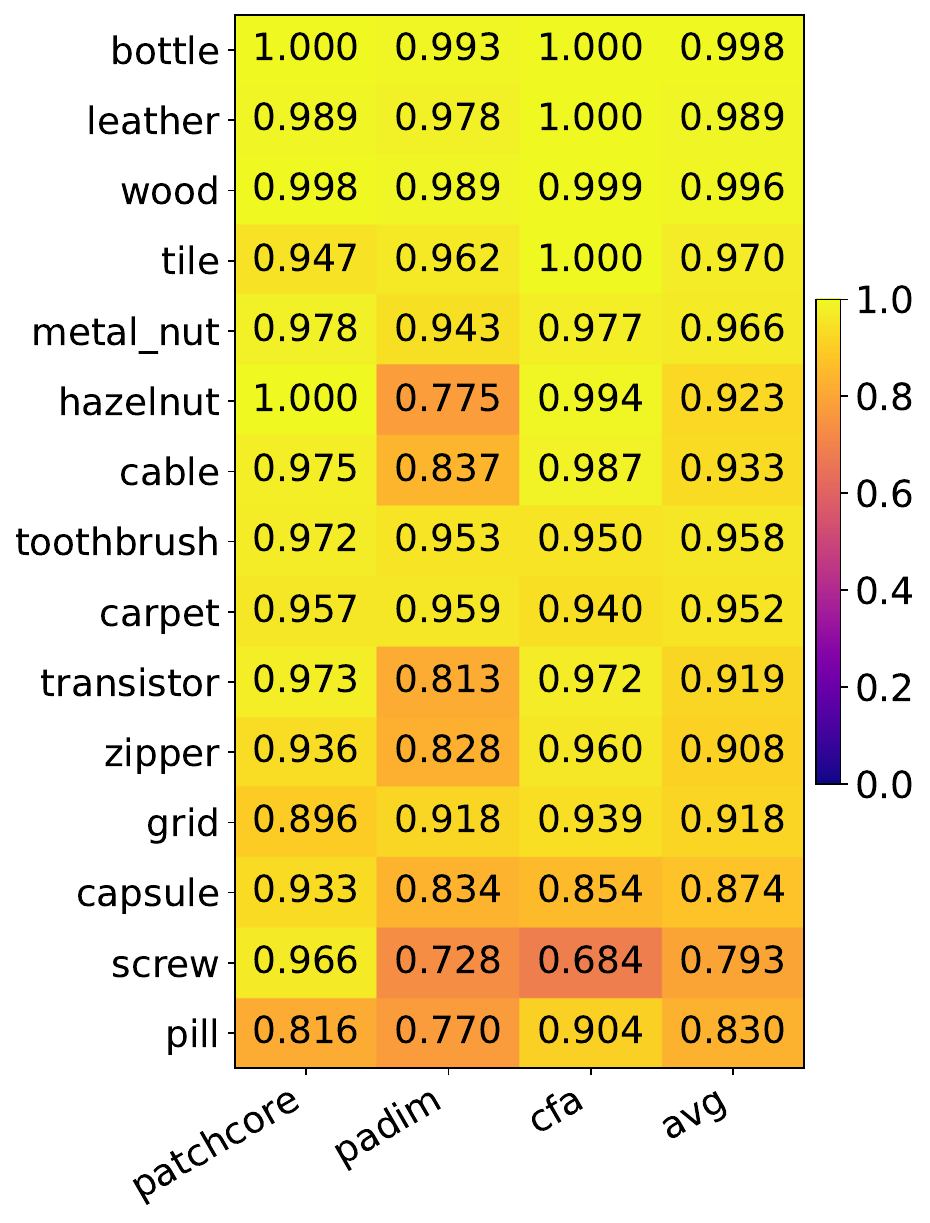}\\
\small MVTec AD
\end{minipage}\hfill
\begin{minipage}{0.32\textwidth}
\centering
\includegraphics[width=\linewidth]{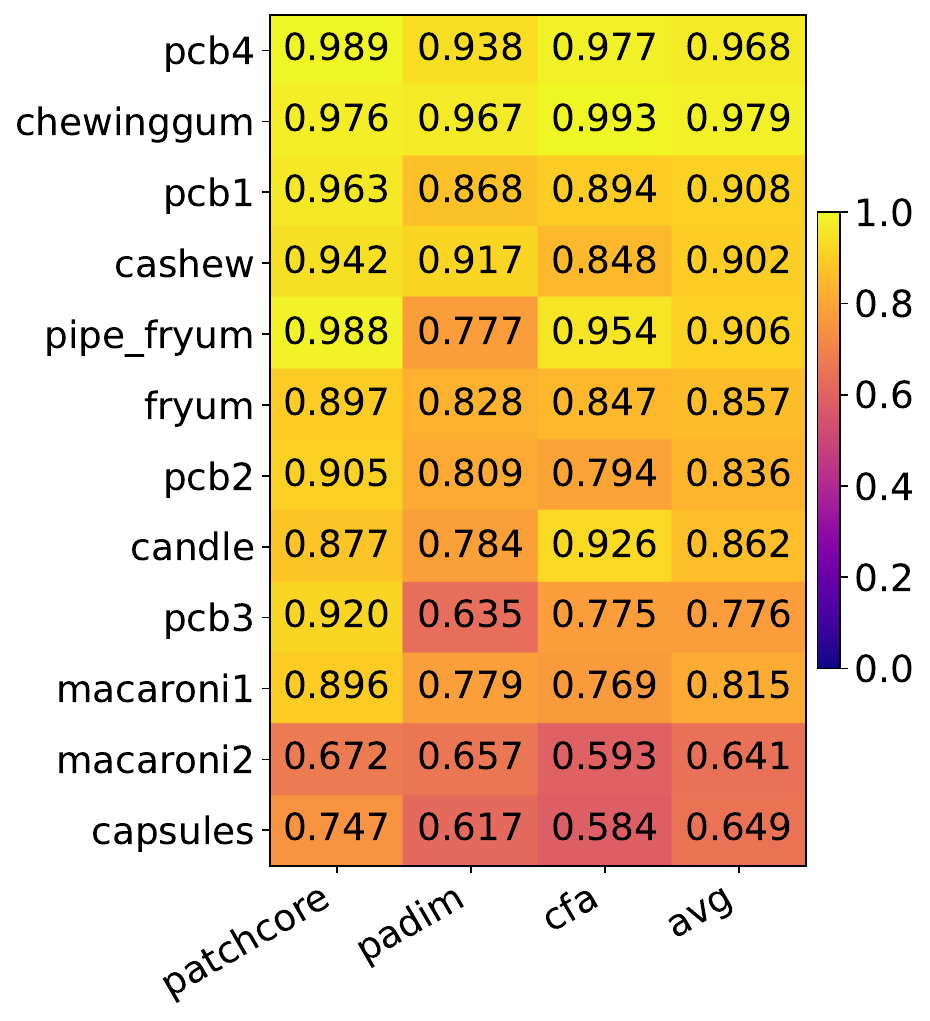}\\
\small VisA
\end{minipage}\hfill
\begin{minipage}{0.32\textwidth}
\centering
\includegraphics[width=\linewidth]{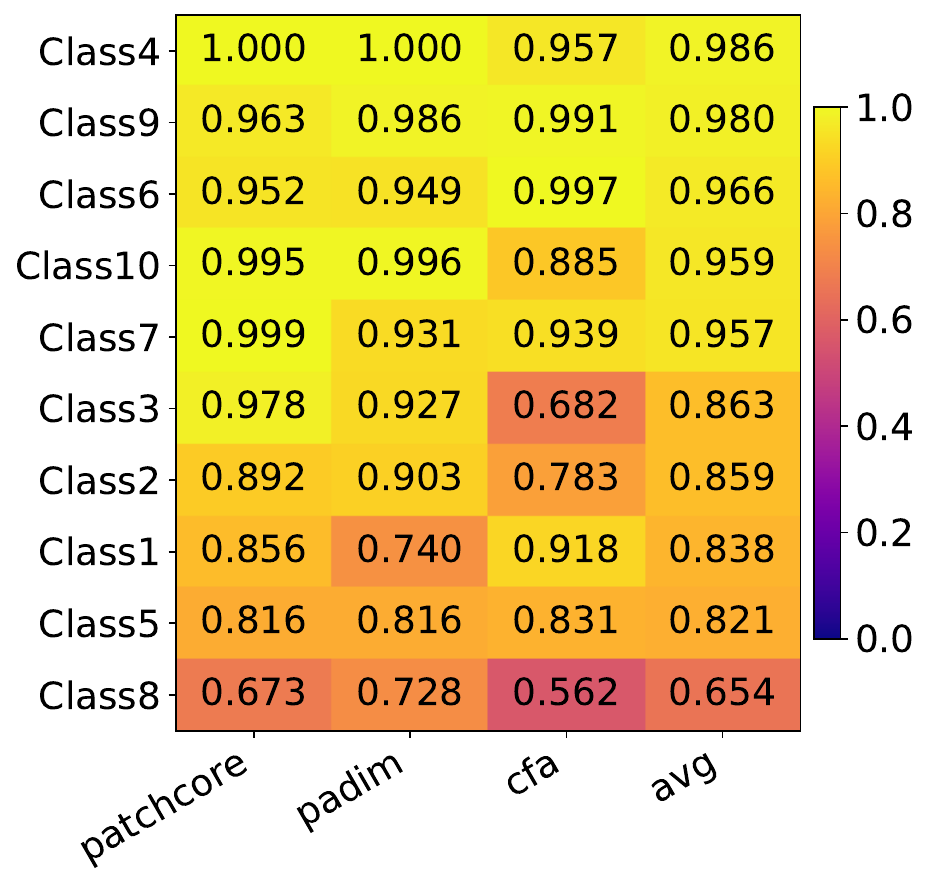}\\
\small DAGM
\end{minipage}
\begin{minipage}{0.42\textwidth}
\centering
\includegraphics[width=\linewidth]{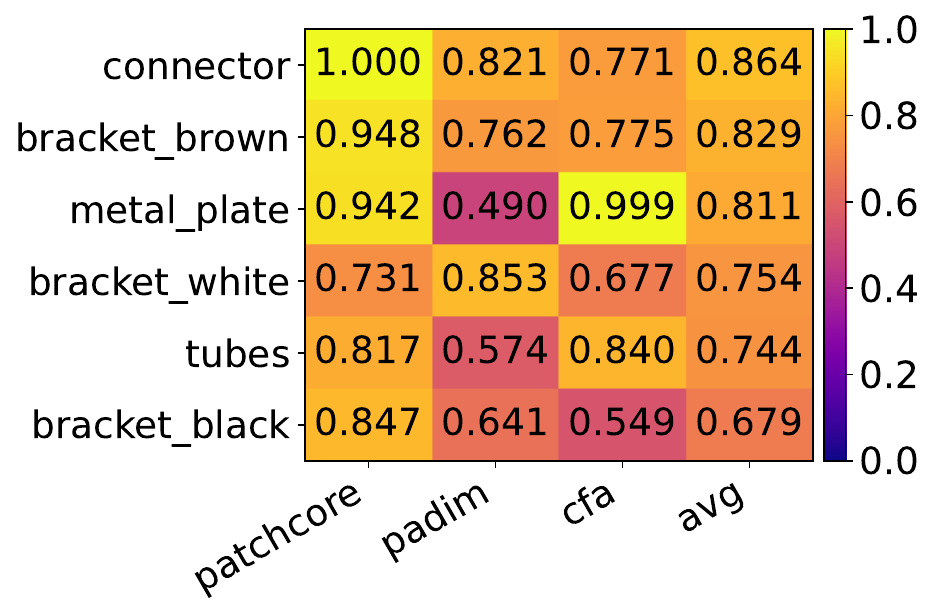}\\
\small MPDD
\end{minipage}\hfill
\begin{minipage}{0.42\textwidth}
\centering
\includegraphics[width=\linewidth]{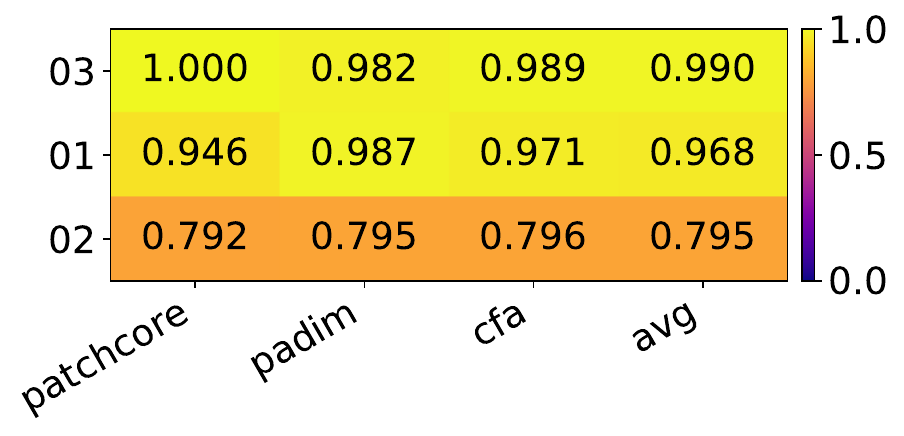}\\
\small BTAD
\end{minipage}
\caption{Single-task ROC-AUC heatmaps used to construct the benchmark scenarios.
For each dataset, rows are ordered by the consensus curriculum induced by the
average rank across PatchCore, PaDiM, and CFA, from easier concepts at the top
to harder concepts at the bottom. The heatmap entries are Single-Task Expert
ROC-AUC values, and the final column reports their average.}
\label{fig:ordering-heatmaps}
\end{figure*}

Figure~\ref{fig:ordering-heatmaps} visualizes the first-stage evidence used to
derive these scenarios. In each heatmap, rows correspond to concepts sorted by
the consensus ordering induced by the average rank, cell values report the
ROC-AUC obtained by each Single-Task Expert, and the final column reports the
average ROC-AUC across PatchCore, PaDiM, and CFA. We use these three methods
because they are competitive industrial anomaly detectors with distinct
inductive biases: PatchCore uses memory-based nearest-neighbor retrieval,
PaDiM models location-aware patch distributions with Gaussian statistics, and
CFA enforces compactness of the normal feature space through hypersphere-based
objectives. This reduces the risk that the scenario definition is dominated by
one detector family. The exact hard-to-easy rankings and pairwise Spearman
correlations are reported in Appendix~\ref{app:ordering-details}.

{
\textbf{Multi-Dataset Scenarios}: Beyond within-dataset streams, we define a multi-dataset scenario in which the learner encounters tasks drawn from the full dataset pool. This setting combines
task-level changes with dataset-level distribution shifts, making it a stress
test for retention and adaptation under broader industrial variability.
The multi-dataset stream is constructed from a balanced subset of tasks selected
according to the Single-Task Expert difficulty estimates. For each dataset with
more than six tasks, we retain six representative concepts using the average
ROC-AUC column in the difficulty heatmaps: the two easiest tasks, the two
hardest tasks, and two median-difficulty tasks. For datasets with six or fewer
tasks, we retain all available tasks. This rule preserves both extremes of the
difficulty spectrum while preventing the largest datasets from dominating the
stream, and it yields a 27-task multi-dataset scenario.
After task selection, we define two cross-dataset orderings: easy-to-hard and
hard-to-easy, obtained by sorting the retained tasks according to the same
average-difficulty criterion.
}

\textbf{Evaluation Protocol}: At step $t$, a model observes normal training data from task $\mathcal{T}_t$ and
updates under a continual-learning constraint, meaning unrestricted access to all
past data is not assumed. After training on $\mathcal{T}_t$, the model is
evaluated on the test sets of all tasks seen so far,
$\mathcal{T}_1,\ldots,\mathcal{T}_t$. The goal is to adapt to the current visual
regime while retaining anomaly-detection ability on earlier tasks.
We report standard anomaly-detection metrics for both perspectives: image-level
ROC-AUC for detection and pixel-level ROC-AUC when segmentation masks are
available. Continual-learning behavior is summarized through final average
performance and backward transfer (BWT). In particular, our continual ROC-AUC
averages the ROC-AUC values obtained throughout the learner's lifetime after
each task update, while BWT measures how later tasks change performance on
previously learned tasks. Negative BWT therefore indicates forgetting, whereas
positive BWT indicates beneficial backward transfer. Formal definitions are
provided in Appendix~\ref{app:benchmark-metrics}. This prevents benchmark
progress from being reduced to current-task performance alone.
To choose the detectors used for the multi-dataset scenario stress test, we
average single-dataset performance across datasets, backbones, and orderings,
then rank the 9 evaluated methods. We evaluate the three highest-performing
methods under this aggregate criterion---CFA, PaSTe, and PaDiM---on the
multi-dataset stream. This is a compute-aware stress test rather than
cherry-picking: the selection rule is fixed before the multi-dataset evaluation
and uses aggregate single-dataset performance across methods, backbones, and
orderings.

\begin{table}[t]
  \centering
  \setlength{\tabcolsep}{14pt}
  \small
  \caption{Overview of the proposed visual anomaly detection scenarios. Each scenario comes in two curriculum orderings (easy-to-hard, hard-to-easy). Single-dataset scenarios are also provided with one additional random ordering, leading to 17 scenarios in total.}
  \label{tab:scenario_overview_visual}
  \begin{tabular}{l c c c}
    \toprule
    \textbf{Scenario} & \textbf{\# Tasks} & \textbf{\# Data samples} & \textbf{\# Anomalies (test)} \\
    \midrule
    BTAD              & 3  & 2,540  & 39.14\% \\
    DAGM              & 10 & 15,054 & 13.09\% \\
    MPDD              & 6  & 1,346  & 61.57\% \\
    MVTec AD          & 15 & 5,354  & 72.93\% \\
    VisA              & 12 & 10,821 & 55.50\% \\
    MultiDataset & 27 & 20,853 & 29.38\% \\
    \bottomrule
  \end{tabular}
\end{table}


Table~\ref{tab:scenario_overview_visual} summarizes our benchmark consisting of
17 scenarios with varying length, sample count, and test-time anomaly ratio,
from compact single-dataset streams to the 27-task multi-dataset scenario. The
released artifacts include scenario manifests, code, task orderings, split
definitions, documentation, and Croissant plus RAI metadata. The scenarios are
available at
\url{https://huggingface.co/datasets/lifelonglab/inclad-benchmark}. The source code
for our framework is available at
\url{https://github.com/LukaszMarcjan/InCLAD-benchmark}.

\section{Experiments}
\label{sec:experiments}

\subsection{Experimental Setup}

All experiments use the benchmark protocol defined above. We evaluate each
method on identical task streams, train/test splits, and ordering metadata, and
compare naive non-continual baselines against their continual variants at both
image and pixel level. Unless otherwise specified, continual versions use replay;
PatchCore, PaDiM, and CFA require method-specific continual adaptations because
their original formulations do not directly support replay-based updates.
Detailed metric definitions, continual adaptations, execution-time measurements,
and hyperparameter settings are provided in
Appendices~\ref{app:benchmark-metrics}, \ref{app:continual-adaptations},
\ref{app:empirical-time-complexity}, and~\ref{app:reproducibility}.

\subsection{Industrial Datasets}

We build our benchmark upon several public industrial datasets. 
%
%
\textbf{DAGM}~\citep{weimer2016dagm} is a long-standing industrial inspection benchmark
built around visually subtle defects embedded in heavily textured backgrounds.
Its value lies in separating weak defect cues from repetitive texture, which is
relevant when anomalies are small relative to the background structure.
%
%
%
Following the protocol summarized in~\citep{DBLP:journals/tii/XuZFN25}, we also consider 
\textbf{MVTec AD}~\citep{DBLP:conf/cvpr/BergmannFSS19}, and
\textbf{VisA}~\citep{DBLP:conf/eccv/ZouJPZD22}. These benchmarks are complementary
industrial anomaly-detection scenarios. 
MVTec AD contains 15 real-world
categories and is a standard benchmark for structural anomaly localization.
VisA covers 12 object categories with both surface and structural
defects under varying object configurations. 
\textbf{BTAD}~\citep{BTAD_website} is a real-world industrial anomaly benchmark introduced
for manufacturing inspection and anomaly-detection research. It contains 2540
images of three product categories, commonly denoted as \texttt{product\_1},
\texttt{product\_2}, and \texttt{product\_3}, together with pixel-level
annotations that can be interpreted as instance segmentation masks and
transformed into semantic segmentation or object-detection labels. Of the 2540
images, 2261 are unlabeled and 279 carry anomaly annotations, and the dataset
can also be viewed through an \texttt{ok}/\texttt{ko} split with 2250 normal and
290 anomalous images. BTAD is useful because it combines pixel-level
localization with a semisupervised industrial setting in which only a limited
subset of images is annotated.
\textbf{MPDD}~\citep{MPDD_website} is a dataset for visual defect detection in industrial
metal-parts manufacturing. It contains more than 1000 images with pixel-precise
defect masks and is explicitly designed to benchmark methods under realistic
conditions for metal-part inspection. The dataset is divided into a training
subset containing anomaly-free samples and a validation subset containing both
normal and anomalous samples. MPDD is especially relevant here because it targets
metallic industrial components, making it closer in spirit to materials-oriented
quality-control problems than many consumer-object anomaly datasets.

Dataset-license details and compute resources are reported in
Appendix~\ref{app:reproducibility}.


\begin{table*}[!t]
\centering
\caption{ROC-AUC and BWT averaged across methods and ResNet18/WideResNet backbones. NCL denotes the naive baseline, while CL denotes the continual variant.} 
\label{tab:summary-avg-dataset-backbones}
\small
\setlength{\tabcolsep}{3pt}
\renewcommand{\arraystretch}{0.88}
\begin{tabular}{lllcccc}
\toprule
\multirow{2}{*}{Dataset} & \multirow{2}{*}{Ordering} & \multirow{2}{*}{Metric} & \multicolumn{2}{c}{Image-level} & \multicolumn{2}{c}{Pixel-level} \\
\cmidrule(lr){4-5}\cmidrule(lr){6-7}
 &  &  & NCL & CL & NCL & CL \\
\midrule


\multirow{6}{*}{MVTec} & \multirow{2}{*}{Easy-to-Hard} & ROC$\uparrow$ & \(0.583\,{\scriptstyle\pm\,0.095}\) & \(0.784\,{\scriptstyle\pm\,0.146}\) & \(0.629\,{\scriptstyle\pm\,0.128}\) & \(0.842\,{\scriptstyle\pm\,0.105}\) \\
 &  & BWT$\uparrow$ & \(-0.039\,{\scriptstyle\pm\,0.018}\) & \(-0.017\,{\scriptstyle\pm\,0.015}\) & \(-0.035\,{\scriptstyle\pm\,0.008}\) & \(-0.010\,{\scriptstyle\pm\,0.008}\) \\
 \cmidrule(lr){4-7}
 & \multirow{2}{*}{Hard-to-Easy} & ROC$\uparrow$ & \(0.557\,{\scriptstyle\pm\,0.062}\) & \(0.710\,{\scriptstyle\pm\,0.117}\) & \(0.623\,{\scriptstyle\pm\,0.125}\) & \(0.839\,{\scriptstyle\pm\,0.103}\) \\
 &  & BWT$\uparrow$ & \(-0.042\,{\scriptstyle\pm\,0.016}\) & \(-0.017\,{\scriptstyle\pm\,0.015}\) & \(-0.042\,{\scriptstyle\pm\,0.010}\) & \(-0.010\,{\scriptstyle\pm\,0.006}\) \\
 \cmidrule(lr){4-7}
 & \multirow{2}{*}{Random} & ROC$\uparrow$ & \(0.593\,{\scriptstyle\pm\,0.037}\) & \(0.766\,{\scriptstyle\pm\,0.124}\) & \(0.679\,{\scriptstyle\pm\,0.080}\) & \(0.858\,{\scriptstyle\pm\,0.090}\) \\
 &  & BWT$\uparrow$ & \(-0.041\,{\scriptstyle\pm\,0.009}\) & \(-0.020\,{\scriptstyle\pm\,0.016}\) & \(-0.030\,{\scriptstyle\pm\,0.007}\) & \(-0.012\,{\scriptstyle\pm\,0.010}\) \\
\cmidrule(lr){1-7}
\multirow{6}{*}{VisA} & \multirow{2}{*}{Easy-to-Hard} & ROC$\uparrow$ & \(0.549\,{\scriptstyle\pm\,0.045}\) & \(0.748\,{\scriptstyle\pm\,0.109}\) & \(0.784\,{\scriptstyle\pm\,0.090}\) & \(0.886\,{\scriptstyle\pm\,0.081}\) \\
 &  & BWT$\uparrow$ & \(-0.056\,{\scriptstyle\pm\,0.019}\) & \(-0.019\,{\scriptstyle\pm\,0.021}\) & \(-0.031\,{\scriptstyle\pm\,0.013}\) & \(-0.013\,{\scriptstyle\pm\,0.015}\) \\
 \cmidrule(lr){4-7}
 & \multirow{2}{*}{Hard-to-Easy} & ROC$\uparrow$ & \(0.556\,{\scriptstyle\pm\,0.023}\) & \(0.677\,{\scriptstyle\pm\,0.070}\) & \(0.765\,{\scriptstyle\pm\,0.092}\) & \(0.866\,{\scriptstyle\pm\,0.092}\) \\
 &  & BWT$\uparrow$ & \(-0.046\,{\scriptstyle\pm\,0.016}\) & \(-0.017\,{\scriptstyle\pm\,0.020}\) & \(-0.026\,{\scriptstyle\pm\,0.010}\) & \(-0.009\,{\scriptstyle\pm\,0.009}\) \\
 \cmidrule(lr){4-7}
 & \multirow{2}{*}{Random} & ROC$\uparrow$ & \(0.555\,{\scriptstyle\pm\,0.040}\) & \(0.703\,{\scriptstyle\pm\,0.092}\) & \(0.777\,{\scriptstyle\pm\,0.079}\) & \(0.877\,{\scriptstyle\pm\,0.075}\) \\
 &  & BWT$\uparrow$ & \(-0.049\,{\scriptstyle\pm\,0.019}\) & \(-0.022\,{\scriptstyle\pm\,0.016}\) & \(-0.035\,{\scriptstyle\pm\,0.015}\) & \(-0.009\,{\scriptstyle\pm\,0.008}\) \\
\cmidrule(lr){1-7}

\multirow{6}{*}{BTAD} & \multirow{2}{*}{Easy-to-Hard} & ROC$\uparrow$ & \(0.666\,{\scriptstyle\pm\,0.094}\) & \(0.861\,{\scriptstyle\pm\,0.148}\) & \(0.773\,{\scriptstyle\pm\,0.086}\) & \(0.925\,{\scriptstyle\pm\,0.080}\) \\
 &  & BWT$\uparrow$ & \(-0.289\,{\scriptstyle\pm\,0.133}\) & \(-0.046\,{\scriptstyle\pm\,0.093}\) & \(-0.247\,{\scriptstyle\pm\,0.039}\) & \(-0.010\,{\scriptstyle\pm\,0.014}\) \\
  \cmidrule(lr){4-7}
 & \multirow{2}{*}{Hard-to-Easy} & ROC$\uparrow$ & \(0.694\,{\scriptstyle\pm\,0.129}\) & \(0.772\,{\scriptstyle\pm\,0.137}\) & \(0.739\,{\scriptstyle\pm\,0.090}\) & \(0.869\,{\scriptstyle\pm\,0.090}\) \\
 &  & BWT$\uparrow$ & \(-0.145\,{\scriptstyle\pm\,0.086}\) & \(-0.031\,{\scriptstyle\pm\,0.068}\) & \(-0.216\,{\scriptstyle\pm\,0.040}\) & \(-0.021\,{\scriptstyle\pm\,0.029}\) \\
  \cmidrule(lr){4-7}
 & \multirow{2}{*}{Random} & ROC$\uparrow$ & \(0.706\,{\scriptstyle\pm\,0.054}\) & \(0.828\,{\scriptstyle\pm\,0.118}\) & \(0.759\,{\scriptstyle\pm\,0.080}\) & \(0.883\,{\scriptstyle\pm\,0.091}\) \\
 &  & BWT$\uparrow$ & \(-0.185\,{\scriptstyle\pm\,0.130}\) & \(-0.025\,{\scriptstyle\pm\,0.033}\) & \(-0.200\,{\scriptstyle\pm\,0.030}\) & \(-0.010\,{\scriptstyle\pm\,0.014}\) \\
\cmidrule(lr){1-7}

\multirow{6}{*}{MPDD} & \multirow{2}{*}{Easy-to-Hard} & ROC$\uparrow$ & \(0.541\,{\scriptstyle\pm\,0.066}\) & \(0.675\,{\scriptstyle\pm\,0.136}\) & \(0.797\,{\scriptstyle\pm\,0.133}\) & \(0.888\,{\scriptstyle\pm\,0.079}\) \\
 &  & BWT$\uparrow$ & \(-0.112\,{\scriptstyle\pm\,0.052}\) & \(-0.061\,{\scriptstyle\pm\,0.048}\) & \(-0.039\,{\scriptstyle\pm\,0.017}\) & \(-0.009\,{\scriptstyle\pm\,0.008}\) \\
 \cmidrule(lr){4-7}
 & \multirow{2}{*}{Hard-to-Easy} & ROC$\uparrow$ & \(0.592\,{\scriptstyle\pm\,0.035}\) & \(0.687\,{\scriptstyle\pm\,0.100}\) & \(0.791\,{\scriptstyle\pm\,0.142}\) & \(0.892\,{\scriptstyle\pm\,0.076}\) \\
 &  & BWT$\uparrow$ & \(-0.095\,{\scriptstyle\pm\,0.046}\) & \(-0.038\,{\scriptstyle\pm\,0.043}\) & \(-0.042\,{\scriptstyle\pm\,0.030}\) & \(-0.008\,{\scriptstyle\pm\,0.013}\) \\
  \cmidrule(lr){4-7}
 & \multirow{2}{*}{Random} & ROC$\uparrow$ & \(0.591\,{\scriptstyle\pm\,0.058}\) & \(0.681\,{\scriptstyle\pm\,0.097}\) & \(0.784\,{\scriptstyle\pm\,0.144}\) & \(0.891\,{\scriptstyle\pm\,0.082}\) \\
 &  & BWT$\uparrow$ & \(-0.108\,{\scriptstyle\pm\,0.056}\) & \(-0.057\,{\scriptstyle\pm\,0.047}\) & \(-0.080\,{\scriptstyle\pm\,0.055}\) & \(-0.010\,{\scriptstyle\pm\,0.013}\) \\
\cmidrule(lr){1-7}
\multirow{6}{*}{DAGM} & \multirow{2}{*}{Easy-to-Hard} & ROC$\uparrow$ & \(0.557\,{\scriptstyle\pm\,0.224}\) & \(0.751\,{\scriptstyle\pm\,0.151}\) & \(0.631\,{\scriptstyle\pm\,0.269}\) & \(0.808\,{\scriptstyle\pm\,0.166}\) \\
 &  & BWT$\uparrow$ & \(-0.035\,{\scriptstyle\pm\,0.019}\) & \(-0.011\,{\scriptstyle\pm\,0.014}\) & \(-0.025\,{\scriptstyle\pm\,0.015}\) & \(-0.007\,{\scriptstyle\pm\,0.007}\) \\
  \cmidrule(lr){4-7}
 & \multirow{2}{*}{Hard-to-Easy} & ROC$\uparrow$ & \(0.547\,{\scriptstyle\pm\,0.213}\) & \(0.694\,{\scriptstyle\pm\,0.098}\) & \(0.618\,{\scriptstyle\pm\,0.245}\) & \(0.779\,{\scriptstyle\pm\,0.122}\) \\
 &  & BWT$\uparrow$ & \(-0.037\,{\scriptstyle\pm\,0.026}\) & \(-0.013\,{\scriptstyle\pm\,0.014}\) & \(-0.027\,{\scriptstyle\pm\,0.017}\) & \(-0.006\,{\scriptstyle\pm\,0.006}\) \\
  \cmidrule(lr){4-7}
 & \multirow{2}{*}{Random} & ROC$\uparrow$ & \(0.539\,{\scriptstyle\pm\,0.210}\) & \(0.713\,{\scriptstyle\pm\,0.128}\) & \(0.614\,{\scriptstyle\pm\,0.248}\) & \(0.795\,{\scriptstyle\pm\,0.137}\) \\
 &  & BWT$\uparrow$ & \(-0.034\,{\scriptstyle\pm\,0.022}\) & \(-0.015\,{\scriptstyle\pm\,0.015}\) & \(-0.028\,{\scriptstyle\pm\,0.016}\) & \(-0.007\,{\scriptstyle\pm\,0.006}\) \\
\midrule
\multirow{4}{*}{MultiDataset} & \multirow{2}{*}{Easy-to-Hard} & ROC$\uparrow$ & \(0.526\,{\scriptstyle\pm\,0.013}\) & \(0.792\,{\scriptstyle\pm\,0.026}\) & \(0.711\,{\scriptstyle\pm\,0.056}\) & \(0.925\,{\scriptstyle\pm\,0.019}\) \\
 &  & BWT$\uparrow$ & \(-0.026\,{\scriptstyle\pm\,0.002}\) & \(-0.004\,{\scriptstyle\pm\,0.005}\) & \(-0.017\,{\scriptstyle\pm\,0.004}\) & \(-0.001\,{\scriptstyle\pm\,0.001}\) \\
\cmidrule(lr){4-7}
 & \multirow{2}{*}{Hard-to-Easy} & ROC$\uparrow$ & \(0.552\,{\scriptstyle\pm\,0.018}\) & \(0.785\,{\scriptstyle\pm\,0.044}\) & \(0.713\,{\scriptstyle\pm\,0.044}\) & \(0.904\,{\scriptstyle\pm\,0.037}\) \\
 &  & BWT$\uparrow$ & \(-0.023\,{\scriptstyle\pm\,0.000}\) & \(-0.006\,{\scriptstyle\pm\,0.006}\) & \(-0.023\,{\scriptstyle\pm\,0.008}\) & \(-0.003\,{\scriptstyle\pm\,0.002}\) \\
\bottomrule
\end{tabular}
\end{table*}

\subsection{Anomaly Detection Methods}

Our experimental comparison includes representative non-continual baselines
from the main method families reviewed in Related Works. Specifically, we
consider nine methods: the \textbf{reconstruction-based} baseline
GANomaly~\citep{ganomaly}; \textbf{pretrained-feature and patch-distribution}
methods PatchCore~\citep{DBLP:conf/cvpr/RothPZSBG22},
PaDiM~\citep{DBLP:conf/icpr/DefardSLA20}, and CFA~\citep{lee2022cfa};
\textbf{teacher--student and distillation} methods
STFPM~\citep{DBLP:journals/corr/abs-2103-04257},
RD4AD~\citep{DBLP:conf/cvpr/DengL22},
EfficientAD~\citep{DBLP:conf/wacv/BatznerHK24}, and
PaSTe~\citep{barusco2025paste}; and the \textbf{flow-based} baseline
FastFlow~\citep{DBLP:journals/corr/abs-2111-07677}.
This set of methods is intended to cover the dominant non-continual design
choices used in current industrial anomaly detection while keeping the
comparison manageable across datasets and evaluation protocols.

For the continual versions, six of the nine methods are adapted with the
standard replay protocol: at each task, the model is updated using the current
normal training samples together with samples retained from previous tasks. In
contrast, PatchCore, PaDiM, and CFA require non-trivial algorithmic adaptations
because their original formulations are not conventional parametric learners to
which replay can be directly applied. PatchCore stores a coreset memory bank of
normal patch embeddings; our continual version maintains task-specific patch
banks, stores the current task at the native coreset resolution, compresses
previous banks with a distance-weighted greedy coreset, and scores each test
image against the closest task bank. PaDiM fits location-aware Gaussian
statistics rather than training with gradient updates; our continual version
therefore stores one Gaussian model per task and selects, at inference time, the
task distribution that yields the lowest image-level Mahalanobis anomaly score.
CFA combines a trainable descriptor with a memory bank used by its
soft-boundary objective. For its continual version, we train the descriptor on
the current task plus replay samples, rebuild all per-task inference banks after
each descriptor update so that they remain in the same embedding space, and
select the bank with the lowest anomaly score for prediction. These adaptations
allow the three memory- or statistics-based detectors to participate in the same
continual benchmark without collapsing their task-specific normality models into
a single mixed distribution. Additional notation and pseudocode for these
adaptations are provided in Appendix~\ref{app:continual-adaptations}.






\subsection{Results Discussion}

Table~\ref{tab:summary-avg-dataset-backbones} summarizes the benchmark at the
dataset level. The single-dataset rows average over all 9 anomaly-detection
models and the two backbones, whereas the MultiDataset rows average over the
three strongest methods selected from the single-dataset study: CFA, PaSTe, and
PaDiM. Across within-dataset streams, continual learning consistently improves
both detection quality and retention. At the image level, the average ROC-AUC
increases from 0.588 for NCL to 0.737 for CL, while BWT improves from $-0.088$
to $-0.027$. At the pixel level, the same pattern is observed: ROC-AUC increases
from 0.718 to 0.860 and BWT improves from $-0.074$ to $-0.010$. Thus, the
continual variants do not merely increase final detection performance; they also reduce forgetting.

\textbf{Effect of task ordering:} Task ordering affects image-level performance more strongly than pixel-level performance. Easy-to-Hard gives the highest average CL ROC-AUC at the image
level (0.764), followed by Random (0.738) and Hard-to-Easy (0.708). This
suggests that starting from easier concepts can stabilize image-level anomaly
scoring before harder concepts are introduced, whereas beginning with harder
concepts leads to weaker adaptation. Pixel-level CL ROC-AUC is more stable
across orderings, ranging from 0.849 to 0.870, indicating that localization
quality is less sensitive to the curriculum direction once replay or
method-specific continual mechanisms are used. BWT follows the same trend:
continual learning makes forgetting consistently less negative for all
orderings, with especially large gains under Easy-to-Hard.

\textbf{Adaptation versus retention:} At the aggregate level of Table~\ref{tab:summary-avg-dataset-backbones}, the CL variants provide the best adaptation--retention balance. They improve ROC-AUC on
every dataset and ordering while also moving BWT closer to zero, which indicates
less interference with previously learned tasks. The effect is most pronounced
on BTAD, where image-level BWT improves from $-0.289$ to $-0.046$ under
Easy-to-Hard and pixel-level BWT improves from $-0.247$ to $-0.010$. MPDD
remains challenging at the image level, with lower average CL ROC-AUC and more
negative BWT than most datasets, but its pixel-level results remain strong. This
gap suggests that some anomalies are easier to localize than to classify at the
image level. Since the single-dataset rows average over all 9 methods, they
support conclusions about the CL protocol as a class; method-specific rankings
require the per-method breakdowns rather than this pooled summary.

\textbf{Multi-dataset streams:} The MultiDataset rows in
Table~\ref{tab:summary-avg-dataset-backbones} show that cross-dataset streams are
substantially more challenging for naive image-level detection, with NCL ROC-AUC
near chance under both Easy-to-Hard and Hard-to-Easy orderings. Continual
learning recovers much of this gap, improving image-level ROC-AUC to roughly
$0.79$ and pixel-level ROC-AUC above $0.90$ while keeping BWT close to zero. This
suggests that the top-performing continual methods can retain useful normality
models even when task transitions cross dataset boundaries, although the
multi-dataset setting remains a stronger stress test than most within-dataset
streams. Appendix~\ref{app:transition-heatmaps} provides the corresponding
transition heatmaps and a finer-grained analysis of cross-dataset interference.

The same appendix also complements the aggregate table for single-dataset
streams. The heatmaps expose the per-task trajectory after each update, making it
possible to inspect when forgetting occurs, whether it is localized to specific
concepts, and how NCL and CL differ under the same ordering. They therefore
provide the fine-grained evidence needed to interpret results beyond
dataset-level averages. The key takeaway is that CL primarily improves
retained-task performance, especially in off-diagonal heatmap cells, and this
effect is more uniform for pixel-level localization. 



%

\section{Conclusion}

This paper presents a benchmark for continual industrial anomaly detection under
evolving production conditions. Rather than treating industrial inspection as a
static train/test problem, we cast it as a sequential setting in which models
must retain anomaly-detection ability on previously seen components while
adapting to new products, acquisition conditions, and dataset-level
shifts. Through a large-scale evaluation across multiple industrial datasets,
task orderings, anomaly-detection perspectives, and continual-learning
strategies, the benchmark exposes how performance depends not only on the
detector architecture, but also on the structure of the task stream. This
framing makes catastrophic forgetting, backward transfer, and cross-dataset
adaptation central evaluation targets.

\textit{Limitations}: Although the benchmark covers diverse datasets and ordering regimes, it remains limited by the scope of existing public industrial anomaly datasets, which only
partially reflect the temporal, operational, and sensor-level variability of
real production lines. The proposed tasks approximate continual
deployment by sequencing components and datasets, but they do not capture every
practical constraint, such as changing inspection hardware, evolving quality
standards, or long-term label drift. Future work should extend the benchmark
with richer temporal data, stronger cross-factory distribution shifts, and
standardized compute and memory budgets. 

\textit{Broader Impacts:} Continual industrial anomaly detection can have positive practical impact by
making visual inspection systems more adaptable to changing production
conditions. Better retention across components, products, and acquisition
settings may reduce unnecessary retraining, improve quality control, and support
earlier detection of manufacturing defects. In safety- or reliability-sensitive
industrial domains, more robust anomaly detection can also help reduce waste and
limit downstream failures caused by defective parts.
At the same time, benchmark progress should not be interpreted as a guarantee of
deployment readiness. Incorrect anomaly scores may lead to missed defects, false
alarms, unnecessary product rejection, or misplaced trust in automated
inspection. Continual systems can also accumulate errors over time if changes in
production are not monitored carefully. 
These risks motivate
human-in-the-loop validation, periodic auditing on recent production data, clear
reporting of failure cases, and deployment-specific checks before using such
models in operational decision-making.

\bibliographystyle{plainnat}
\bibliography{references}

\newpage
\appendix

\section{Benchmark Comparison}
\label{app:benchmark-comparison}

Table~\ref{tab:benchmark-comparison} positions InCLAD relative to representative
anomaly-detection and industrial-inspection benchmarks. The comparison emphasizes
that our contribution is not a new raw image corpus, but a continual-learning
benchmark layer built on top of public industrial datasets. InCLAD combines
normal-only training, image-level detection, pixel-level localization,
curriculum-style task orderings, a multi-dataset stream, and forgetting-aware
metrics within a single protocol.

\begin{table*}[h]
\centering
\caption{Comparison with representative anomaly-detection benchmarks. ``Partial''
indicates that the property is covered only for some datasets, tasks, or
evaluation settings.}
\label{tab:benchmark-comparison}
\scriptsize
\setlength{\tabcolsep}{2.6pt}
\renewcommand{\arraystretch}{0.95}
\resizebox{\textwidth}{!}{%
\begin{tabular}{lcccccccc}
\toprule
\textbf{Benchmark} &
\textbf{Industrial} &
\textbf{Continual} &
\textbf{Normal-only} &
\textbf{Image} &
\textbf{Pixel} &
\textbf{Curricula} &
\textbf{Multi-dataset} &
\textbf{BWT} \\
 & \textbf{domain} & \textbf{/ task-inc.} & \textbf{training} & \textbf{eval.} & \textbf{eval.} & \textbf{orderings} & \textbf{stream} & \textbf{metrics} \\
\midrule
MVTec AD~\citep{DBLP:conf/cvpr/BergmannFSS19} & Yes & No & Yes & Yes & Yes & No & No & No \\
VisA~\citep{DBLP:conf/eccv/ZouJPZD22} & Yes & No & Yes & Yes & Yes & No & No & No \\
ADBench~\citep{han2022adbench} & Partial & No & Partial & No & No & No & Yes & No \\
CableInspect-AD~\citep{arodi2024cableinspect} & Yes & No & Partial & Yes & Yes & No & No & No \\
ReinAD~\citep{wang2025reinad} & Yes & No & Yes & Yes & Yes & No & No & No \\
\textbf{InCLAD (ours)} & Yes & Yes & Yes & Yes & Yes & Yes & Yes & Yes \\
\bottomrule
\end{tabular}
}
\end{table*}

\section{Benchmark Metrics}
\label{app:benchmark-metrics}

Continual-learning metrics should characterize more than performance on the
most recent task. Following prior work on lifelong-learning evaluation
\cite{2016lopez_paz_gradient_episodic_memory,diazrodriguez2018dont},
we evaluate whether a method can learn incoming tasks, retain knowledge of
previously learned tasks, and transfer knowledge across the stream while
operating under the memory constraints imposed by the benchmark protocol.

Let $\mathcal{T}_1,\ldots,\mathcal{T}_N$ denote a task stream. During
evaluation, we construct a results matrix $R\in\mathbb{R}^{N\times N}$, where
$R_{i,j}$ is the ROC-AUC obtained on the test set of task $\mathcal{T}_j$ after
the learner has been trained up to task $\mathcal{T}_i$. Entries with $j>i$ are
not used because future tasks have not yet been observed. The same definition is
used independently for image-level and pixel-level evaluation: in the
image-level case, $R_{i,j}$ stores image-level ROC-AUC; in the pixel-level case,
it stores pixel-level ROC-AUC when segmentation masks are available.

\textbf{Continual ROC-AUC.} We report the lifetime average ROC-AUC over the
lower triangular part of $R$, including the diagonal. This summarizes
performance over the entire learning trajectory rather than only after the final
task:
\begin{equation}
    \mathrm{ROC\mbox{-}AUC}_{\mathrm{CL}}
    =
    \frac{2}{N(N+1)}
    \sum_{i=1}^{N}\sum_{j=1}^{i} R_{i,j}.
    \label{eq:continual-rocauc}
\end{equation}
For readability, the experimental tables refer to
$\mathrm{ROC\mbox{-}AUC}_{\mathrm{CL}}$ simply as ROC-AUC.

\textbf{Backward Transfer.} BWT measures the average effect of learning later
tasks on earlier tasks. It compares each post-update score $R_{i,j}$, with
$i>j$, against the score obtained immediately after learning task
$\mathcal{T}_j$, namely $R_{j,j}$:
\begin{equation}
    \mathrm{BWT}
    =
    \frac{2}{N(N-1)}
    \sum_{i=2}^{N}\sum_{j=1}^{i-1}\left(R_{i,j}-R_{j,j}\right).
    \label{eq:bwt}
\end{equation}
Positive BWT indicates that subsequent tasks improve earlier-task performance,
whereas negative BWT indicates forgetting. Strongly negative BWT corresponds to
severe forgetting and is especially undesirable in industrial inspection, where
a deployed system may revisit components learned earlier in the stream.

\section{Continual Adaptations for PatchCore, PaDiM, and CFA}
\label{app:continual-adaptations}

This appendix details the adaptations used to make PatchCore, PaDiM, and CFA
compatible with our continual anomaly-detection protocol. Let
$\mathcal{T}_1,\ldots,\mathcal{T}_K$ denote the task stream and let
$X_t=\{x_i^{(t)}\}_{i=1}^{n_t}$ be the normal training images for task
$\mathcal{T}_t$. At evaluation time, task identity is not used to select a
model-specific memory. Instead, each method scores an input against the
task-specific normality models accumulated so far and selects the one that best
explains the input as normal.

\subsection{Continual PatchCore}

Standard PatchCore extracts patch embeddings with a fixed feature extractor and
stores a coreset memory bank for nearest-neighbor anomaly scoring. A naive
continual implementation that simply merges all task banks would mix distinct
normality distributions and make old tasks dominate memory as the stream grows.
We therefore maintain separate task banks and use an explicit budget for
previous tasks.

Let $f$ be the frozen feature extractor and let
$Z_t=\{z_{i,p}^{(t)}\}$ be all patch embeddings extracted from $X_t$. For each
new task, we first compute the native PatchCore coreset
\[
    C_t = \operatorname{Coreset}(Z_t),
\]
using the same coreset ratio as the non-continual PatchCore baseline. The
current task is kept at this native resolution. Previous task banks
$C_1,\ldots,C_{t-1}$ are then compressed under a history budget
$B_t=\lfloor \rho |C_t| \rfloor$, where $\rho$ is the previous-task budget
ratio. Each previous task receives approximately
$b_t=\max(1,\lfloor B_t/(t-1)\rfloor)$ patches.

Compression is distance-weighted. For a patch $z$ in a previous bank $C_j$, we
compute its distance to the current task,
\[
    w_t(z)=\min_{c\in C_t}\|z-c\|_2 .
\]
Patches far from the current distribution are more task-specific and are kept
preferentially. The weighted greedy coreset selects patches by
\[
    z^\star = \arg\max_{z\in C_j}
    w_t(z)\min_{s\in S}\|z-s\|_2 ,
\]
where $S$ is the set of already selected patches for the compressed bank.

\paragraph{PatchCore CL update.}
\begin{enumerate}
    \item Extract patch embeddings $Z_t=f(X_t)$.
    \item Build the current native coreset $C_t=\operatorname{Coreset}(Z_t)$.
    \item For every previous bank $C_j$, compute $w_t(z)$ and shrink $C_j$ to
    $b_t$ patches using weighted greedy coreset selection.
    \item Store the list of task banks $\{C_1,\ldots,C_t\}$ and calibrate the
    threshold on $X_t$.
\end{enumerate}

For an input image $x$, let $Z(x)$ denote its patch embeddings. We compute
nearest-neighbor patch distances to every task bank $C_j$, aggregate them into a
task-level normality score, and select the closest bank:
\[
    j^\star(x)=\arg\min_j \frac{1}{|Z(x)|}
    \sum_{z\in Z(x)} d_{\mathrm{NN}}(z,C_j).
\]
The final image score and anomaly map are then computed using $C_{j^\star}$.

\subsection{Continual PaDiM}

PaDiM is not updated through gradient-based replay. It estimates a location-wise
Gaussian distribution over patch embeddings for normal images. To avoid mixing
heterogeneous components into one Gaussian model, we fit and store one PaDiM
distribution per task.

Let $z_p(x)$ be the selected PaDiM feature vector at spatial location $p$ for
image $x$. For task $\mathcal{T}_t$, we estimate
\[
    \mu_{t,p}=\frac{1}{n_t}\sum_{x\in X_t} z_p(x),
    \qquad
    \Sigma_{t,p}=\operatorname{Cov}\{z_p(x):x\in X_t\}+\epsilon I ,
\]
for every spatial location $p$, and store
$G_t=\{(\mu_{t,p},\Sigma_{t,p}^{-1})\}_p$.

\paragraph{PaDiM CL update.}
\begin{enumerate}
    \item Extract PaDiM feature maps for normal images in $X_t$.
    \item Estimate one location-aware Gaussian model $G_t$ for the current task.
    \item Append $G_t$ to the list of task distributions
    $\{G_1,\ldots,G_t\}$.
    \item Calibrate the anomaly threshold on $X_t$ using the task-selection rule.
\end{enumerate}

At inference time, each stored task distribution produces a Mahalanobis map:
\[
    m_{j,p}(x)=
    \sqrt{(z_p(x)-\mu_{j,p})^\top \Sigma_{j,p}^{-1}
    (z_p(x)-\mu_{j,p})}.
\]
Let $A_j(x)$ be the image-level score obtained by aggregating
$\{m_{j,p}(x)\}_p$. We select
\[
    j^\star(x)=\arg\min_j A_j(x),
\]
and report the image score and pixel-level map from $G_{j^\star}$.

\subsection{Continual CFA}

CFA differs from PatchCore and PaDiM because it contains a trainable descriptor
network and a memory bank used by the CFA soft-boundary objective. The memory
bank used for training must remain compatible with the descriptor being updated,
and old inference banks become stale after every descriptor update. Our
continual CFA therefore separates the training bank from the inference banks.

Let $g_\theta$ be the CFA descriptor. At task $t$, let $R_t$ denote replay
samples from previous tasks and define $U_t=X_t\cup R_t$. Before descriptor
training, we initialize a training bank
\[
    B_t^{\mathrm{train}}=\operatorname{Bank}(g_{\theta_{t-1}}(U_t)),
\]
which serves as the fixed memory target for the CFA soft-boundary loss. We then
update the descriptor parameters by minimizing the standard CFA training loss on
$U_t$:
\[
    \theta_t = \arg\min_\theta
    \mathcal{L}_{\mathrm{CFA}}(g_\theta(U_t),B_t^{\mathrm{train}}).
\]
After the descriptor update, all per-task inference banks are rebuilt using the
current descriptor so that every bank lies in the same embedding space:
\[
    B_j^{\mathrm{inf}}=\operatorname{Bank}(g_{\theta_t}(X_j)),
    \qquad j=1,\ldots,t .
\]

\paragraph{CFA CL update.}
\begin{enumerate}
    \item Store the current task samples $X_t$ for future bank reconstruction.
    \item Form $U_t=X_t\cup R_t$ from current data and replay samples.
    \item Initialize the training bank $B_t^{\mathrm{train}}$ from
    $g_{\theta_{t-1}}(U_t)$.
    \item Train the CFA descriptor on $U_t$ with the fixed training bank.
    \item Rebuild every per-task inference bank
    $B_1^{\mathrm{inf}},\ldots,B_t^{\mathrm{inf}}$ using the updated descriptor
    $g_{\theta_t}$.
    \item Calibrate the threshold on $X_t$ and update the replay buffer.
\end{enumerate}

For prediction, CFA scores the image against each inference bank and selects the
bank that yields the lowest image-level anomaly score:
\[
    j^\star(x)=\arg\min_j A(x;B_j^{\mathrm{inf}},g_{\theta_t}).
\]
The final score and anomaly map are computed using
$B_{j^\star}^{\mathrm{inf}}$. This avoids comparing embeddings generated by
different descriptor states and prevents old banks from becoming incompatible
with the current CFA model.

{

\section{Reproducibility Details}
\label{app:reproducibility}

\subsection{Dataset Licenses and Scenario Release}
\label{app:dataset-licenses}

The continual scenarios are built on existing public datasets, and their release
must respect the corresponding dataset licenses. MVTec AD is released under
CC BY-NC-SA 4.0, VisA under CC BY 4.0, BTAD under CC BY-SA 4.0, MPDD under
CC BY-NC-SA 4.0, and DAGM under CC BY 4.0. Because some of these licenses are
non-commercial and share-alike, we do not redistribute the raw images or masks in
our repository. Instead, we release code, configuration files, and scenario
manifests containing dataset names, task orderings, relative file paths, split
definitions, and random seeds. Practitioners are expected to obtain the raw
datasets from the official sources, place them in a local directory, and run our
scripts to materialize the continual scenarios. This keeps the benchmark
reproducible while avoiding repackaging datasets under incompatible or more
permissive repository terms.

\subsection{Computational Resources}
\label{app:compute-resources}

Experiments were run on an institutional Red Hat Enterprise Linux HPC system and
on two local high-performance workstations. The HPC system includes one 2~TB
big-memory node, one node equipped with two NVIDIA H100 GPUs with 80~GB of GPU
memory each, and six nodes equipped with four NVIDIA L40 GPUs with 40~GB of GPU
memory each. Each compute node has 512~GB of RAM and dual-socket Intel Xeon
Platinum 8462Y+ CPUs with 32 cores per socket.

\subsection{Empirical Time-Complexity}
\label{app:empirical-time-complexity}

Empirical time complexity is measured through wall-clock training and evaluation
time for each method, dataset, task ordering, and anomaly-detection level.
Tables~\ref{tab:execution_times_wide-resnet50-2}
and~\ref{tab:execution_times_special} report training and evaluation times
separately, averaged across available orderings. The measurements show that
runtime is driven by both the dataset scale and the model family. Feature-bank
methods such as PatchCore and PaDiM can have modest update costs but expensive
evaluation, especially on larger streams such as VisA and DAGM, because test
images must be compared against accumulated task-specific statistics or memory
banks. Gradient-trained methods such as FastFlow, RD4AD, STFPM, PaSTe, and CFA
instead concentrate more cost in training. Continual variants usually increase
runtime relative to the naive setting because they include replay, accumulated
task models, or repeated evaluation over previously seen tasks. The effect is
most visible for evaluation-heavy methods on long streams, where retaining
separate task-specific normality models improves continual behavior but also
increases the cost of scoring.

\begin{table*}[p]
\centering
\caption{Execution times (train and eval separately, mean $\pm$ population std across available orderings) for WideResNet-50-2-based models. Continual denotes CL for CFA, PaDiM, and PatchCore, and replay for the remaining models. When image and pixel results were produced as separate runs, their times are summed to reflect the total wall-clock cost of obtaining both modalities.}
\label{tab:execution_times_wide-resnet50-2}
\scriptsize
\setlength{\tabcolsep}{2pt}
\renewcommand{\arraystretch}{0.82}
\begin{tabular}{lllrr}
\toprule
\textbf{Dataset} & \textbf{Model} & \textbf{Time} & \textbf{Naive} & \textbf{Continual} \\
\midrule
\multirow{14}{*}{MVTec} & \multirow{2}{*}{CFA} & Train & 6m\,32s ($\pm$ 1m\,42s) & 15m\,42s ($\pm$ 10m\,57s) \\
 &  & Eval & 9m\,35s ($\pm$ 1m\,40s) & 52m\,7s ($\pm$ 31m\,50s) \\
 & \multirow{2}{*}{PaDiM} & Train & 6m\,19s ($\pm$ 1m\,39s) & 7m\,9s ($\pm$ 1m\,18s) \\
 &  & Eval & 30m\,28s ($\pm$ 3m\,22s) & 65m\,24s ($\pm$ 3m\,37s) \\
 & \multirow{2}{*}{PatchCore} & Train & 3m\,26s ($\pm$ 12.6s) & 4m\,17s ($\pm$ 10.2s) \\
 &  & Eval & 34m\,27s ($\pm$ 3m\,58s) & 49m\,49s ($\pm$ 1m\,42s) \\
 & \multirow{2}{*}{FastFlow} & Train & 112m\,29s ($\pm$ 5m\,0s) & 122m\,35s ($\pm$ 1m\,59s) \\
 &  & Eval & 16m\,51s ($\pm$ 7.9s) & 16m\,12s ($\pm$ 1m\,21s) \\
 & \multirow{2}{*}{PaSTe} & Train & 19m\,33s ($\pm$ 1m\,30s) & 27m\,13s ($\pm$ 7m\,40s) \\
 &  & Eval & 9m\,27s ($\pm$ 10.9s) & 11m\,4s ($\pm$ 2m\,13s) \\
 & \multirow{2}{*}{RD4AD} & Train & 57m\,11s ($\pm$ 6.6s) & 79m\,58s ($\pm$ 22m\,31s) \\
 &  & Eval & 10m\,46s ($\pm$ 25.5s) & 11m\,47s ($\pm$ 2m\,27s) \\
 & \multirow{2}{*}{STFPM} & Train & 29m\,9s ($\pm$ 6m\,1s) & 26m\,3s ($\pm$ 7m\,30s) \\
 &  & Eval & 14m\,5s ($\pm$ 2m\,52s) & 11m\,53s ($\pm$ 1m\,30s) \\
\midrule
\multirow{14}{*}{VisA} & \multirow{2}{*}{CFA} & Train & 39m\,22s ($\pm$ 5m\,10s) & 64m\,40s ($\pm$ 13m\,51s) \\
 &  & Eval & 19m\,36s ($\pm$ 2m\,35s) & 102m\,38s ($\pm$ 24m\,20s) \\
 & \multirow{2}{*}{PaDiM} & Train & 11m\,11s ($\pm$ 33.2s) & 20m\,20s ($\pm$ 5m\,14s) \\
 &  & Eval & 51m\,32s ($\pm$ 17m\,21s) & 105m\,0s ($\pm$ 28m\,37s) \\
 & \multirow{2}{*}{PatchCore} & Train & 29m\,40s ($\pm$ 4m\,1s) & 32m\,23s ($\pm$ 2m\,55s) \\
 &  & Eval & 129m\,17s ($\pm$ 19m\,19s) & 181m\,7s ($\pm$ 22m\,33s) \\
 & \multirow{2}{*}{FastFlow} & Train & 283m\,25s ($\pm$ 31m\,56s) & 346m\,6s ($\pm$ 52m\,30s) \\
 &  & Eval & 22m\,0s ($\pm$ 2m\,55s) & 20m\,36s ($\pm$ 56.7s) \\
 & \multirow{2}{*}{PaSTe} & Train & 59m\,20s ($\pm$ 24m\,4s) & 120m\,40s ($\pm$ 29m\,47s) \\
 &  & Eval & 12m\,41s ($\pm$ 4m\,19s) & 21m\,29s ($\pm$ 3m\,16s) \\
 & \multirow{2}{*}{RD4AD} & Train & 253m\,25s ($\pm$ 89m\,11s) & 274m\,39s ($\pm$ 146m\,15s) \\
 &  & Eval & 19m\,57s ($\pm$ 6m\,57s) & 18m\,17s ($\pm$ 8m\,9s) \\
 & \multirow{2}{*}{STFPM} & Train & 79m\,47s ($\pm$ 20m\,39s) & 105m\,24s ($\pm$ 31m\,44s) \\
 &  & Eval & 16m\,32s ($\pm$ 5m\,9s) & 18m\,18s ($\pm$ 4m\,22s) \\
\midrule
\multirow{14}{*}{BTAD} & \multirow{2}{*}{CFA} & Train & 2m\,57s ($\pm$ 3m\,22s) & 3m\,31s ($\pm$ 4m\,5s) \\
 &  & Eval & 1m\,11s ($\pm$ 9.9s) & 1m\,30s ($\pm$ 36.1s) \\
 & \multirow{2}{*}{PaDiM} & Train & 8m\,11s ($\pm$ 3m\,56s) & 10m\,38s ($\pm$ 3m\,13s) \\
 &  & Eval & 10m\,10s ($\pm$ 1m\,3s) & 14m\,17s ($\pm$ 56.6s) \\
 & \multirow{2}{*}{PatchCore} & Train & 18m\,42s ($\pm$ 55.7s) & 20m\,27s ($\pm$ 15.7s) \\
 &  & Eval & 21m\,32s ($\pm$ 2m\,23s) & 26m\,28s ($\pm$ 1m\,1s) \\
 & \multirow{2}{*}{FastFlow} & Train & 55m\,17s ($\pm$ 7m\,55s) & 62m\,46s ($\pm$ 8m\,42s) \\
 &  & Eval & 1m\,37s ($\pm$ 23.7s) & 1m\,36s ($\pm$ 20.8s) \\
 & \multirow{2}{*}{PaSTe} & Train & 10m\,39s ($\pm$ 3m\,14s) & 12m\,5s ($\pm$ 3m\,38s) \\
 &  & Eval & 1m\,3s ($\pm$ 22.1s) & 1m\,2s ($\pm$ 22.0s) \\
 & \multirow{2}{*}{RD4AD} & Train & 30m\,22s ($\pm$ 9m\,6s) & 34m\,28s ($\pm$ 10m\,37s) \\
 &  & Eval & 1m\,5s ($\pm$ 24.0s) & 1m\,5s ($\pm$ 24.4s) \\
 & \multirow{2}{*}{STFPM} & Train & 11m\,56s ($\pm$ 3.7s) & 13m\,31s ($\pm$ 3.1s) \\
 &  & Eval & 58.7s ($\pm$ 1.2s) & 58.4s ($\pm$ 1.2s) \\
\midrule
\multirow{14}{*}{MPDD} & \multirow{2}{*}{CFA} & Train & 3m\,3s ($\pm$ 10.8s) & 3m\,37s ($\pm$ 17.2s) \\
 &  & Eval & 1m\,33s ($\pm$ 5.0s) & 2m\,54s ($\pm$ 25.6s) \\
 & \multirow{2}{*}{PaDiM} & Train & 10m\,5s ($\pm$ 6m\,55s) & 9m\,55s ($\pm$ 1m\,6s) \\
 &  & Eval & 11m\,35s ($\pm$ 42.5s) & 30m\,12s ($\pm$ 2m\,47s) \\
 & \multirow{2}{*}{PatchCore} & Train & 2m\,38s ($\pm$ 5.5s) & 3m\,54s ($\pm$ 1m\,5s) \\
 &  & Eval & 11m\,15s ($\pm$ 2m\,19s) & 14m\,2s ($\pm$ 2m\,42s) \\
 & \multirow{2}{*}{FastFlow} & Train & 21m\,36s ($\pm$ 19.8s) & 25m\,15s ($\pm$ 37.3s) \\
 &  & Eval & 1m\,59s ($\pm$ 7.7s) & 2m\,1s ($\pm$ 8.2s) \\
 & \multirow{2}{*}{PaSTe} & Train & 6m\,34s ($\pm$ 2.1s) & 7m\,38s ($\pm$ 5.9s) \\
 &  & Eval & 1m\,30s ($\pm$ 2.3s) & 1m\,31s ($\pm$ 2.3s) \\
 & \multirow{2}{*}{RD4AD} & Train & 16m\,24s ($\pm$ 10.5s) & 18m\,51s ($\pm$ 12.4s) \\
 &  & Eval & 1m\,32s ($\pm$ 2.1s) & 1m\,32s ($\pm$ 2.5s) \\
 & \multirow{2}{*}{STFPM} & Train & 6m\,49s ($\pm$ 9.0s) & 7m\,59s ($\pm$ 4.8s) \\
 &  & Eval & 1m\,54s ($\pm$ 2.6s) & 1m\,53s ($\pm$ 2.6s) \\
\midrule
\multirow{14}{*}{DAGM} & \multirow{2}{*}{CFA} & Train & 20m\,39s ($\pm$ 17m\,42s) & 30m\,44s ($\pm$ 17m\,39s) \\
 &  & Eval & 33m\,38s ($\pm$ 18m\,9s) & 114m\,58s ($\pm$ 61m\,21s) \\
 & \multirow{2}{*}{PaDiM} & Train & 4m\,42s ($\pm$ 2m\,47s) & 10m\,55s ($\pm$ 7m\,41s) \\
 &  & Eval & 69m\,55s ($\pm$ 15m\,47s) & 133m\,15s ($\pm$ 35m\,49s) \\
 & \multirow{2}{*}{PatchCore} & Train & 18m\,33s ($\pm$ 6m\,6s) & 20m\,40s ($\pm$ 6m\,12s) \\
 &  & Eval & 185m\,22s ($\pm$ 67m\,21s) & 224m\,32s ($\pm$ 69m\,37s) \\
 & \multirow{2}{*}{FastFlow} & Train & 207m\,1s ($\pm$ 15m\,13s) & 283m\,59s ($\pm$ 92m\,53s) \\
 &  & Eval & 44m\,14s ($\pm$ 1m\,32s) & 49m\,52s ($\pm$ 14m\,32s) \\
 & \multirow{2}{*}{PaSTe} & Train & 34m\,2s ($\pm$ 1m\,22s) & 41m\,12s ($\pm$ 4m\,40s) \\
 &  & Eval & 24m\,32s ($\pm$ 48.2s) & 24m\,43s ($\pm$ 1m\,53s) \\
 & \multirow{2}{*}{RD4AD} & Train & 110m\,2s ($\pm$ 16m\,3s) & 112m\,44s ($\pm$ 6m\,31s) \\
 &  & Eval & 26m\,28s ($\pm$ 2m\,8s) & 25m\,49s ($\pm$ 1m\,54s) \\
 & \multirow{2}{*}{STFPM} & Train & 50m\,34s ($\pm$ 4m\,21s) & 59m\,2s ($\pm$ 4m\,14s) \\
 &  & Eval & 33m\,49s ($\pm$ 2m\,52s) & 32m\,46s ($\pm$ 1m\,59s) \\
\bottomrule
\end{tabular}
\end{table*}

\begin{table*}[p]
\centering
\caption{Execution times (train and eval separately, mean $\pm$ population std across available orderings) for models without the two-backbone setup: GANomaly (default) and EfficientAD (small). When image and pixel results were produced as separate runs, their times are summed to reflect the total wall-clock cost of obtaining both modalities.}
\label{tab:execution_times_special}
\scriptsize
\setlength{\tabcolsep}{3pt}
\renewcommand{\arraystretch}{0.85}
\begin{tabular}{lllrr}
\toprule
\textbf{Dataset} & \textbf{Model} & \textbf{Time} & \textbf{Naive} & \textbf{Continual} \\
\midrule
\multirow{4}{*}{MVTec} & \multirow{2}{*}{GANomaly (default)} & Train & 4m\,59s ($\pm$ 0.7s) & 5m\,56s ($\pm$ 6.1s) \\
 &  & Eval & 9m\,38s ($\pm$ 8.4s) & 9m\,44s ($\pm$ 7.1s) \\
 & \multirow{2}{*}{EfficientAD (small)} & Train & -- & 80m\,15s ($\pm$ 1m\,6s) \\
 &  & Eval & -- & 11m\,55s ($\pm$ 1.4s) \\
\midrule
\multirow{4}{*}{VisA} & \multirow{2}{*}{GANomaly (default)} & Train & 11m\,32s ($\pm$ 10.5s) & 13m\,23s ($\pm$ 18.4s) \\
 &  & Eval & 10m\,44s ($\pm$ 35.1s) & 10m\,35s ($\pm$ 41.3s) \\
 & \multirow{2}{*}{EfficientAD (small)} & Train & 202m\,50s ($\pm$ 50m\,42s) & 293m\,14s ($\pm$ 84m\,23s) \\
 &  & Eval & 15m\,57s ($\pm$ 2m\,41s) & 22m\,37s ($\pm$ 8m\,48s) \\
\midrule
\multirow{4}{*}{BTAD} & \multirow{2}{*}{GANomaly (default)} & Train & 3m\,7s ($\pm$ 14.7s) & 3m\,32s ($\pm$ 17.1s) \\
 &  & Eval & 1m\,28s ($\pm$ 2.6s) & 1m\,30s ($\pm$ 3.8s) \\
 & \multirow{2}{*}{EfficientAD (small)} & Train & 24m\,57s ($\pm$ 8.2s) & 28m\,18s ($\pm$ 11.0s) \\
 &  & Eval & 1m\,35s ($\pm$ 5.4s) & 1m\,36s ($\pm$ 5.0s) \\
\midrule
\multirow{4}{*}{MPDD} & \multirow{2}{*}{GANomaly (default)} & Train & 1m\,14s ($\pm$ 5.5s) & 1m\,23s ($\pm$ 6.4s) \\
 &  & Eval & 1m\,39s ($\pm$ 5.8s) & 1m\,36s ($\pm$ 2.7s) \\
 & \multirow{2}{*}{EfficientAD (small)} & Train & 7m\,42s ($\pm$ 4.7s) & 8m\,55s ($\pm$ 9.6s) \\
 &  & Eval & 1m\,43s ($\pm$ 3.3s) & 1m\,44s ($\pm$ 5.0s) \\
\midrule
\multirow{2}{*}{DAGM} & \multirow{2}{*}{GANomaly (default)} & Train & 10m\,50s ($\pm$ 1m\,25s) & 12m\,52s ($\pm$ 1m\,41s) \\
 &  & Eval & 41m\,40s ($\pm$ 9m\,42s) & 45m\,0s ($\pm$ 11m\,48s) \\
\bottomrule
\end{tabular}
\end{table*}

\subsection{Hyperparameter Configurations}
\label{app:hyperparameter-configurations}

Because runtime depends on training schedules, backbones, batch sizes, and
memory-bank construction, we report the effective hyperparameters used in the
visual anomaly-detection experiments. Tables~\ref{tab:hyperparameters_visual_cl}
and~\ref{tab:hyperparameters_visual_noncl} summarize the CL-specific and non-CL
settings, respectively. The CL table covers the adapted PatchCore, PaDiM, and
CFA variants: CFA is trained for multiple epochs, while PaDiM and PatchCore use
fixed pretrained features followed by task-specific statistics or memory-bank
updates. These settings provide the context needed to interpret the timing
results in Appendix~\ref{app:empirical-time-complexity}.

\begin{table}[t]
  \centering
  \small
  \caption{Effective hyperparameters of the CL-specific visual anomaly detection models used in our experiments. The runner passes \texttt{epochs=50} to all CL jobs, but this setting is effective only for CFA.}
  \label{tab:hyperparameters_visual_cl}
  \resizebox{\linewidth}{!}{%
  \begin{tabular}{lccc}
  \toprule
  \textbf{Hyperparameter}
    & \textbf{CFA}
    & \textbf{PaDiM}
    & \textbf{PatchCore} \\
  \midrule
  \multicolumn{4}{l}{\textit{Training}} \\
  \midrule
  Backbone          & resnet18 / wide\_resnet50\_2 & resnet18 / wide\_resnet50\_2 & resnet18 / wide\_resnet50\_2 \\
  Pretrained        & \checkmark & \checkmark & \checkmark \\
  Batch size        & 32 & 32 & 32 \\
  Epochs            & 50 & --- & --- \\
  Learning rate     & $10^{-3}$ & --- & --- \\
  Weight decay      & $5 \cdot 10^{-4}$ & --- & --- \\
  \midrule
  \multicolumn{4}{l}{\textit{Architecture / scoring}} \\
  \midrule
  Input size         & $(224,224)$ & $(224,224)$ & $(224,224)$ \\
  Threshold quantile & 0.99 & 0.99 & 0.99 \\
  Freeze backbone    & \checkmark & \checkmark & \checkmark \\
  \midrule
  \multicolumn{4}{l}{\textit{Model-specific}} \\
  \midrule
  $k$-neighbors              & 3 & --- & --- \\
  Repulsion neighbors        & 3 & --- & --- \\
  $n_{\text{features}}$      & --- & 100 & --- \\
  Covariance regularization  & --- & 0.01 & --- \\
  Coreset sampling ratio     & --- & --- & 0.1 \\
  Coreset projection dim.    & --- & --- & 128 \\
  Coreset starting points    & --- & --- & 10 \\
  Neighbors at inference     & --- & --- & 1 \\
  Previous-ratio (CL)        & --- & --- & 0.25 \\
  \bottomrule
  \end{tabular}}
\end{table}

\begin{table*}[t]
  \centering
  \small
  \caption{Effective hyperparameters of the non-CL visual anomaly detection models used in our experiments. When two backbones are listed, separate runs were executed for both settings.}
  \label{tab:hyperparameters_visual_noncl}
  \resizebox{\linewidth}{!}{%
  \begin{tabular}{lcccccc}
  \toprule
  \textbf{Hyperparameter}
    & \textbf{FastFlow}
    & \textbf{GANomaly}
    & \textbf{PaSTe}
    & \textbf{RD4AD}
    & \textbf{STFPM}
    & \textbf{EfficientAD} \\
  \midrule
  \multicolumn{7}{l}{\textit{Training}} \\
  \midrule
  Backbone          & resnet18 / wide\_resnet50\_2 & --- & resnet18 / wide\_resnet50\_2 & resnet18 / wide\_resnet50\_2 & resnet18 / wide\_resnet50\_2 & --- \\
  Pretrained        & \checkmark & --- & \checkmark & \checkmark & \checkmark & --- \\
  Batch size        & 8   & 16 & 32 & 16 & 32 & 1 \\
  Epochs            & 200 & 15 & 100 & 200 & 100 & 50 \\
  Learning rate     & $10^{-3}$ & $2 \cdot 10^{-4}$ & 0.4 & $5 \cdot 10^{-3}$ & 0.4 & $10^{-4}$ \\
  Weight decay      & 0 & 0 & $10^{-4}$ & 0 & $10^{-4}$ & $10^{-5}$ \\
  \midrule
  \multicolumn{7}{l}{\textit{Architecture / scoring}} \\
  \midrule
  Input size        & $(256,256)$ & $(256,256)$ & $(224,224)$ & $(224,224)$ & $(256,256)$ & $(256,256)$ \\
  Threshold quantile & 0.99 & 0.99 & 0.99 & 0.99 & 0.99 & 0.99 \\
  Score mode        & max & --- & max & max & max & max \\
  \midrule
  \multicolumn{7}{l}{\textit{Model-specific}} \\
  \midrule
  Flow steps              & 8 & --- & --- & --- & --- & --- \\
  Hidden ratio            & 1.0 & --- & --- & --- & --- & --- \\
  Feature width $n_{\text{features}}$ & --- & 64 & --- & --- & --- & --- \\
  Latent vector size      & --- & 100 & --- & --- & --- & --- \\
  Student bootstrap layer & --- & --- & 0 & --- & --- & --- \\
  Score smoothing kernel  & --- & --- & --- & 1 & --- & --- \\
  Model size              & --- & --- & --- & --- & --- & small \\
  Teacher out channels    & --- & --- & --- & --- & --- & 384 \\
  \bottomrule
  \end{tabular}}
\end{table*}
}

\newpage

\section{Scenario Ordering Details}
\label{app:ordering-details}

This appendix reports the exact concept orderings used to derive the structured
continual scenarios in the first stage of the study (see Tables
~\ref{tab:ordering-spearman}--\ref{tab:ordering-visa}). 
%
The ordering tables serve two purposes. First, they make the scenario
construction reproducible by listing the hard-to-easy sequence used for each
dataset; the easy-to-hard curriculum is obtained by reversing the same sequence.
Second, they show how stable the notion of task difficulty is across detector
families. Table~\ref{tab:ordering-spearman} reports the pairwise Spearman
correlations among PatchCore, PaDiM, and CFA. High correlations indicate that the
three Single-Task Experts agree on which concepts are easier or harder, while low
or negative correlations indicate that difficulty is partly method-dependent.
This is why the benchmark uses average rank rather than the score of a single
detector to define curriculum orderings.

The agreement patterns differ substantially across datasets. BTAD has only three
concepts, but the models consistently identify product \texttt{02} as the
hardest and product \texttt{03} as the easiest. VisA also shows strong agreement
across all detector pairs, with capsules and macaroni variants near the hard end
and chewinggum or \texttt{pcb4} near the easy end. DAGM has generally high
agreement between PatchCore and PaDiM, but weaker agreement involving CFA,
showing that some synthetic texture classes favor different modeling biases.
MVTec AD occupies an intermediate regime: the easiest concepts such as bottle,
leather, and wood are consistently strong, while concepts such as pill, screw,
and capsule form the hard end of the stream. MPDD is the clearest case where
ordering cannot be inferred reliably from one detector alone. PaDiM and CFA are
negatively correlated on MPDD, and concepts such as \texttt{metal\_plate} are
easy for CFA but difficult for PaDiM. The average-rank ordering therefore
prevents the curriculum from being dominated by one method's failure mode.

These differences matter for interpreting continual-learning results. Datasets
with strong agreement, such as VisA and BTAD, provide curricula where task
difficulty is relatively stable across model families. Datasets with weaker
agreement, especially MPDD and parts of MVTec AD, create more ambiguous streams:
a task may be easy for memory-based retrieval but difficult for Gaussian
statistics, or vice versa. Such streams are useful stress tests because
forgetting and backward transfer are then affected not only by nominal task
order, but also by how well a continual method's inductive bias matches each
concept.

\begin{table*}[t]
\centering
\small
\caption{Pairwise Spearman rank correlations between the three Single-Task
Expert models for each dataset. Higher values indicate stronger agreement on
concept difficulty, while low or negative values indicate that the relative
difficulty of concepts depends more strongly on the chosen detector.}
\label{tab:ordering-spearman}
\begin{tabular}{lccc}
\toprule
Dataset & PatchCore vs PaDiM & PatchCore vs CFA & PaDiM vs CFA \\
\midrule
BTAD  & 0.500 & 1.000 & 0.500 \\
DAGM  & 0.855 & 0.455 & 0.648 \\
MPDD  & -0.029 & 0.200 & -0.657 \\
MVTec AD & 0.382 & 0.693 & 0.654 \\
VisA  & 0.622 & 0.797 & 0.748 \\
\bottomrule
\end{tabular}
\end{table*}

\begin{table*}[t]
\centering
\scriptsize
\setlength{\tabcolsep}{5pt}
\caption{BTAD hard-to-easy concept ordering derived from average rank across
PatchCore, PaDiM, and CFA. The easy-to-hard scenario is the reverse of this
table.}
\label{tab:ordering-btad}
\resizebox{\textwidth}{!}{%
\begin{tabular}{ccccccc}
\toprule
Rank & Concept & Avg. ROC-AUC & Avg. Rank & PatchCore & PaDiM & CFA \\
\midrule
1 & 02 & 0.7947 & 1.00 & 0.7923 & 0.7955 & 0.7963 \\
2 & 01 & 0.9679 & 2.33 & 0.9456 & 0.9874 & 0.9708 \\
3 & 03 & 0.9903 & 2.67 & 0.9999 & 0.9819 & 0.9890 \\
\bottomrule
\end{tabular}}
\end{table*}

\begin{table*}[t]
\centering
\scriptsize
\setlength{\tabcolsep}{5pt}
\caption{DAGM hard-to-easy concept ordering derived from average rank across
PatchCore, PaDiM, and CFA.}
\label{tab:ordering-dagm}
\resizebox{\textwidth}{!}{%
\begin{tabular}{ccccccc}
\toprule
Rank & Concept & Avg. ROC-AUC & Avg. Rank & PatchCore & PaDiM & CFA \\
\midrule
1 & Class8  & 0.6541 & 1.00 & 0.6729 & 0.7278 & 0.5616 \\
2 & Class5  & 0.8210 & 3.00 & 0.8164 & 0.8160 & 0.8308 \\
3 & Class2  & 0.8594 & 3.67 & 0.8920 & 0.9027 & 0.7835 \\
4 & Class1  & 0.8380 & 3.67 & 0.8561 & 0.7395 & 0.9183 \\
5 & Class3  & 0.8625 & 4.67 & 0.9776 & 0.9274 & 0.6825 \\
6 & Class7  & 0.9565 & 7.33 & 0.9990 & 0.9310 & 0.9394 \\
7 & Class10 & 0.9587 & 7.33 & 0.9946 & 0.9960 & 0.8854 \\
8 & Class6  & 0.9660 & 7.33 & 0.9519 & 0.9489 & 0.9972 \\
9 & Class9  & 0.9802 & 7.67 & 0.9630 & 0.9863 & 0.9912 \\
10 & Class4 & 0.9857 & 9.33 & 1.0000 & 1.0000 & 0.9571 \\
\bottomrule
\end{tabular}}
\end{table*}

\begin{table*}[t]
\centering
\scriptsize
\setlength{\tabcolsep}{5pt}
\caption{MPDD hard-to-easy concept ordering derived from average rank across
PatchCore, PaDiM, and CFA.}
\label{tab:ordering-mpdd}
\resizebox{\textwidth}{!}{%
\begin{tabular}{ccccccc}
\toprule
Rank & Concept & Avg. ROC-AUC & Avg. Rank & PatchCore & PaDiM & CFA \\
\midrule
1 & bracket\_black & 0.6791 & 2.33 & 0.8471 & 0.6410 & 0.5492 \\
2 & tubes          & 0.7435 & 3.00 & 0.8170 & 0.5738 & 0.8397 \\
3 & bracket\_white & 0.7537 & 3.00 & 0.7311 & 0.8533 & 0.6767 \\
4 & metal\_plate   & 0.8106 & 3.67 & 0.9420 & 0.4902 & 0.9995 \\
5 & bracket\_brown & 0.8286 & 4.33 & 0.9480 & 0.7624 & 0.7753 \\
6 & connector      & 0.8643 & 4.67 & 1.0000 & 0.8214 & 0.7714 \\
\bottomrule
\end{tabular}}
\end{table*}

\begin{table*}[t]
\centering
\scriptsize
\setlength{\tabcolsep}{5pt}
\caption{MVTec AD hard-to-easy concept ordering derived from average rank
across PatchCore, PaDiM, and CFA.}
\label{tab:ordering-mvtec}
\resizebox{\textwidth}{!}{%
\begin{tabular}{ccccccc}
\toprule
Rank & Concept & Avg. ROC-AUC & Avg. Rank & PatchCore & PaDiM & CFA \\
\midrule
1 & pill        & 0.8300 & 2.00 & 0.8161 & 0.7700 & 0.9037 \\
2 & screw       & 0.7930 & 3.00 & 0.9664 & 0.7284 & 0.6842 \\
3 & capsule     & 0.8738 & 3.67 & 0.9334 & 0.8337 & 0.8544 \\
4 & grid        & 0.9176 & 4.67 & 0.8956 & 0.9181 & 0.9390 \\
5 & zipper      & 0.9079 & 5.33 & 0.9362 & 0.8277 & 0.9598 \\
6 & transistor  & 0.9193 & 7.00 & 0.9729 & 0.8129 & 0.9721 \\
7 & carpet      & 0.9522 & 7.33 & 0.9575 & 0.9591 & 0.9402 \\
8 & toothbrush  & 0.9583 & 8.00 & 0.9722 & 0.9528 & 0.9500 \\
9 & cable       & 0.9333 & 9.00 & 0.9751 & 0.8373 & 0.9874 \\
10 & metal\_nut & 0.9658 & 9.67 & 0.9780 & 0.9428 & 0.9765 \\
11 & hazelnut   & 0.9231 & 9.67 & 1.0000 & 0.7754 & 0.9939 \\
12 & tile       & 0.9698 & 10.33 & 0.9470 & 0.9625 & 1.0000 \\
13 & wood       & 0.9956 & 13.00 & 0.9982 & 0.9895 & 0.9991 \\
14 & leather    & 0.9890 & 13.33 & 0.9888 & 0.9783 & 1.0000 \\
15 & bottle     & 0.9976 & 14.00 & 1.0000 & 0.9929 & 1.0000 \\
\bottomrule
\end{tabular}}
\end{table*}

\begin{table*}[t]
\centering
\scriptsize
\setlength{\tabcolsep}{5pt}
\caption{VisA hard-to-easy concept ordering derived from average rank across
PatchCore, PaDiM, and CFA.}
\label{tab:ordering-visa}
\resizebox{\textwidth}{!}{%
\begin{tabular}{ccccccc}
\toprule
Rank & Concept & Avg. ROC-AUC & Avg. Rank & PatchCore & PaDiM & CFA \\
\midrule
1 & capsules   & 0.6493 & 1.33 & 0.7465 & 0.6175 & 0.5840 \\
2 & macaroni2  & 0.6407 & 2.00 & 0.6722 & 0.6567 & 0.5931 \\
3 & macaroni1  & 0.8148 & 4.00 & 0.8965 & 0.7791 & 0.7687 \\
4 & pcb3       & 0.7765 & 4.33 & 0.9196 & 0.6352 & 0.7746 \\
5 & pcb2       & 0.8363 & 6.00 & 0.9055 & 0.8094 & 0.7941 \\
6 & candle     & 0.8623 & 6.00 & 0.8770 & 0.7839 & 0.9259 \\
7 & fryum      & 0.8574 & 6.33 & 0.8968 & 0.8282 & 0.8472 \\
8 & cashew     & 0.9023 & 8.33 & 0.9420 & 0.9168 & 0.8482 \\
9 & pipe\_fryum & 0.9063 & 8.33 & 0.9882 & 0.7768 & 0.9540 \\
10 & pcb1      & 0.9082 & 8.67 & 0.9626 & 0.8681 & 0.8940 \\
11 & chewinggum & 0.9788 & 11.33 & 0.9764 & 0.9668 & 0.9932 \\
12 & pcb4      & 0.9680 & 11.33 & 0.9892 & 0.9376 & 0.9770 \\
\bottomrule
\end{tabular}}
\end{table*}

\newpage
\section{Fine-grained Heatmap Results}
\label{app:transition-heatmaps}

\begin{figure*}[!ht]
  \centering
  \includegraphics[width=\textwidth]{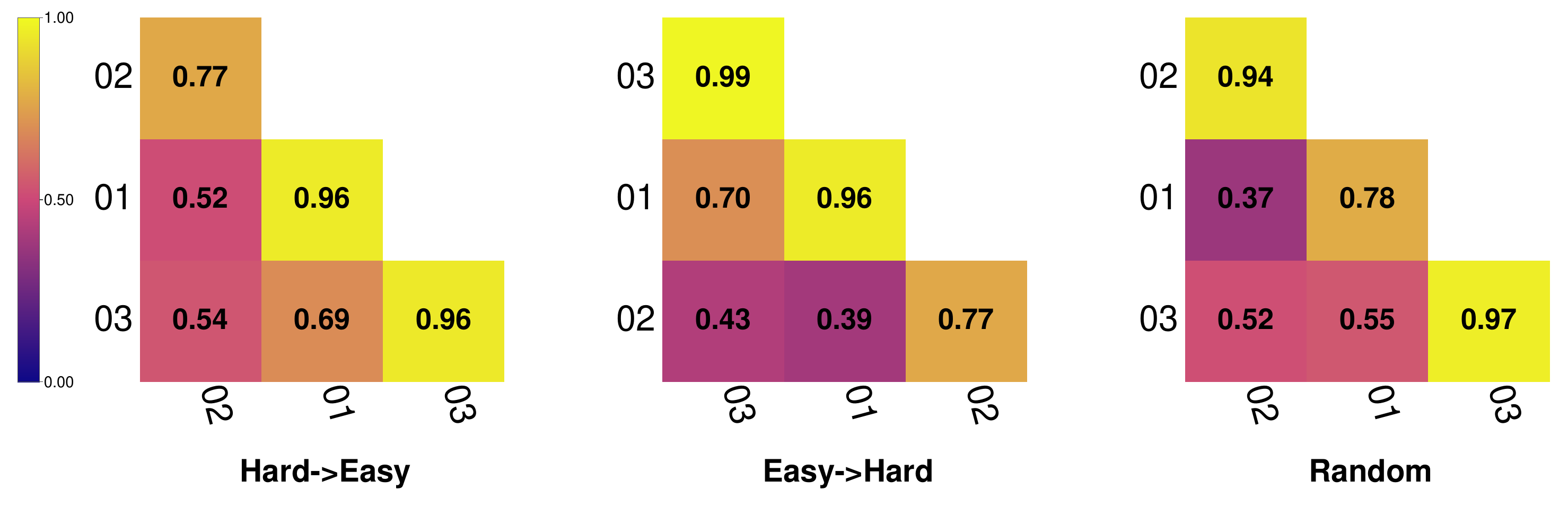}\\
  \includegraphics[width=\textwidth]{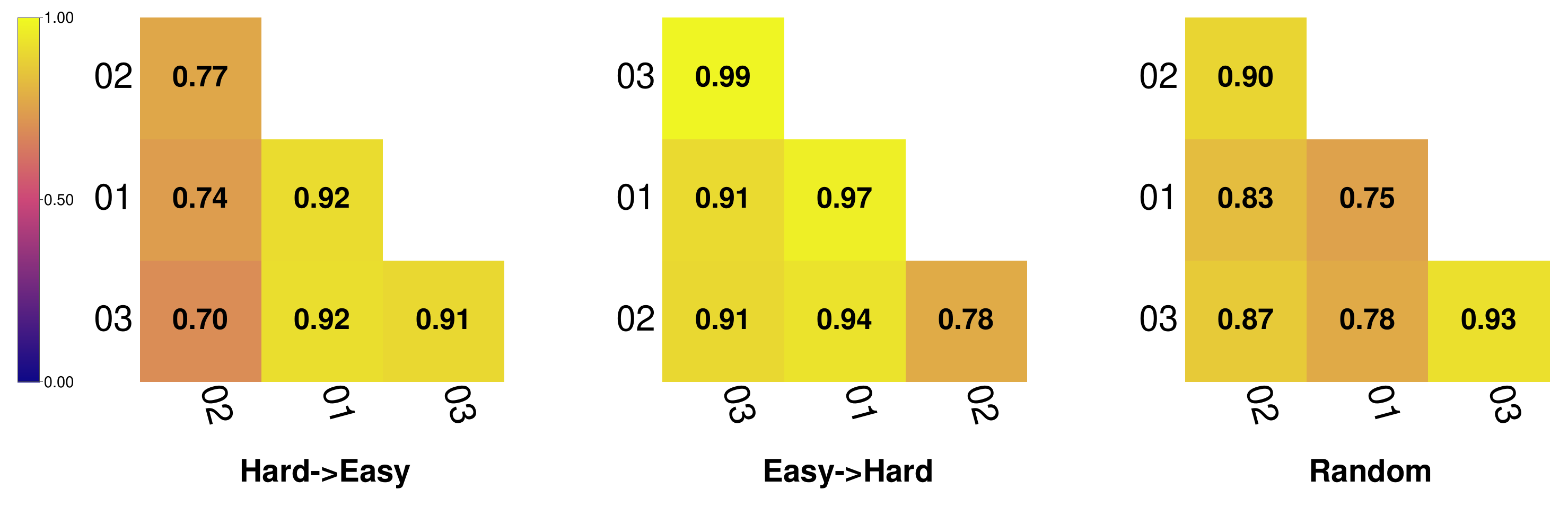}\\
 \rule{\textwidth}{0.4pt}\\[2pt]    
 \includegraphics[width=\textwidth]{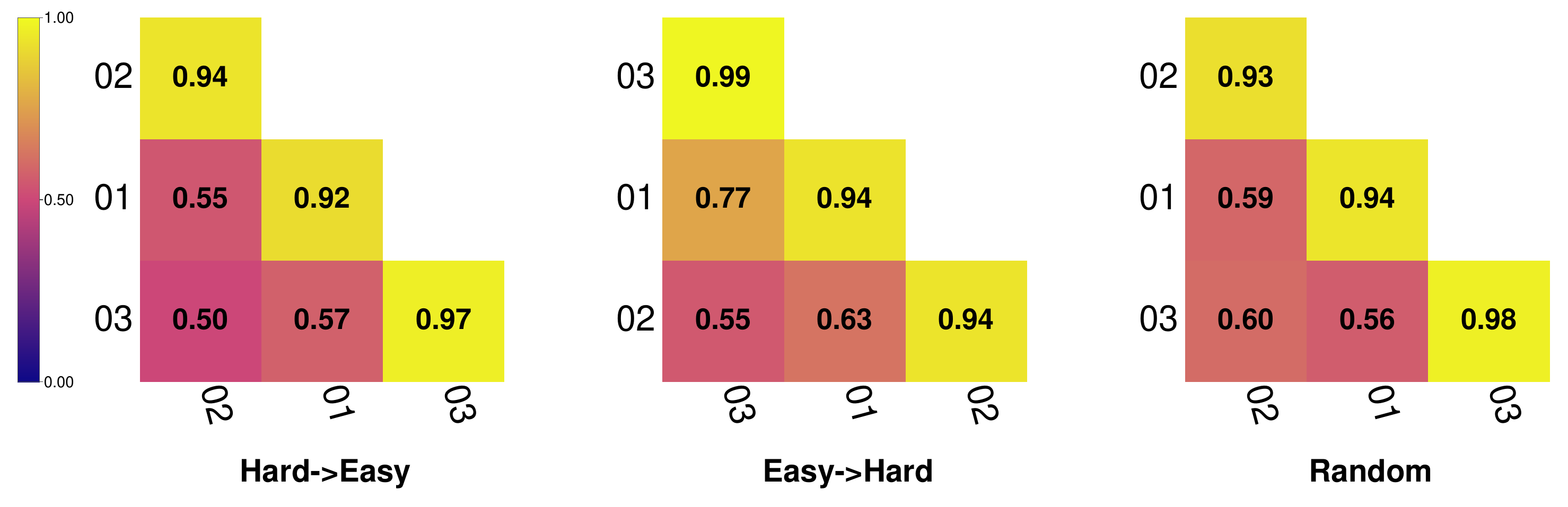}\\[4pt]
  \includegraphics[width=\textwidth]{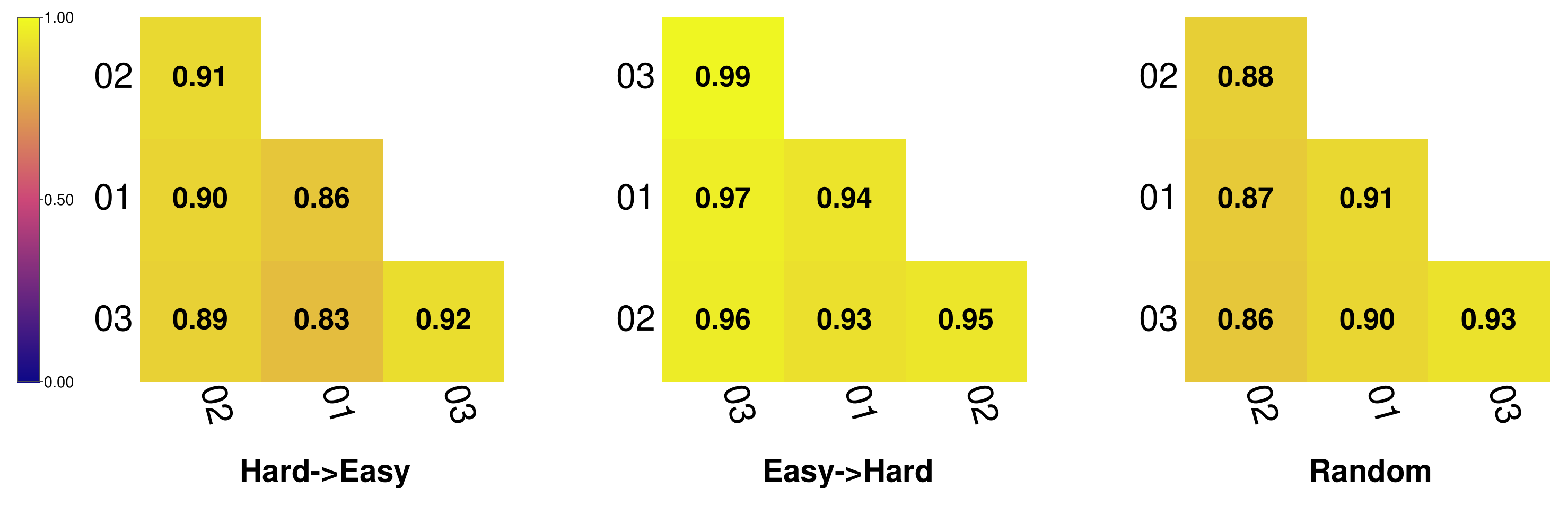}\\
  \caption{\textbf{BTAD}, ResNet-18 : concept transition matrices across three orderings
    (Hard$\to$Easy, Easy$\to$Hard, Random): NCL (top) vs.\ CL (bottom),
    image-level and pixel-level.}
  \label{fig:ordering-btad-r18}
\end{figure*}

\begin{figure*}[!ht]
  \centering
  \includegraphics[width=\textwidth]{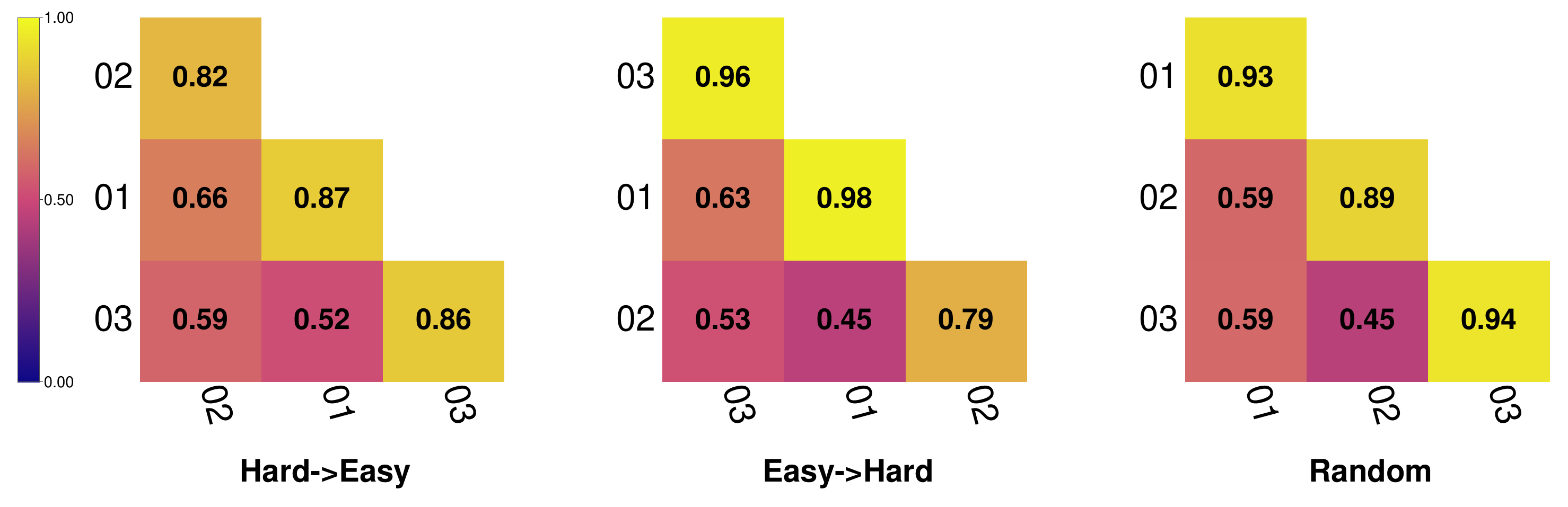}\\
  \includegraphics[width=\textwidth]{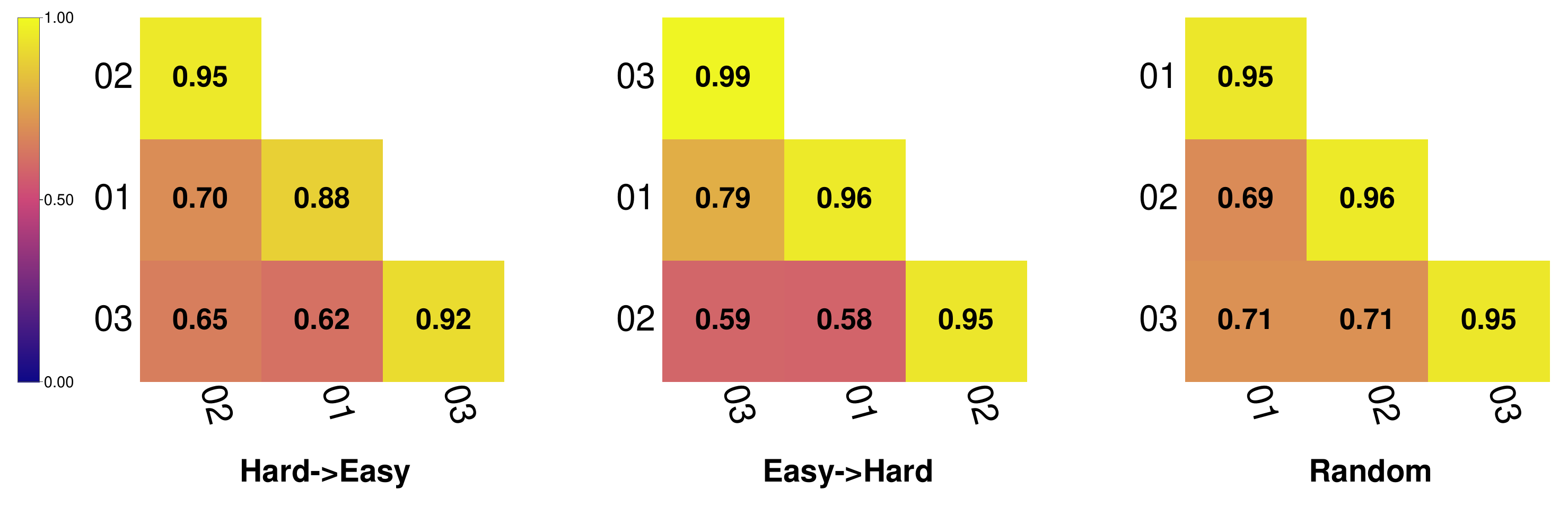}\\
  \rule{\textwidth}{0.4pt}\\[2pt]
  \includegraphics[width=\textwidth]{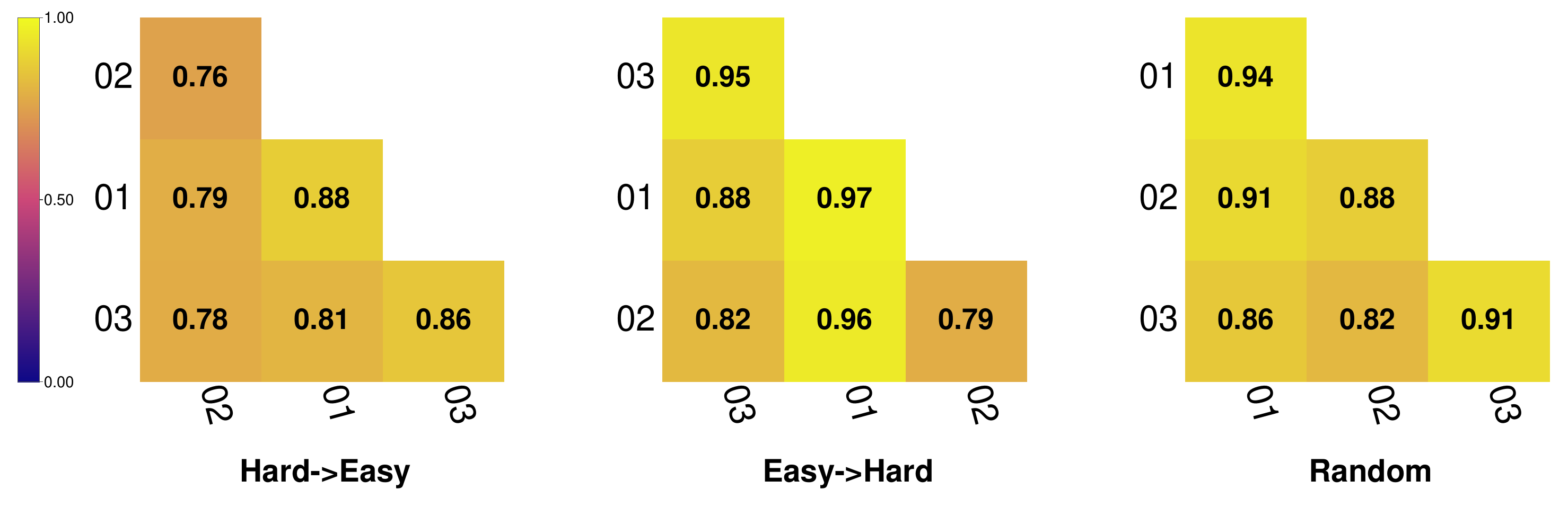}\\
  \includegraphics[width=\textwidth]{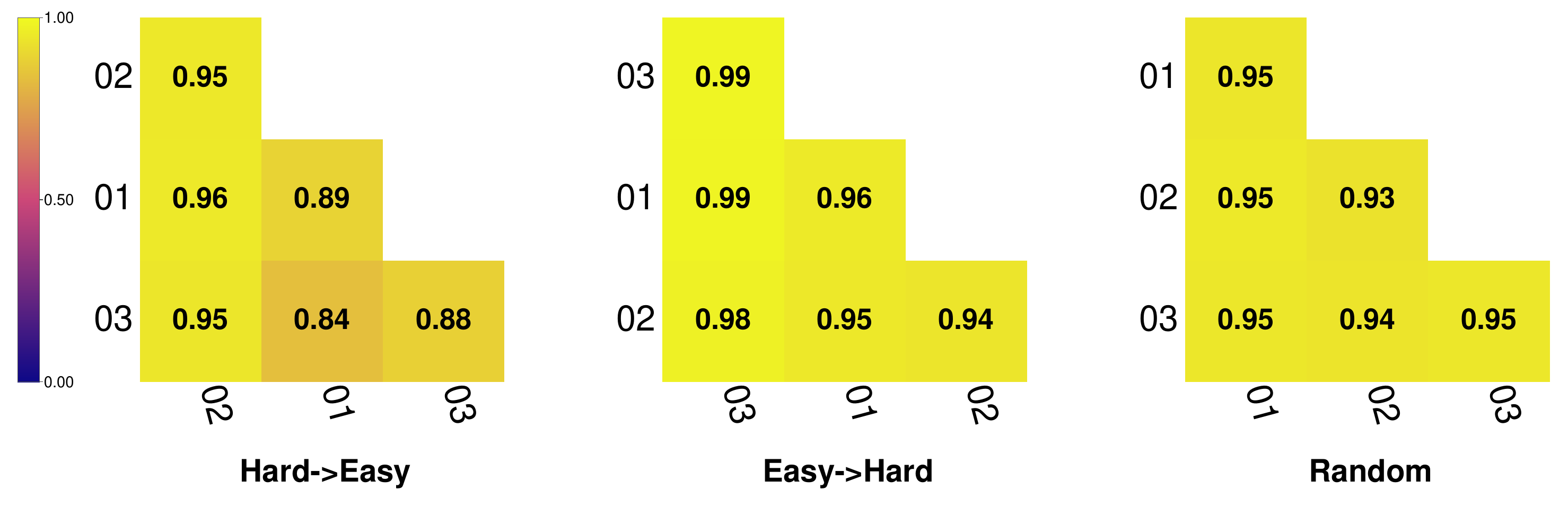}\\
  \caption{\textbf{BTAD}, Wide ResNet-50-2 : concept transition matrices:
    NCL (top) vs.\ CL (bottom), image-level and pixel-level.}
  \label{fig:ordering-btad-wr50}
\end{figure*}

\begin{figure*}[!ht]
  \centering
  \includegraphics[width=\textwidth]{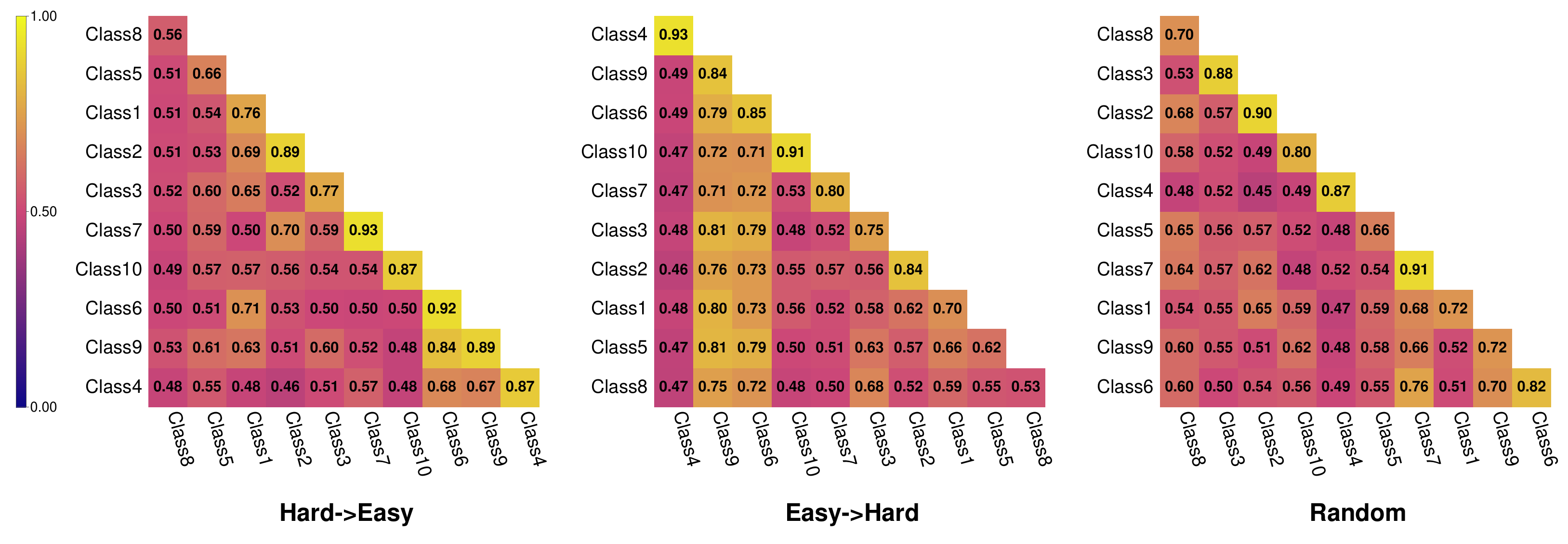}\\
  \includegraphics[width=\textwidth]{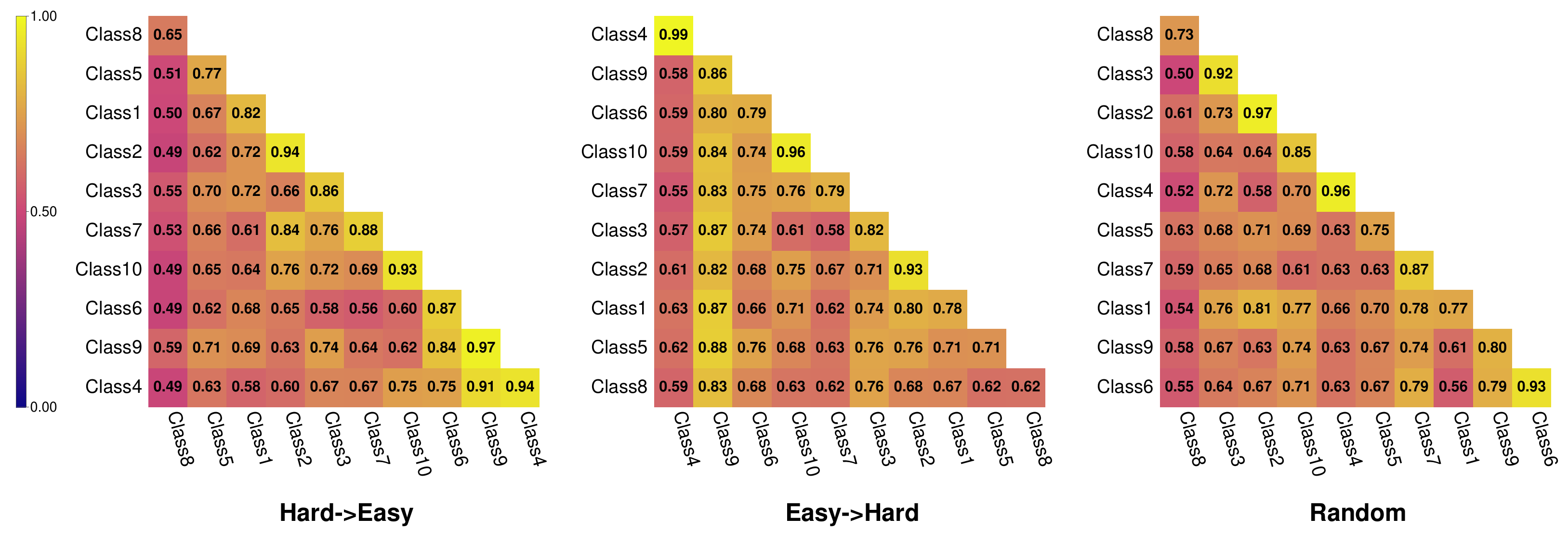}\\
  \rule{\textwidth}{0.4pt}\\[2pt]
  \includegraphics[width=\textwidth]{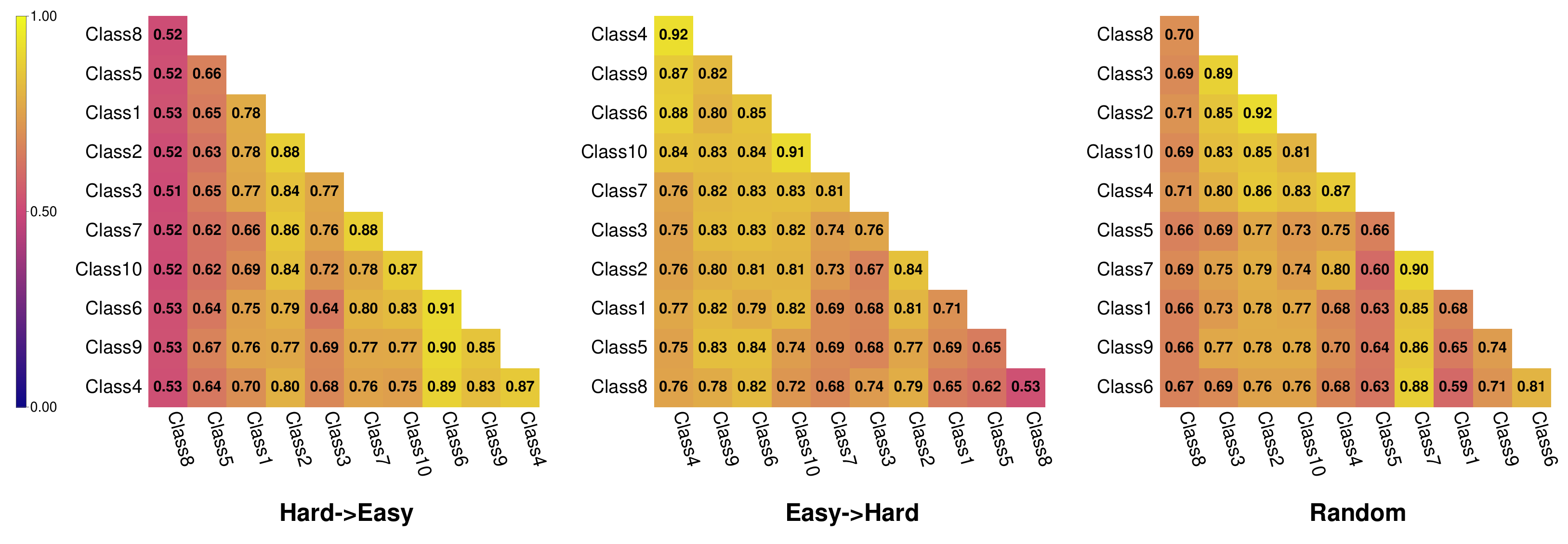}\\
  \includegraphics[width=\textwidth]{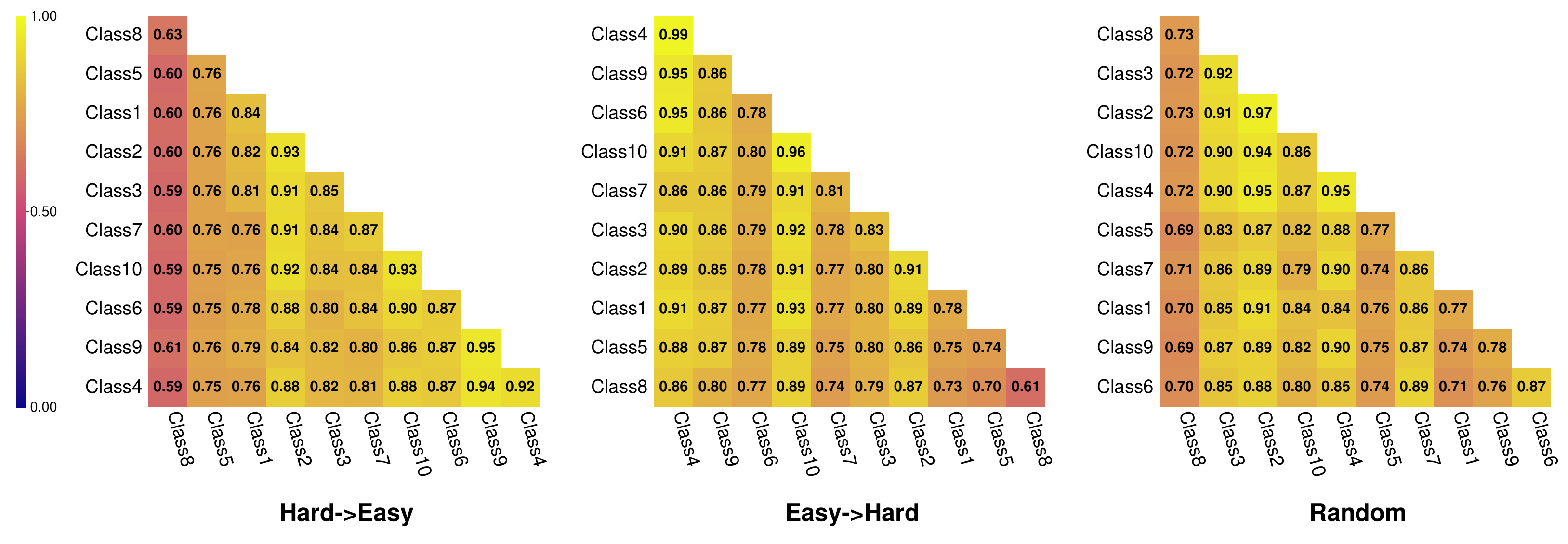}\\
  \caption{\textbf{DAGM}, ResNet-18 : concept transition matrices:
    NCL (top) vs.\ CL (bottom), image-level and pixel-level.}
  \label{fig:ordering-dagm-r18}
\end{figure*}

\begin{figure*}[!ht]
  \centering
  \includegraphics[width=\textwidth]{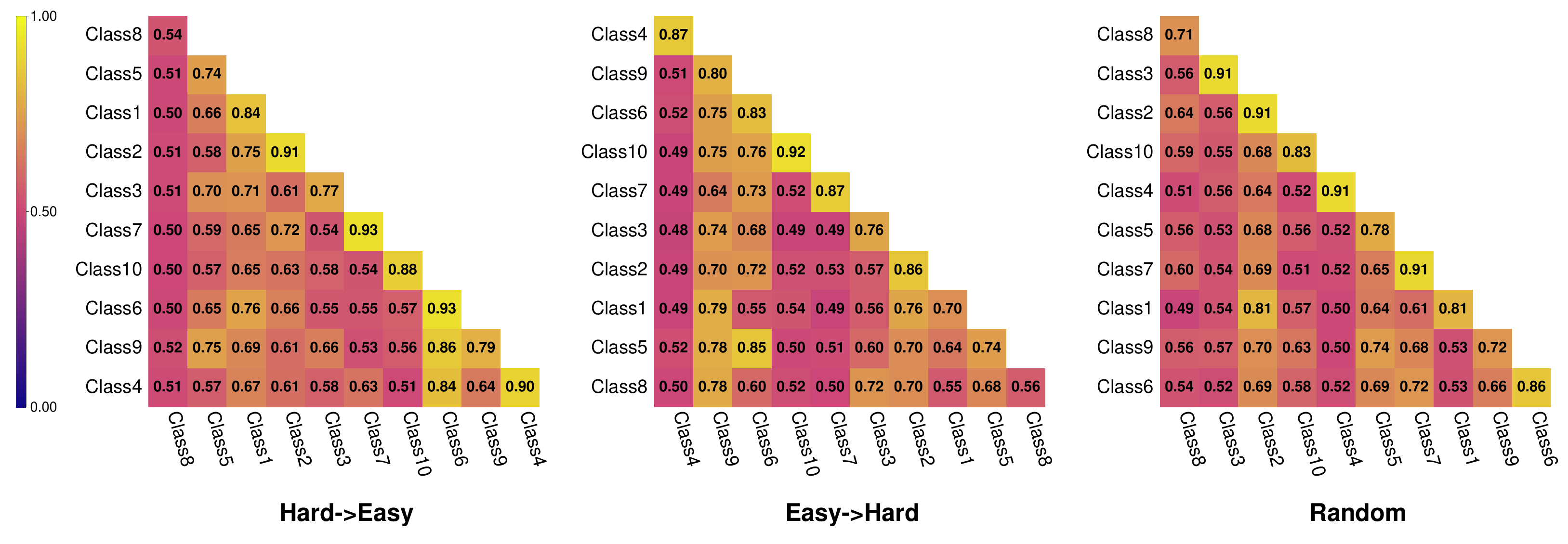}\\
  \includegraphics[width=\textwidth]{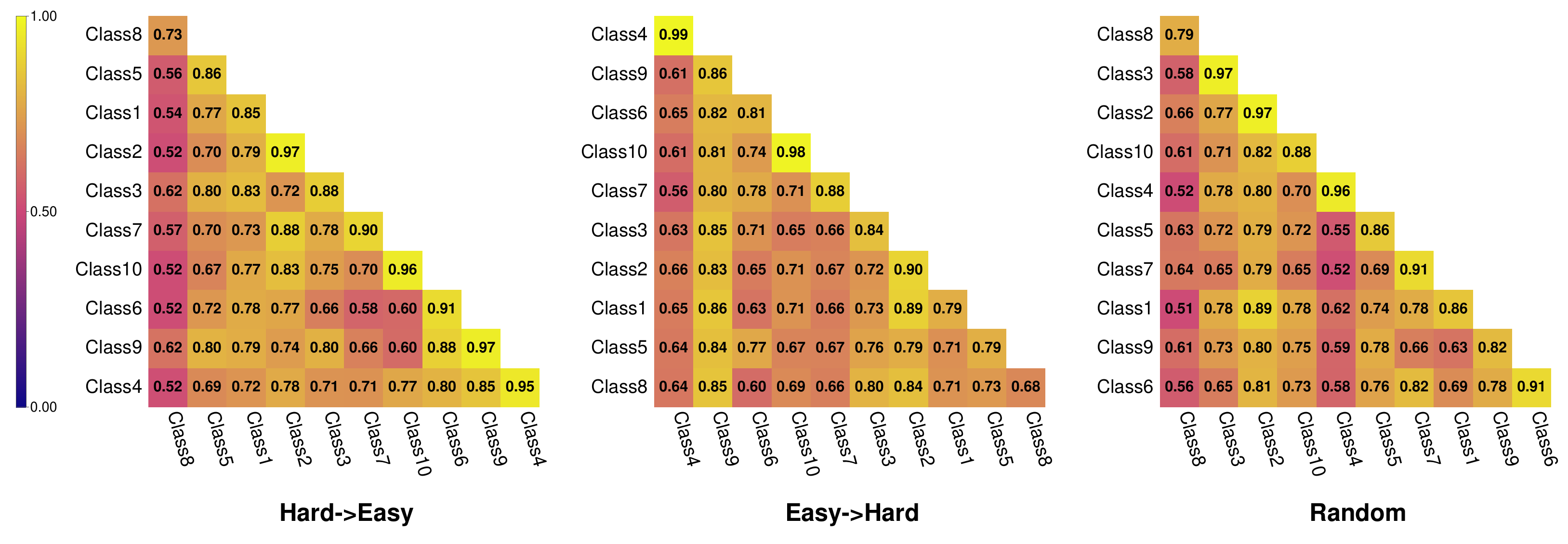}\\
  \rule{\textwidth}{0.4pt}\\[2pt]
  \includegraphics[width=\textwidth]{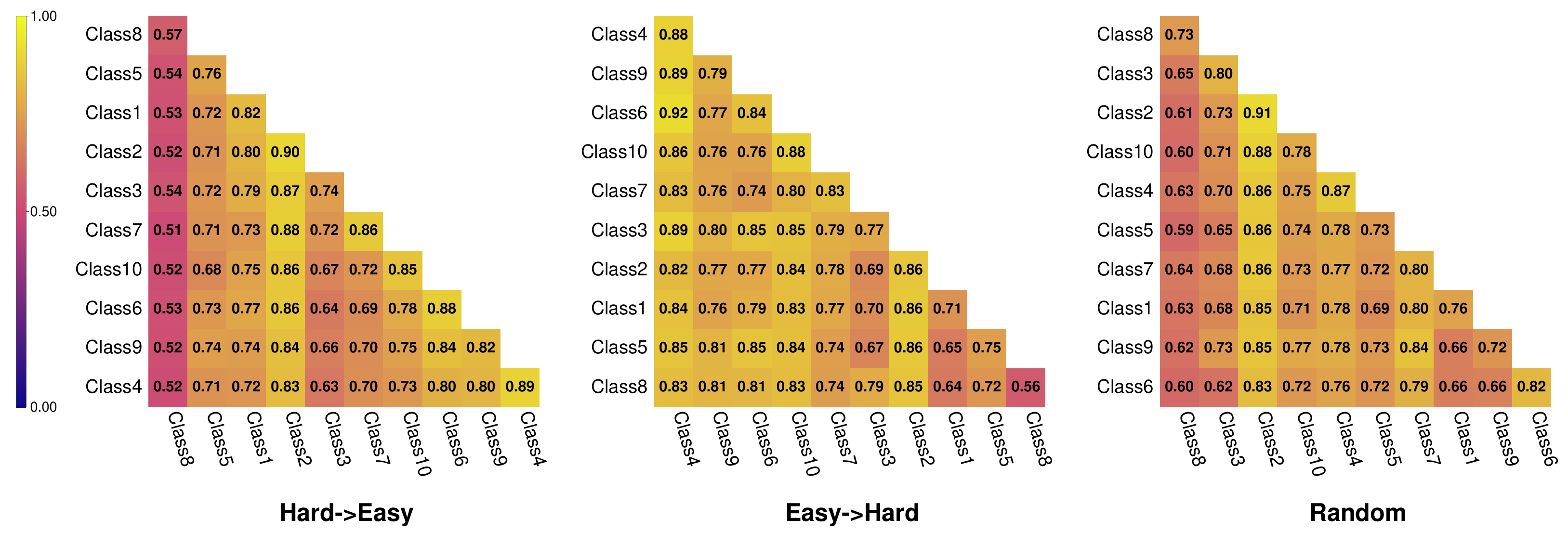}\\
  \includegraphics[width=\textwidth]{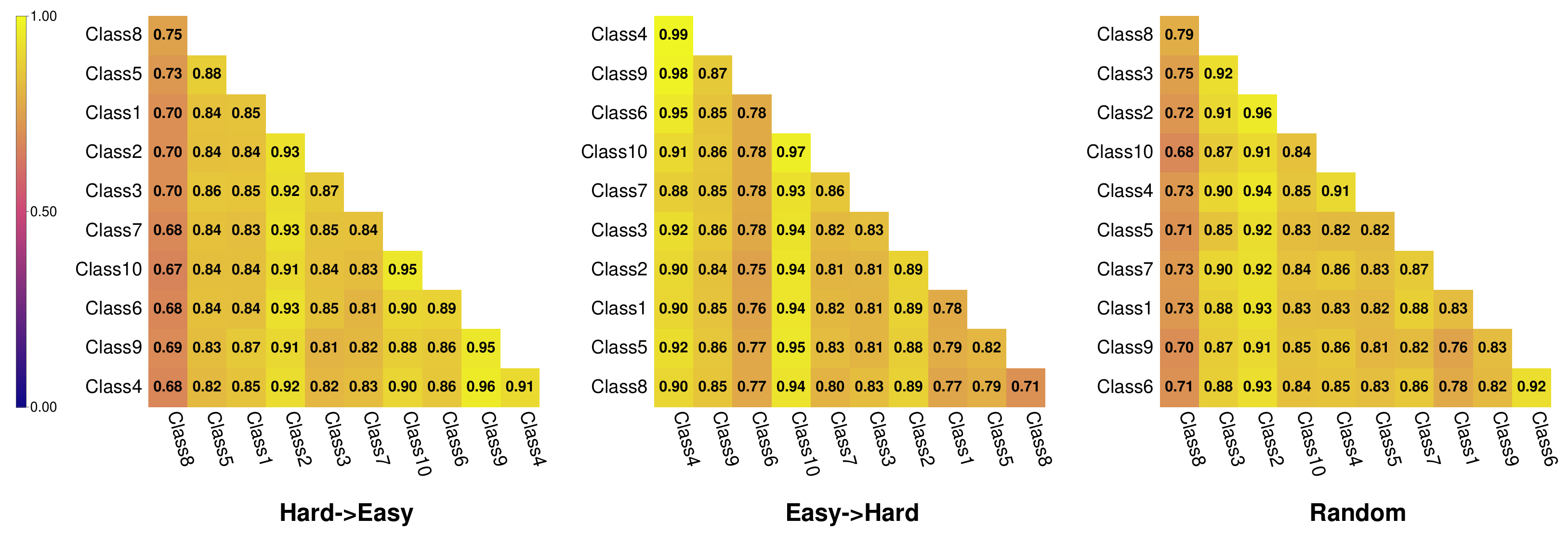}\\
  \caption{\textbf{DAGM}, Wide ResNet-50-2 : concept transition matrices:
    NCL (top) vs.\ CL (bottom), image-level and pixel-level.}
  \label{fig:ordering-dagm-wr50}
\end{figure*}

\begin{figure*}[!ht]
  \centering
  \includegraphics[width=\textwidth]{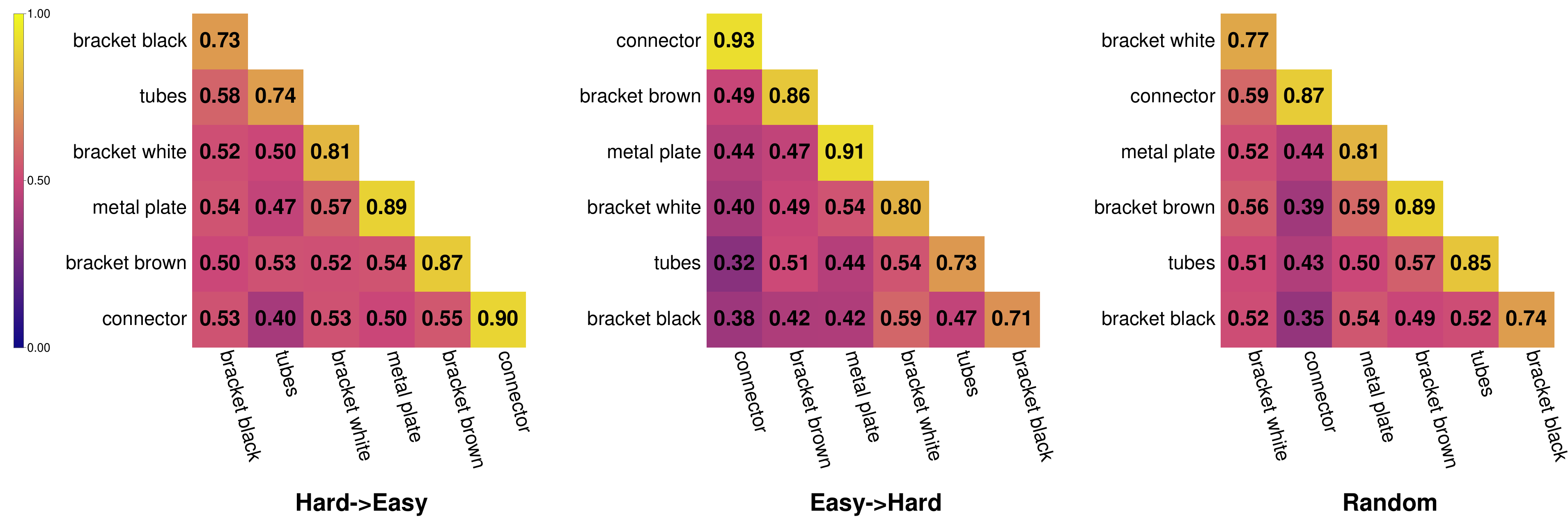}\\
  \includegraphics[width=\textwidth]{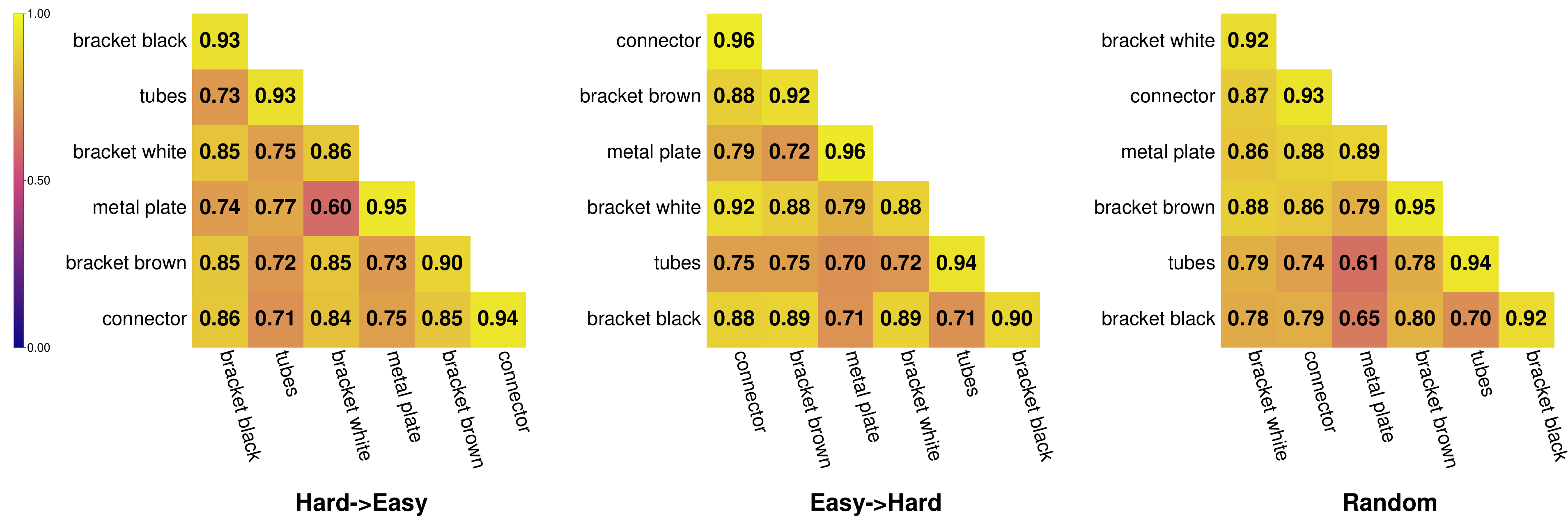}\\
  \rule{\textwidth}{0.4pt}\\[2pt]
  \includegraphics[width=\textwidth]{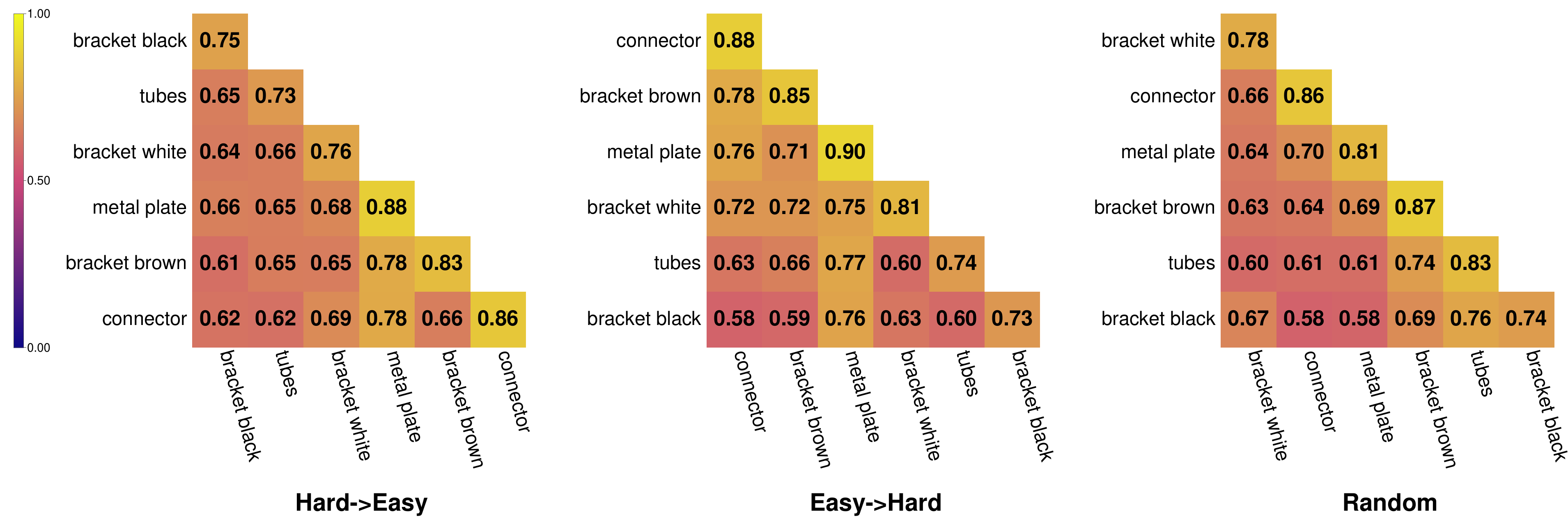}\\
  \includegraphics[width=\textwidth]{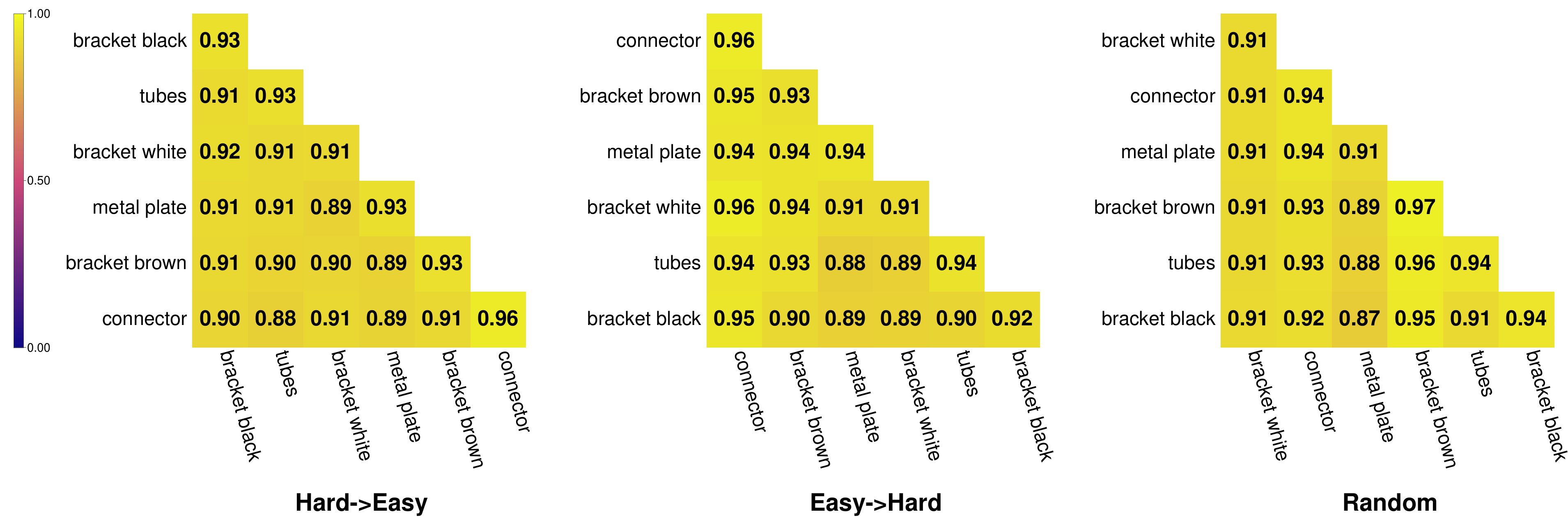}\\
  \caption{\textbf{MPDD}, ResNet-18 : concept transition matrices:
    NCL (top) vs.\ CL (bottom), image-level and pixel-level.}
  \label{fig:ordering-mpdd-r18}
\end{figure*}

\begin{figure*}[!ht]
  \centering
  \includegraphics[width=\textwidth]{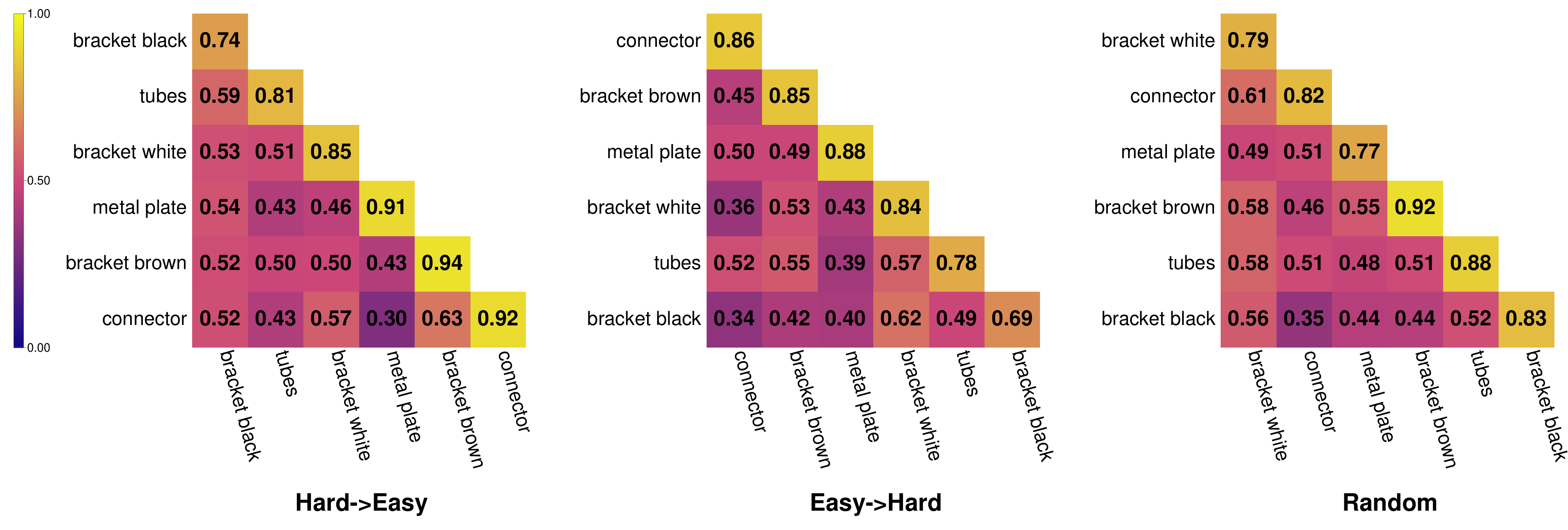}\\
  \includegraphics[width=\textwidth]{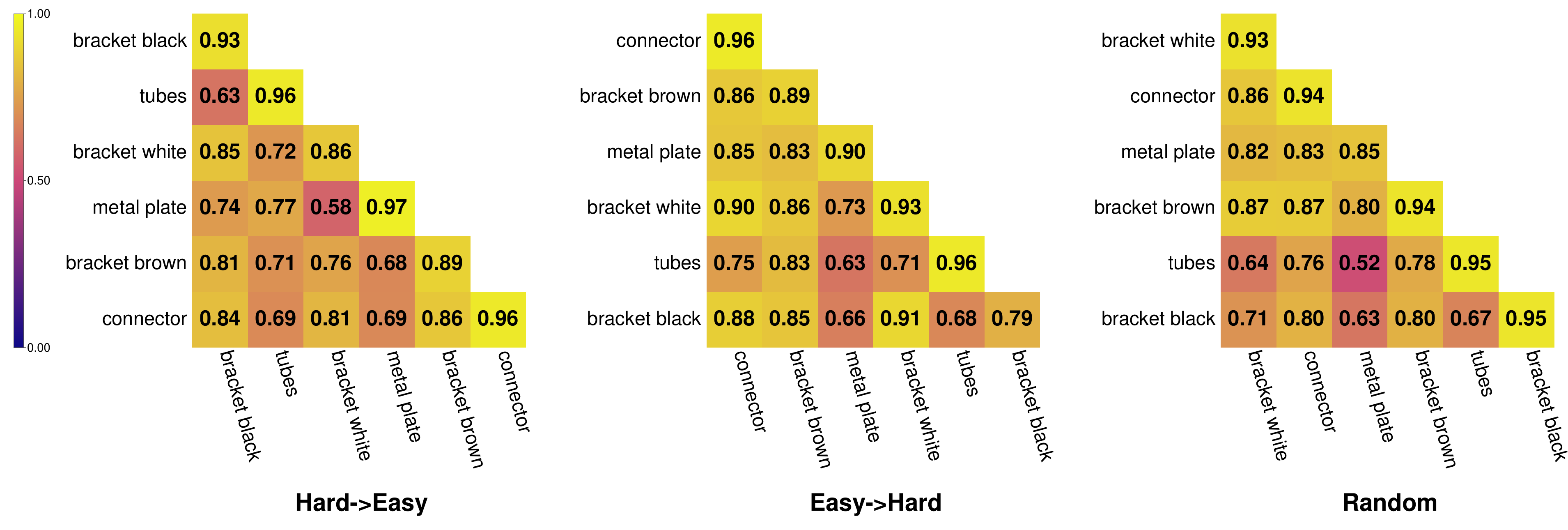}\\
  \rule{\textwidth}{0.4pt}\\[2pt]
  \includegraphics[width=\textwidth]{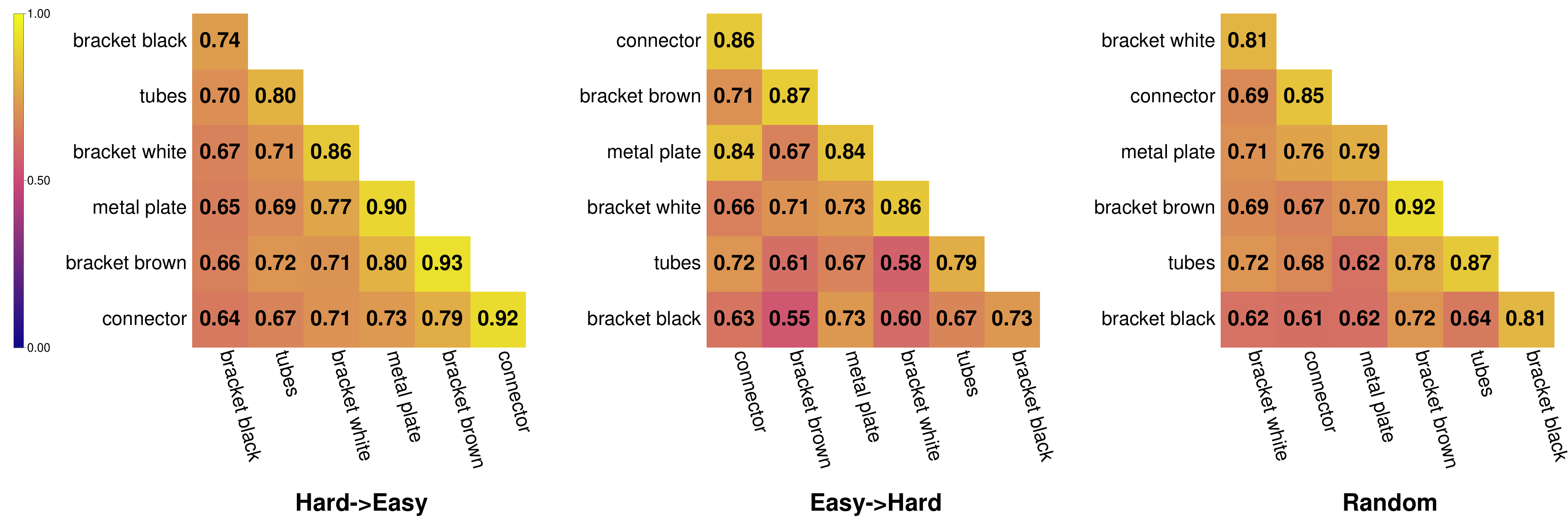}\\
  \includegraphics[width=\textwidth]{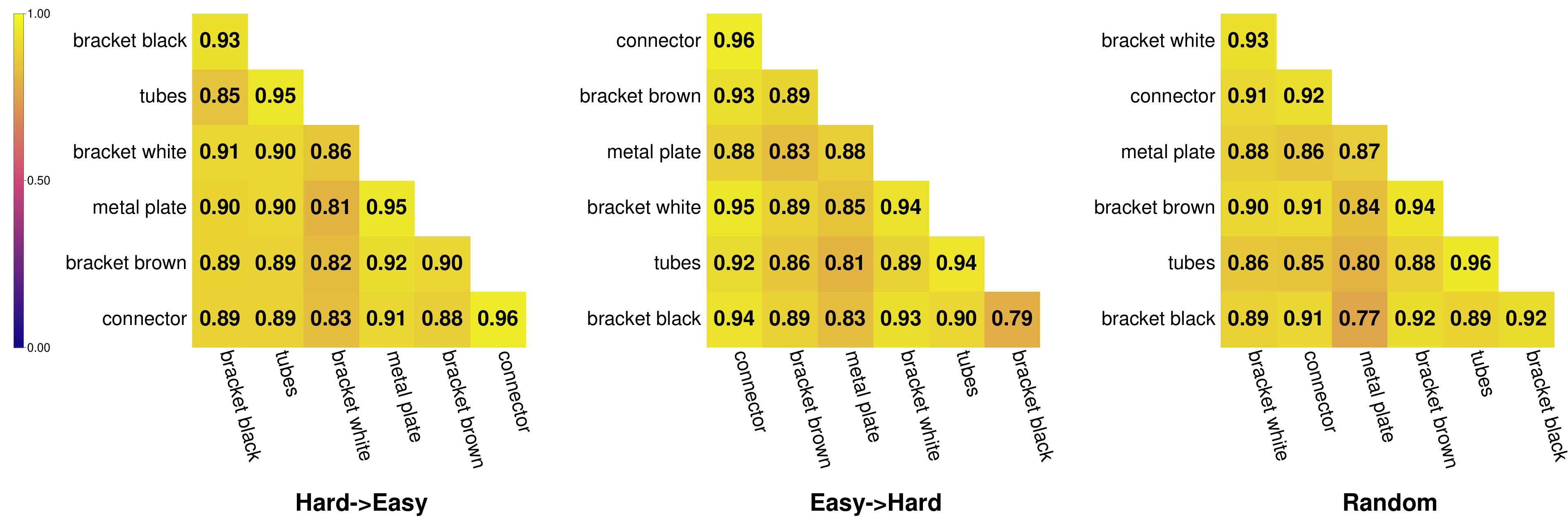}\\
  \caption{\textbf{MPDD}, Wide ResNet-50-2 : concept transition matrices:
    NCL (top) vs.\ CL (bottom), image-level and pixel-level.}
  \label{fig:ordering-mpdd-wr50}
\end{figure*}

\begin{figure*}[!ht]
  \centering
  \includegraphics[width=\textwidth]{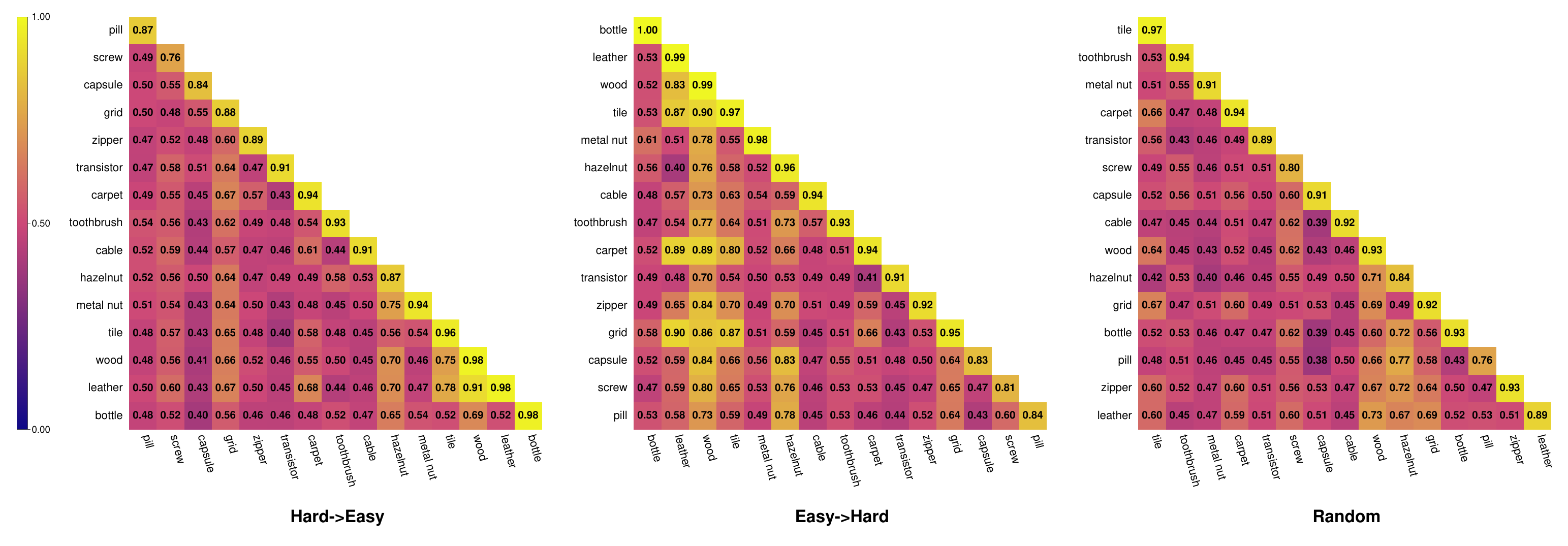}\\
  \includegraphics[width=\textwidth]{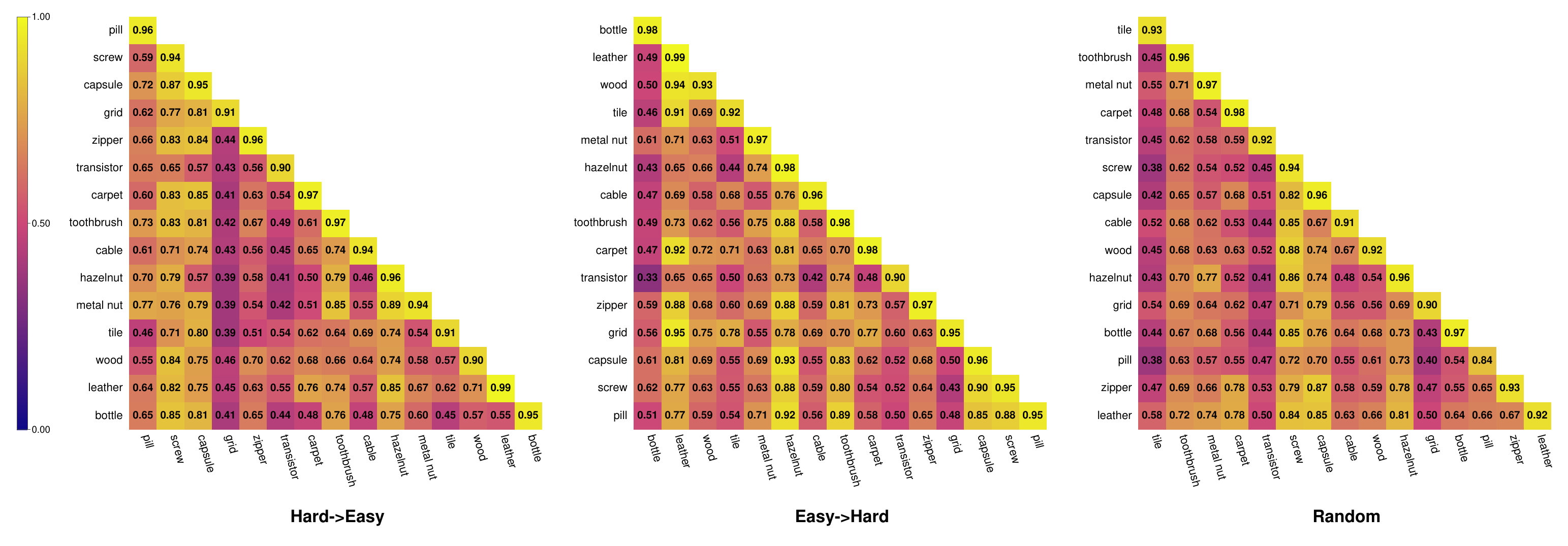}\\
  \rule{\textwidth}{0.4pt}\\[2pt]
  \includegraphics[width=\textwidth]{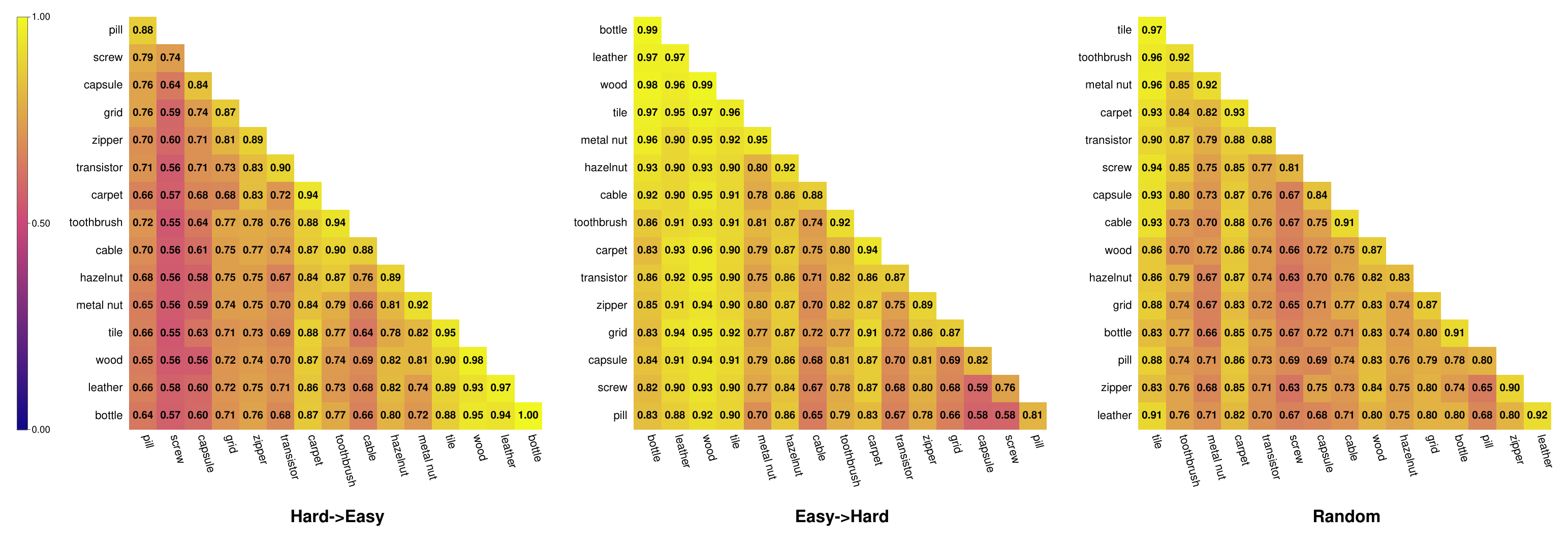}\\
  \includegraphics[width=\textwidth]{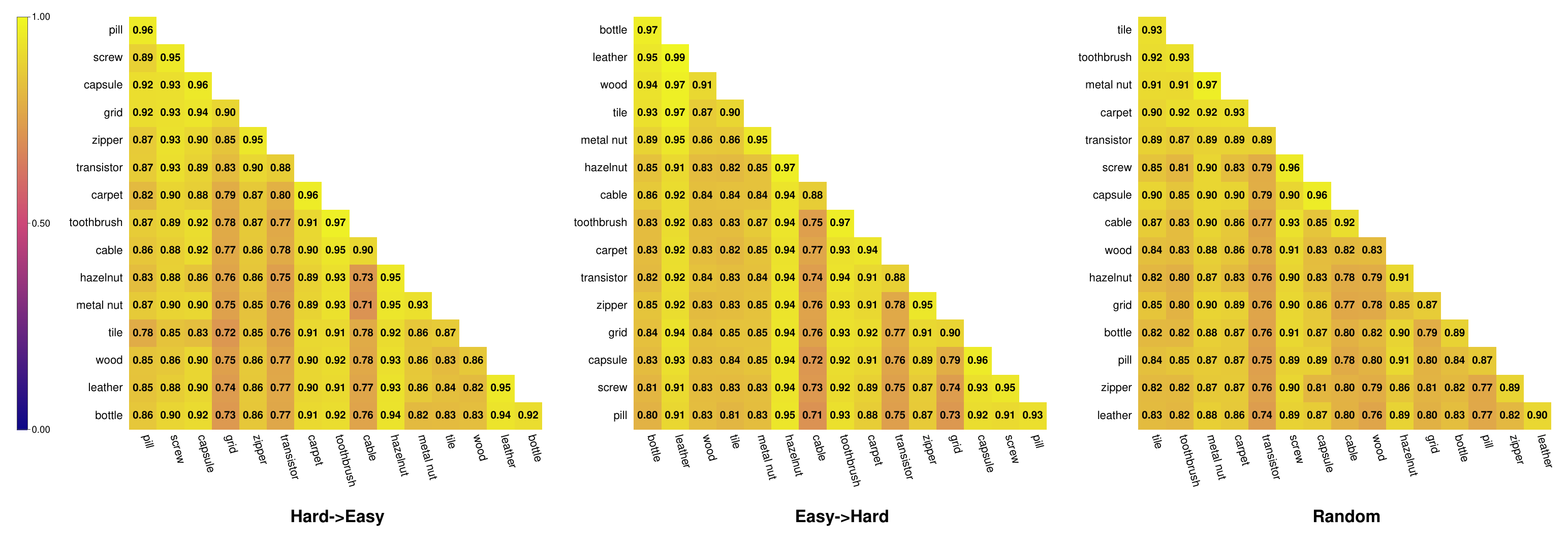}\\
  \caption{\textbf{MVTec}, ResNet-18 : concept transition matrices:
    NCL (top) vs.\ CL (bottom), image-level and pixel-level.}
  \label{fig:ordering-mvtec-r18}
\end{figure*}

\begin{figure*}[!ht]
  \centering
  \includegraphics[width=\textwidth]{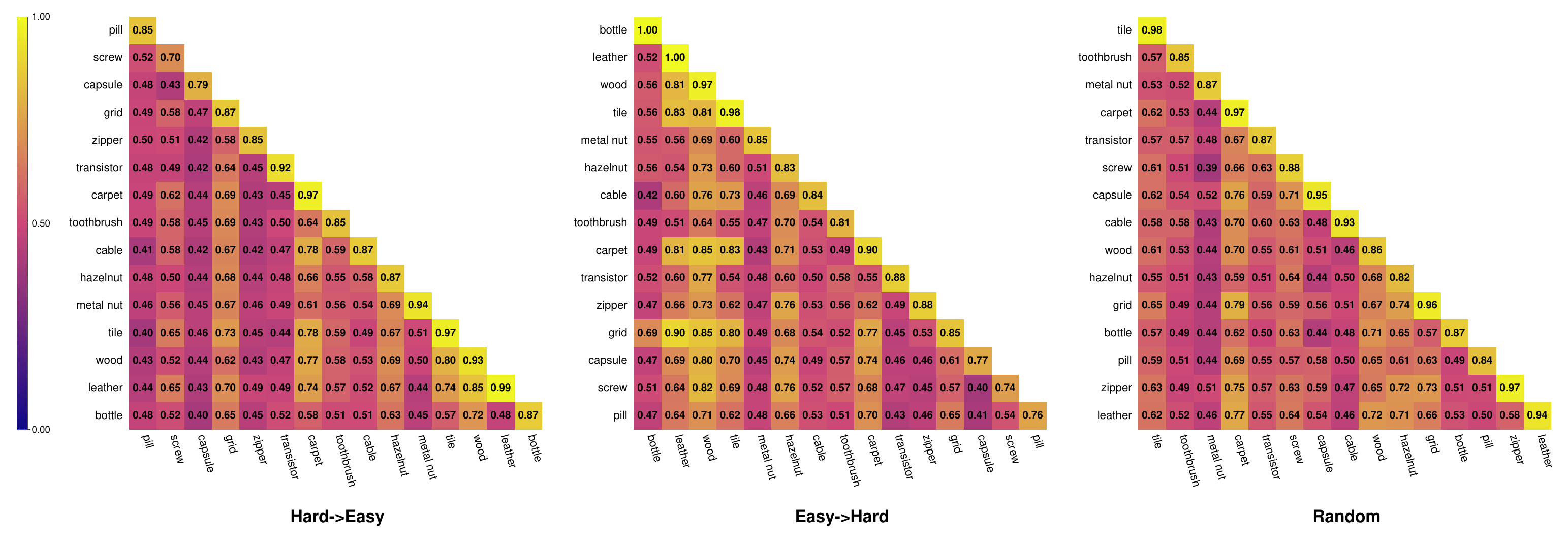}\\
  \includegraphics[width=\textwidth]{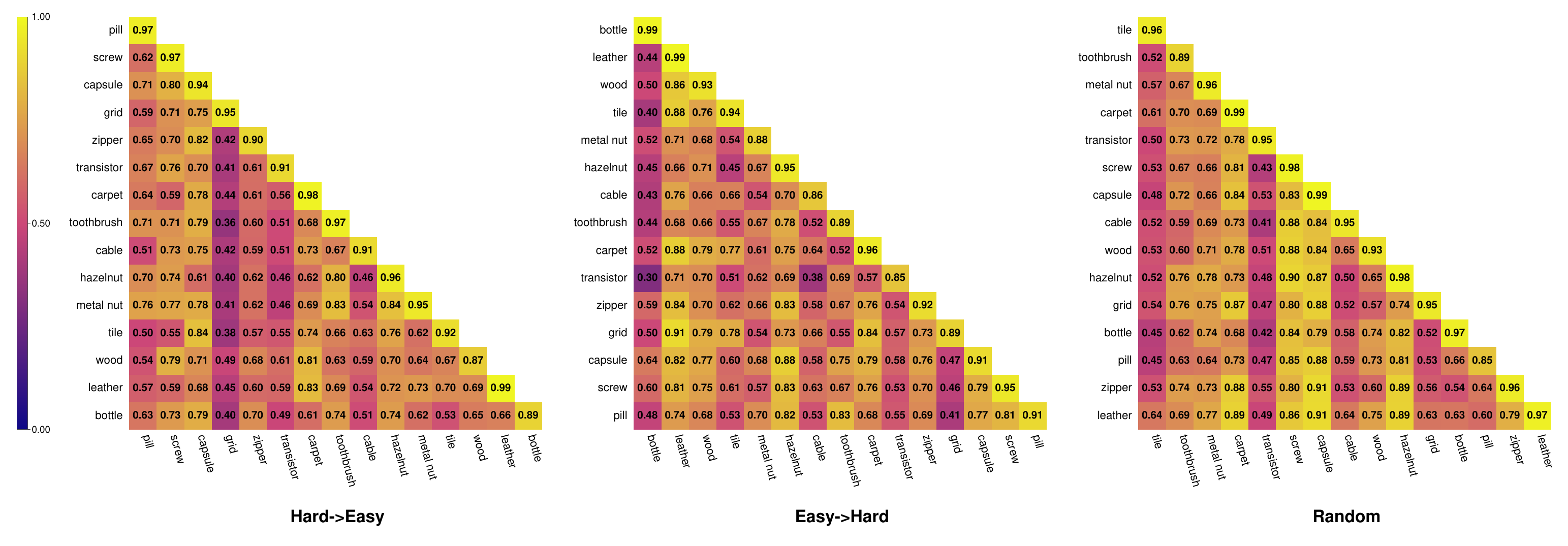}\\
  \rule{\textwidth}{0.4pt}\\[2pt]
  \includegraphics[width=\textwidth]{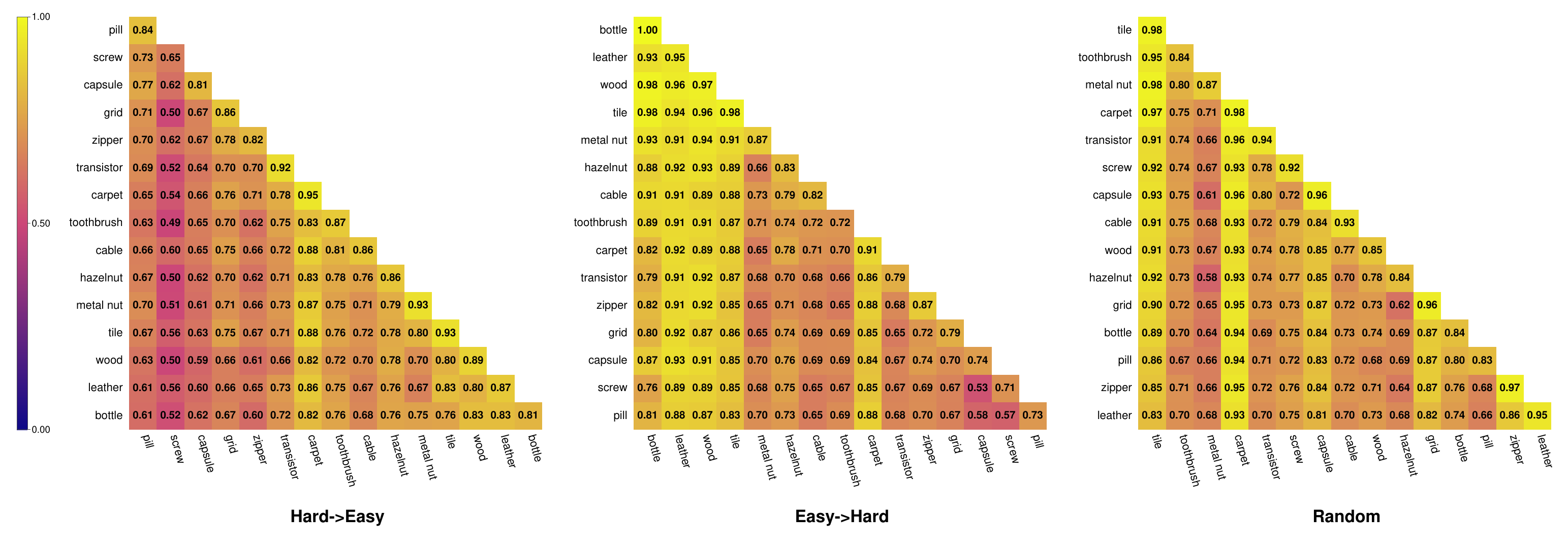}\\
  \includegraphics[width=\textwidth]{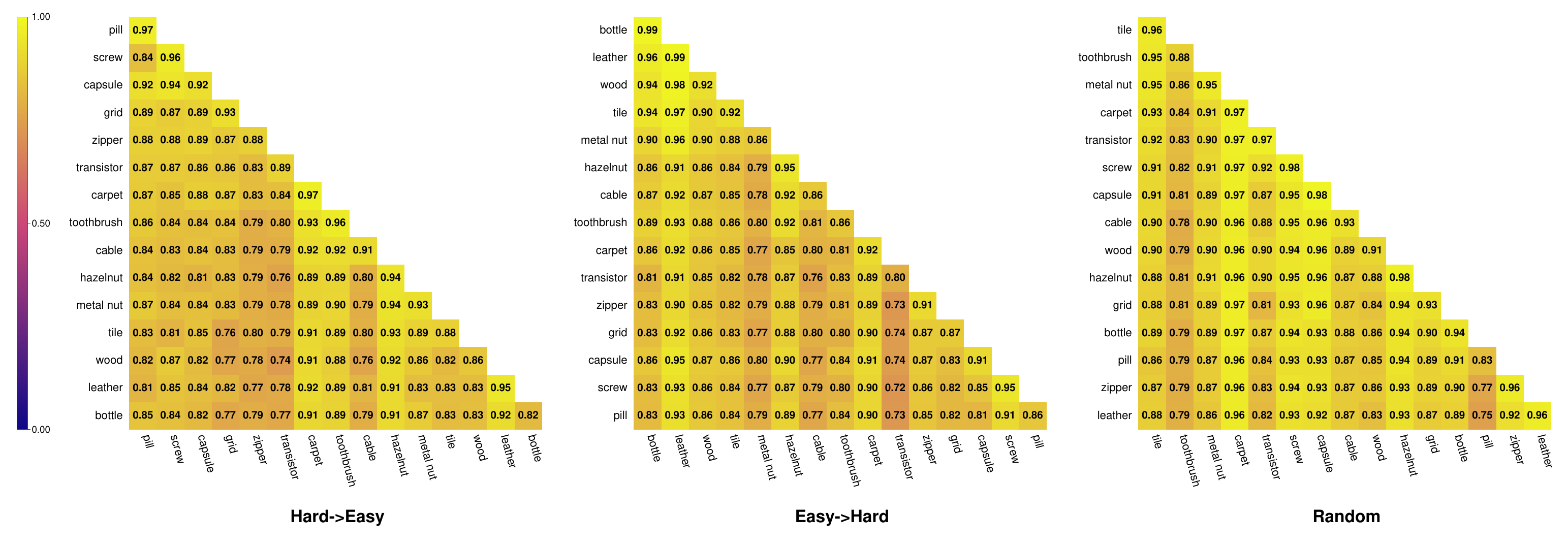}\\
  \caption{\textbf{MVTec}, Wide ResNet-50-2 : concept transition matrices:
    NCL (top) vs.\ CL (bottom), image-level and pixel-level.}
  \label{fig:ordering-mvtec-wr50}
\end{figure*}

\begin{figure*}[!ht]
  \centering
  \includegraphics[width=\textwidth]{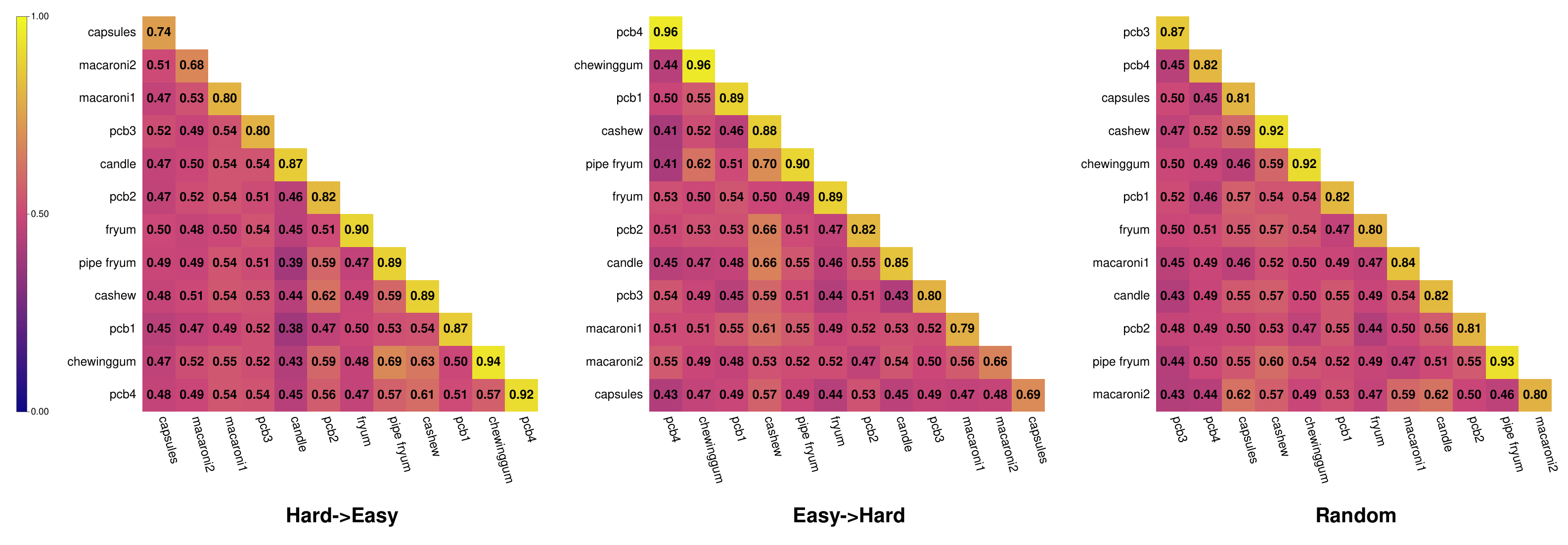}\\
  \includegraphics[width=\textwidth]{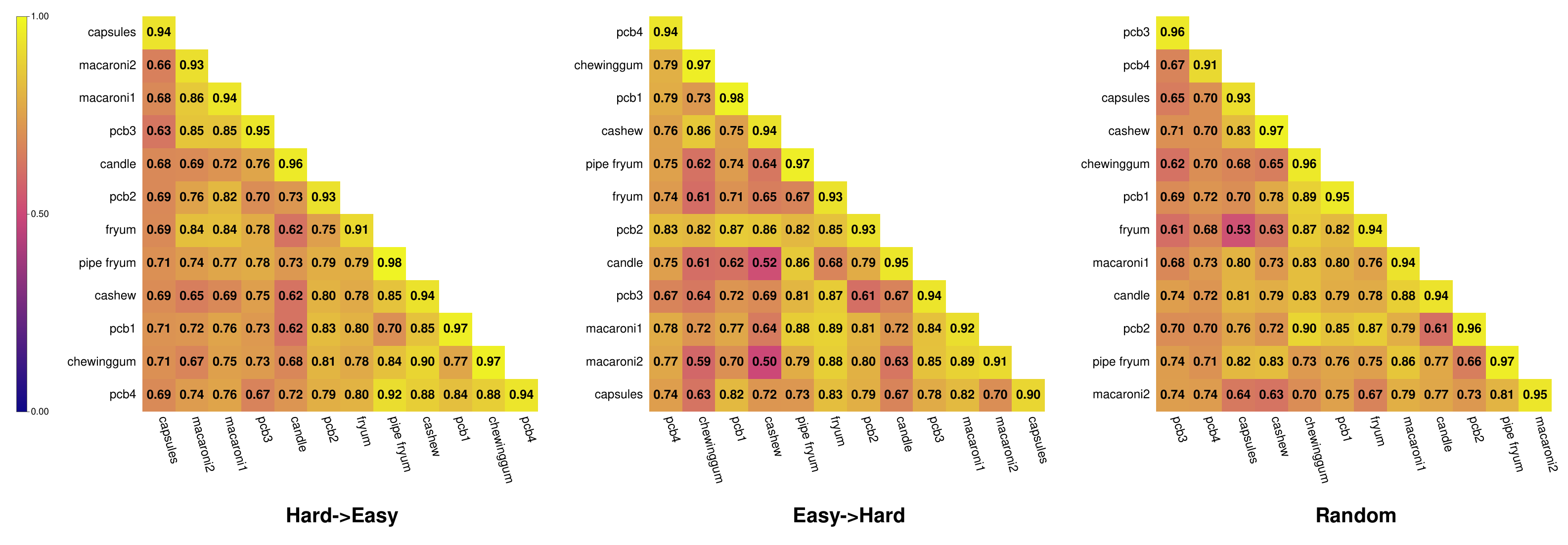}\\
  \rule{\textwidth}{0.4pt}\\[2pt]
  \includegraphics[width=\textwidth]{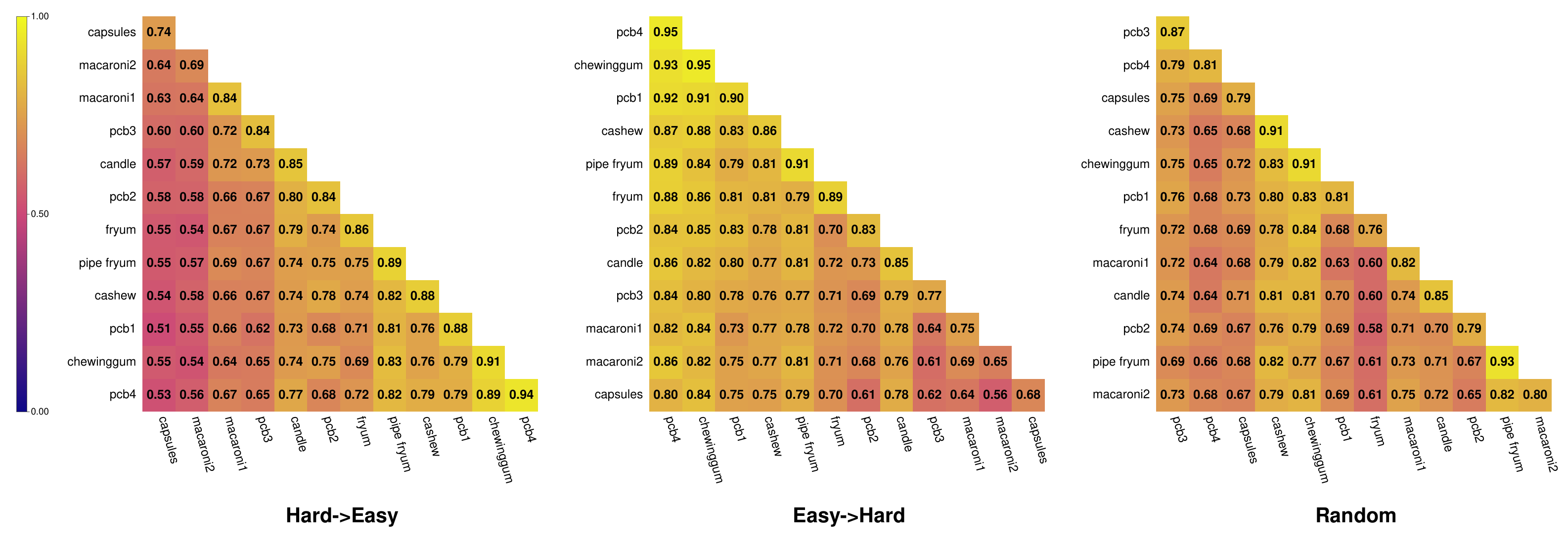}\\
  \includegraphics[width=\textwidth]{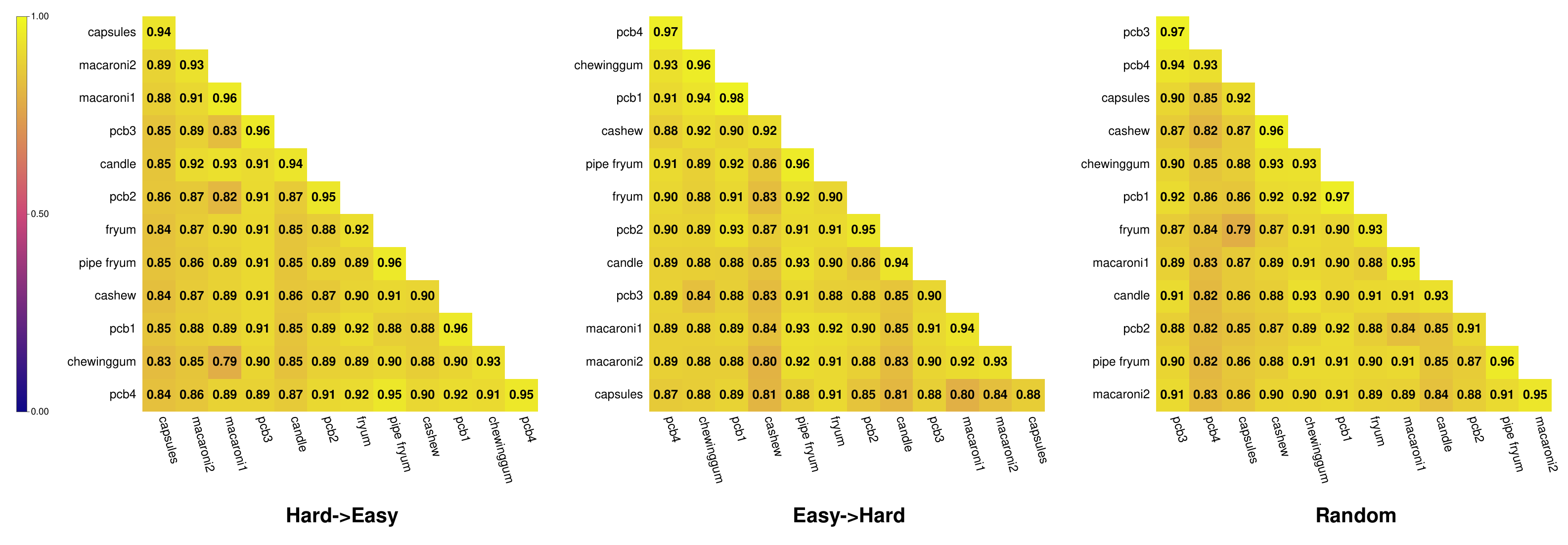}\\
  \caption{\textbf{VisA}, ResNet-18 : concept transition matrices:
    NCL (top) vs.\ CL (bottom), image-level and pixel-level.}
  \label{fig:ordering-visa-r18}
\end{figure*}

\begin{figure*}[!ht]
  \centering
  \includegraphics[width=\textwidth]{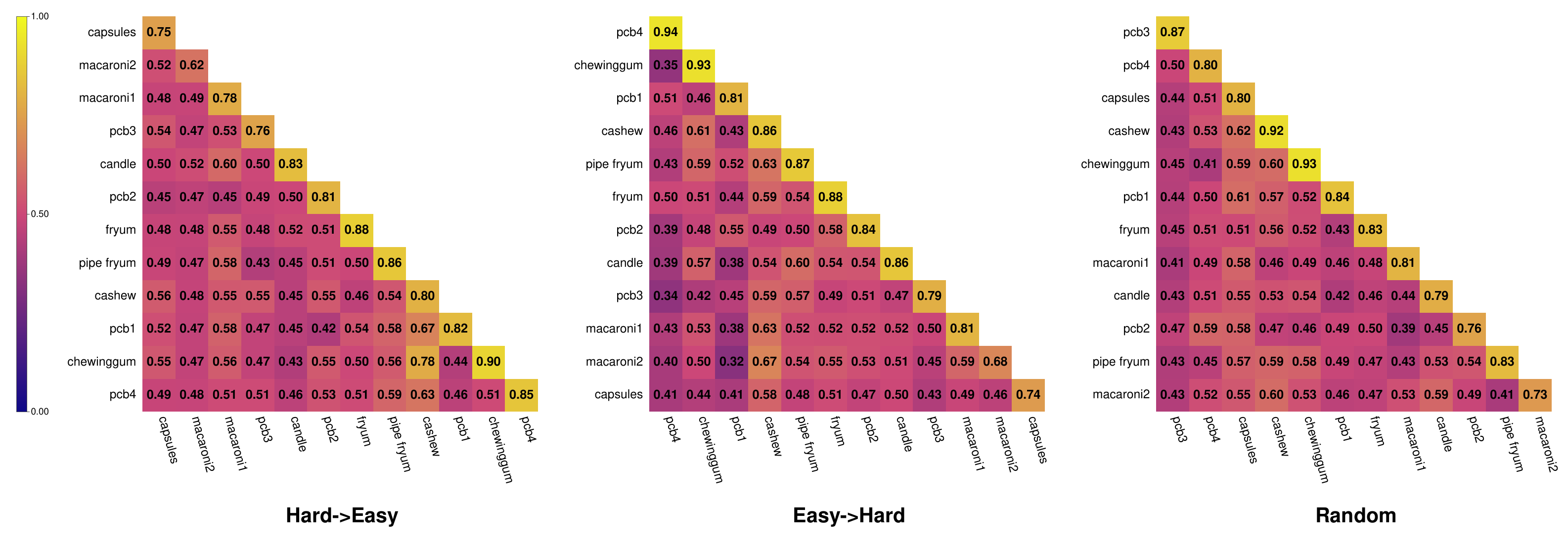}\\
  \includegraphics[width=\textwidth]{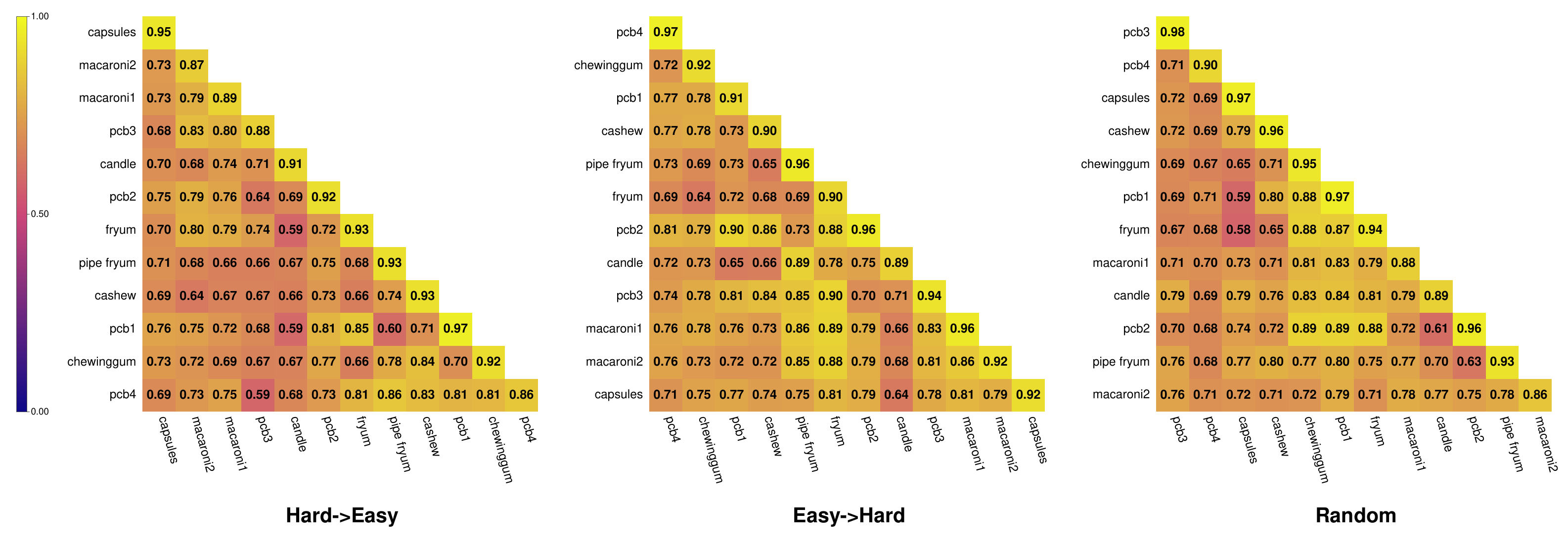}\\
  \rule{\textwidth}{0.4pt}\\[2pt]
  \includegraphics[width=\textwidth]{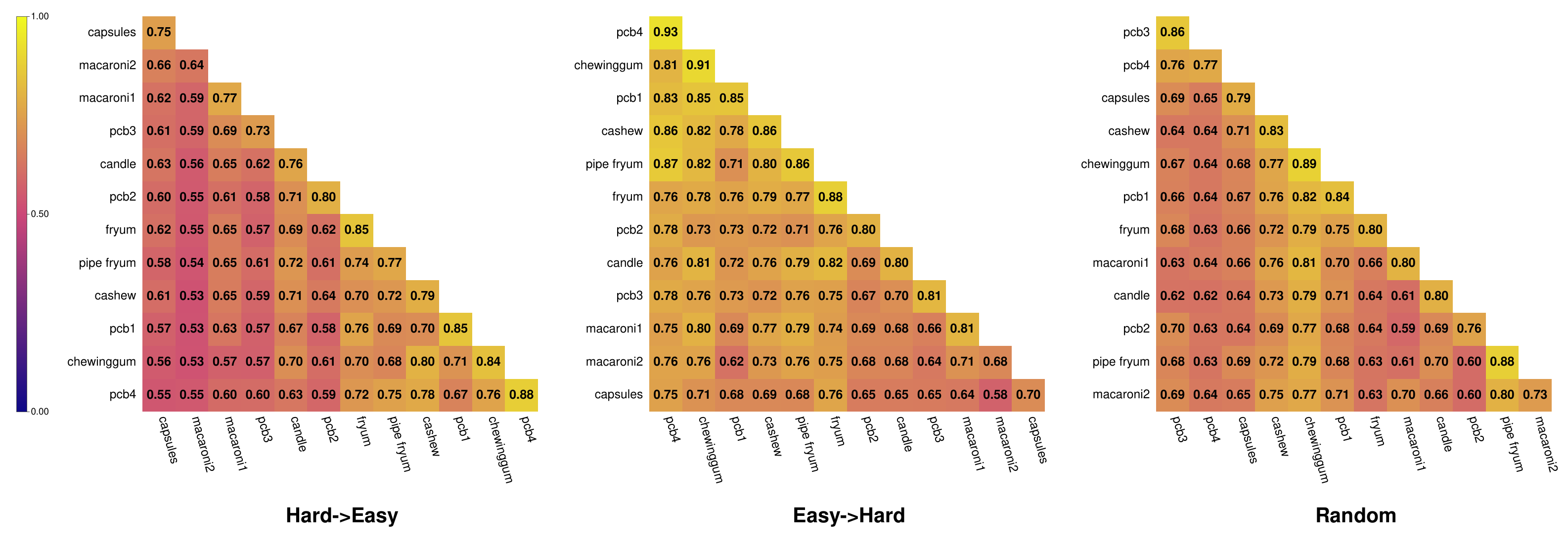}\\
  \includegraphics[width=\textwidth]{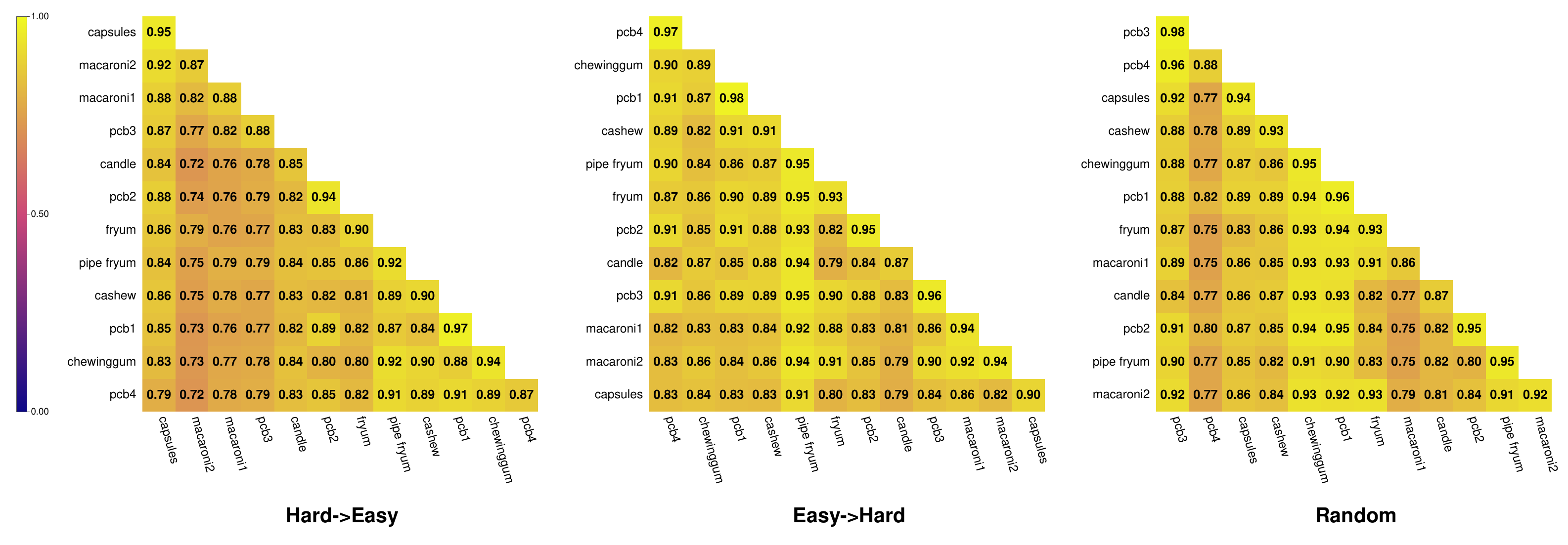}\\
  \caption{\textbf{VisA}, Wide ResNet-50-2 : concept transition matrices:
    NCL (top) vs.\ CL (bottom), image-level and pixel-level.}
  \label{fig:ordering-visa-wr50}
\end{figure*}

Figures~\ref{fig:ordering-btad-r18}--\ref{fig:ordering-visa-wr50} provide the
task-transition heatmaps used to complement the aggregate results in
Table~\ref{tab:summary-avg-dataset-backbones}. Each triangular block corresponds
to one ordering. Diagonal cells show performance when a task is first learned,
whereas off-diagonal cells show whether that task is retained after subsequent
updates. Across datasets, the main pattern is that NCL often preserves strong
diagonal performance but loses intensity in off-diagonal cells. CL produces
warmer and more uniform lower-triangular regions, indicating that the gains in
the aggregate table are driven not only by better final scores but also by
better retention of individual tasks.

\paragraph{Ordering effects.}
Ordering effects are most visible in the off-diagonal cells. In NCL, the same
task can perform well immediately after training and then degrade sharply after
later tasks are introduced. This is especially visible in BTAD and MPDD at the
image level, where several early-task cells become much darker after subsequent
updates. Random streams usually fall between the two curriculum directions,
confirming that the benchmark captures the interaction between task sequence and
the update rule rather than only average dataset difficulty. CL reduces this
ordering sensitivity: the color contrast between hard-to-easy, easy-to-hard, and
random blocks becomes less severe, particularly for pixel-level evaluation. Thus,
the heatmaps refine the aggregate result by showing that ordering effects are
often task-local, with a small number of vulnerable concepts driving much of the
forgetting signal.

\paragraph{Adaptation versus retention.}
The NCL--CL contrast is strongest on retained-task entries rather than on newly
learned diagonal entries. BTAD provides the clearest example: NCL includes high
diagonal values, but retained performance on earlier tasks drops substantially;
CL raises many of those off-diagonal cells and makes the three ordering blocks
more homogeneous. MPDD shows a related but more nuanced pattern. Image-level CL
improves retention relative to NCL, but the heatmaps remain less uniform than
their pixel-level counterparts, matching the main-paper observation that MPDD is
harder for image-level detection than for localization. MVTec and VisA also show
reduced contrast between newly learned and previously learned tasks under CL,
although some low-intensity regions remain. DAGM exhibits more moderate gains
and retains class-dependent variability, suggesting that continual adaptation
reduces but does not eliminate concept-specific interference.
Across datasets, pixel-level heatmaps are generally warmer and more stable than
image-level heatmaps. This is particularly evident for MPDD, MVTec, and VisA,
where CL produces high and relatively uniform localization scores across many
task transitions. This separates two effects that are conflated in a single
aggregate score: a model may preserve local defect evidence well while still
struggling to convert that evidence into robust image-level anomaly scores. The
heatmaps therefore explain why the main table can show strong pixel-level CL
performance even when image-level BWT remains more negative for some datasets.

\begin{figure*}[p]
  \centering
  \includegraphics[width=\textwidth]{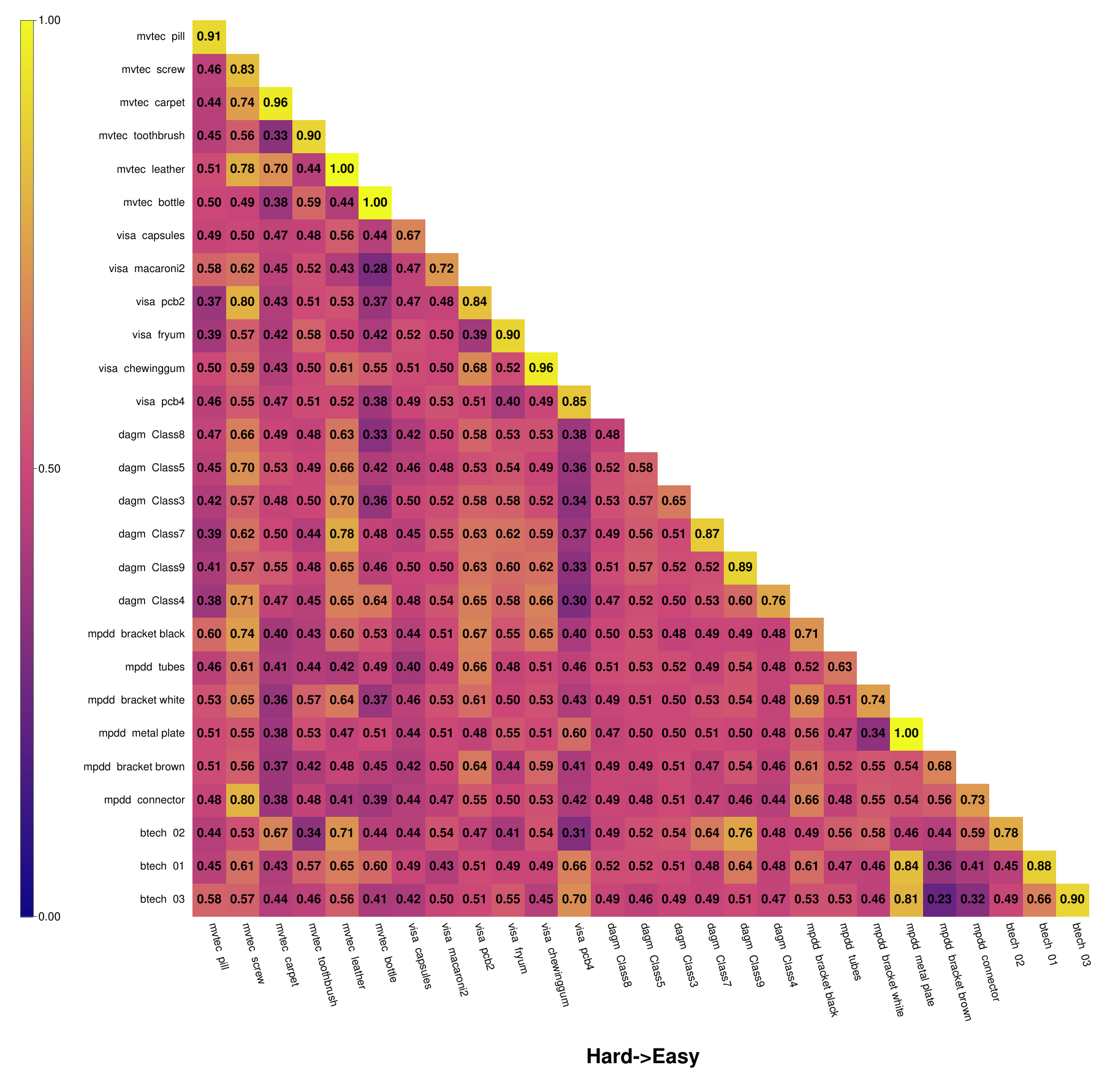}
  \caption{\textbf{MultiDataset}, ResNet-18, Hard-to-Easy: NCL, image-level concept transition matrix.}
  \label{fig:ordering-multidataset-r18-h2e-ncl-image}
\end{figure*}

\begin{figure*}[p]
  \centering
  \includegraphics[width=\textwidth]{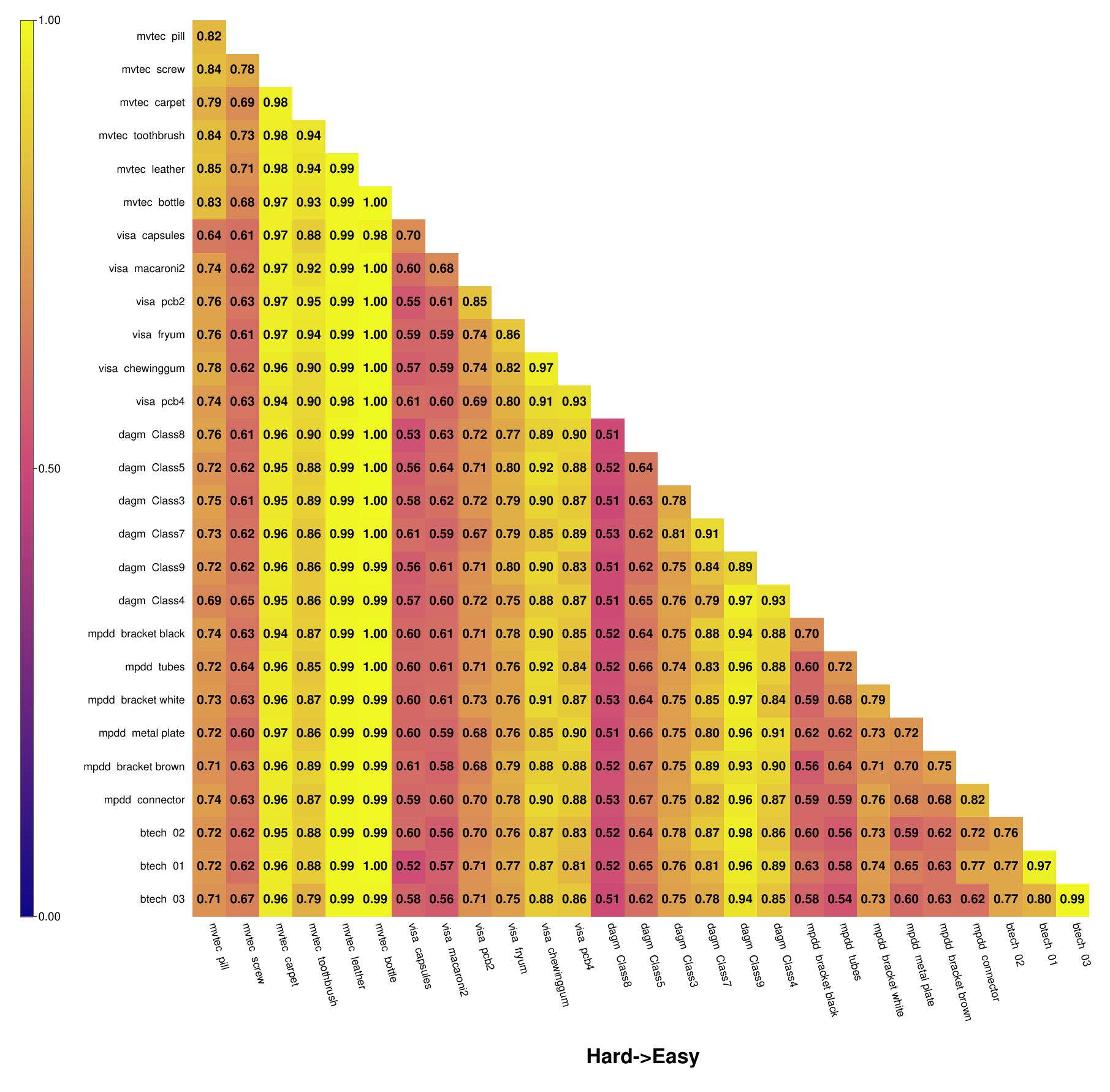}
  \caption{\textbf{MultiDataset}, ResNet-18, Hard-to-Easy: CL, image-level concept transition matrix.}
  \label{fig:ordering-multidataset-r18-h2e-cl-image}
\end{figure*}

\begin{figure*}[p]
  \centering
  \includegraphics[width=\textwidth]{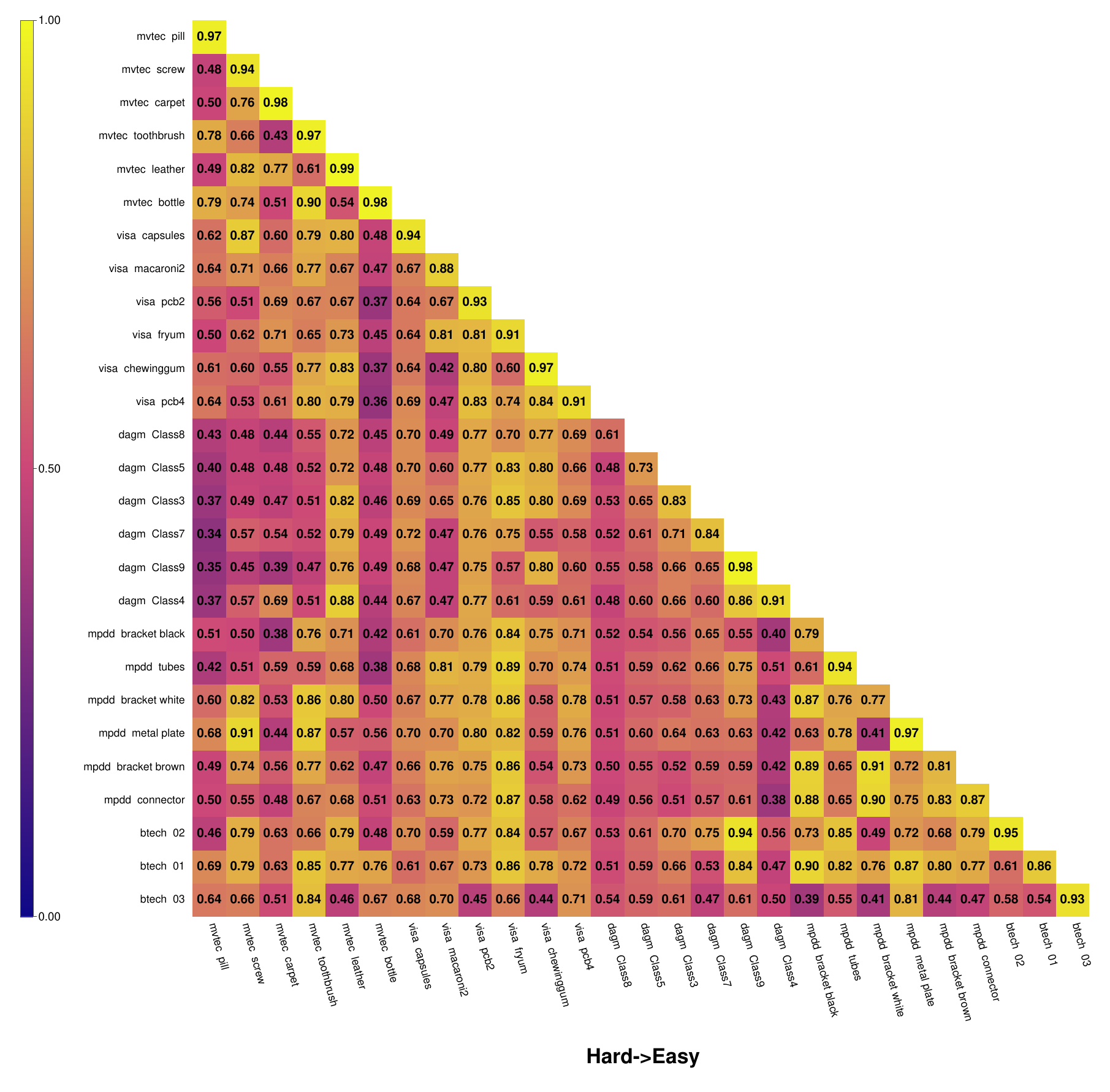}
  \caption{\textbf{MultiDataset}, ResNet-18, Hard-to-Easy: NCL, pixel-level concept transition matrix.}
  \label{fig:ordering-multidataset-r18-h2e-ncl-pixel}
\end{figure*}

\begin{figure*}[p]
  \centering
  \includegraphics[width=\textwidth]{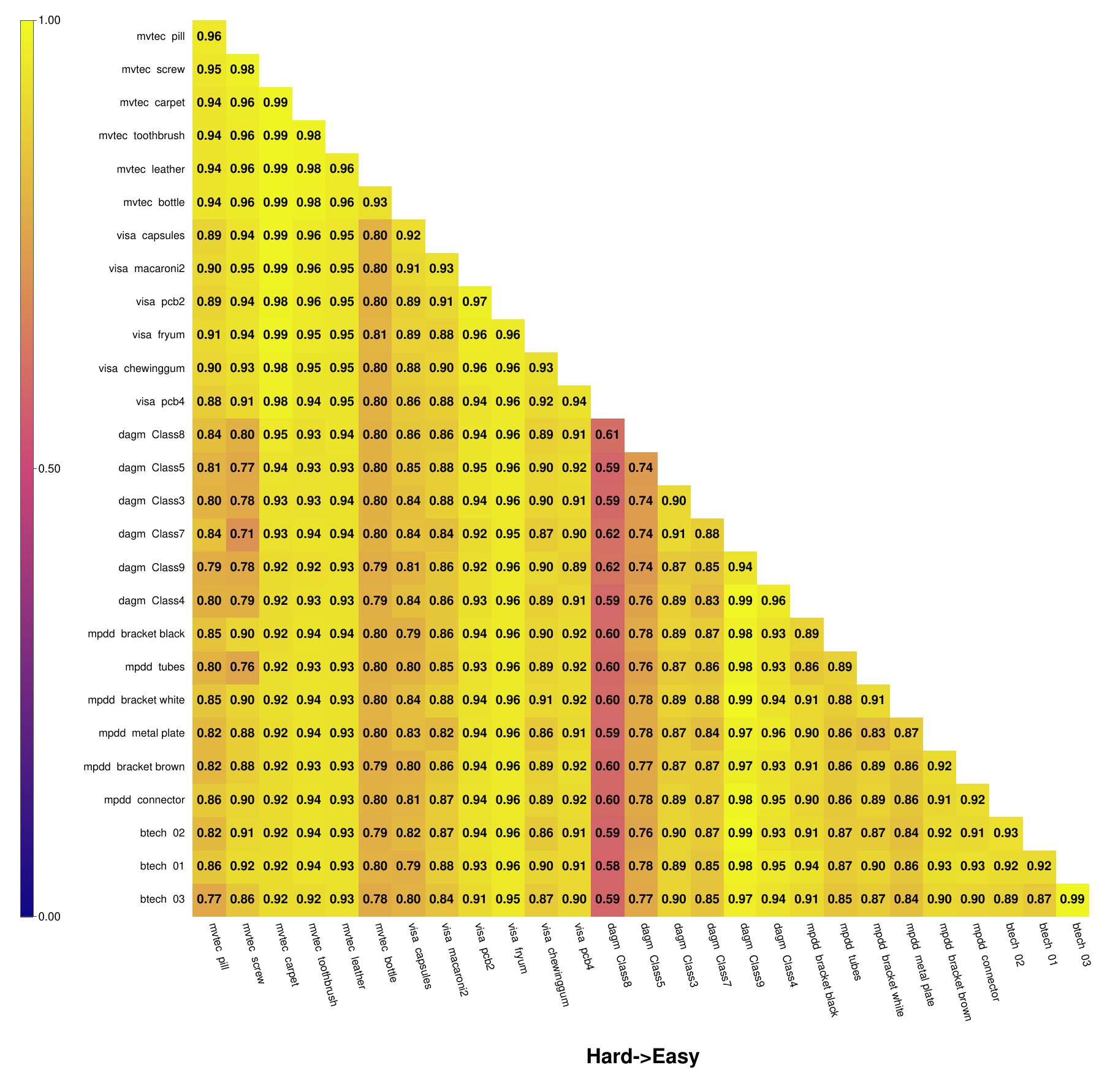}
  \caption{\textbf{MultiDataset}, ResNet-18, Hard-to-Easy: CL, pixel-level concept transition matrix.}
  \label{fig:ordering-multidataset-r18-h2e-cl-pixel}
\end{figure*}

\begin{figure*}[p]
  \centering
  \includegraphics[width=\textwidth]{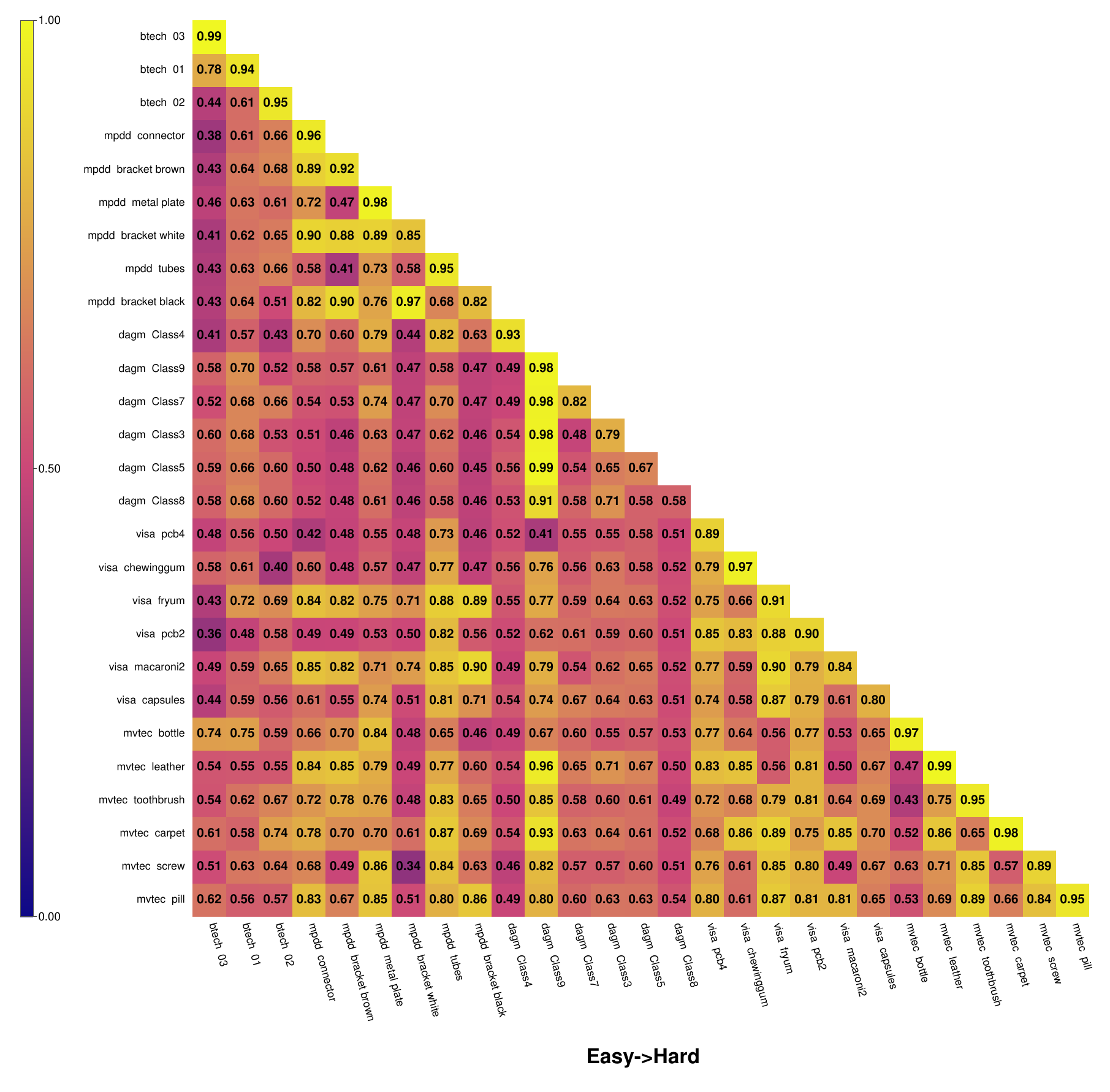}
  \caption{\textbf{MultiDataset}, ResNet-18, Easy-to-Hard: NCL, pixel-level concept transition matrix.}
  \label{fig:ordering-multidataset-r18-e2h-ncl-pixel}
\end{figure*}

\begin{figure*}[p]
  \centering
  \includegraphics[width=\textwidth]{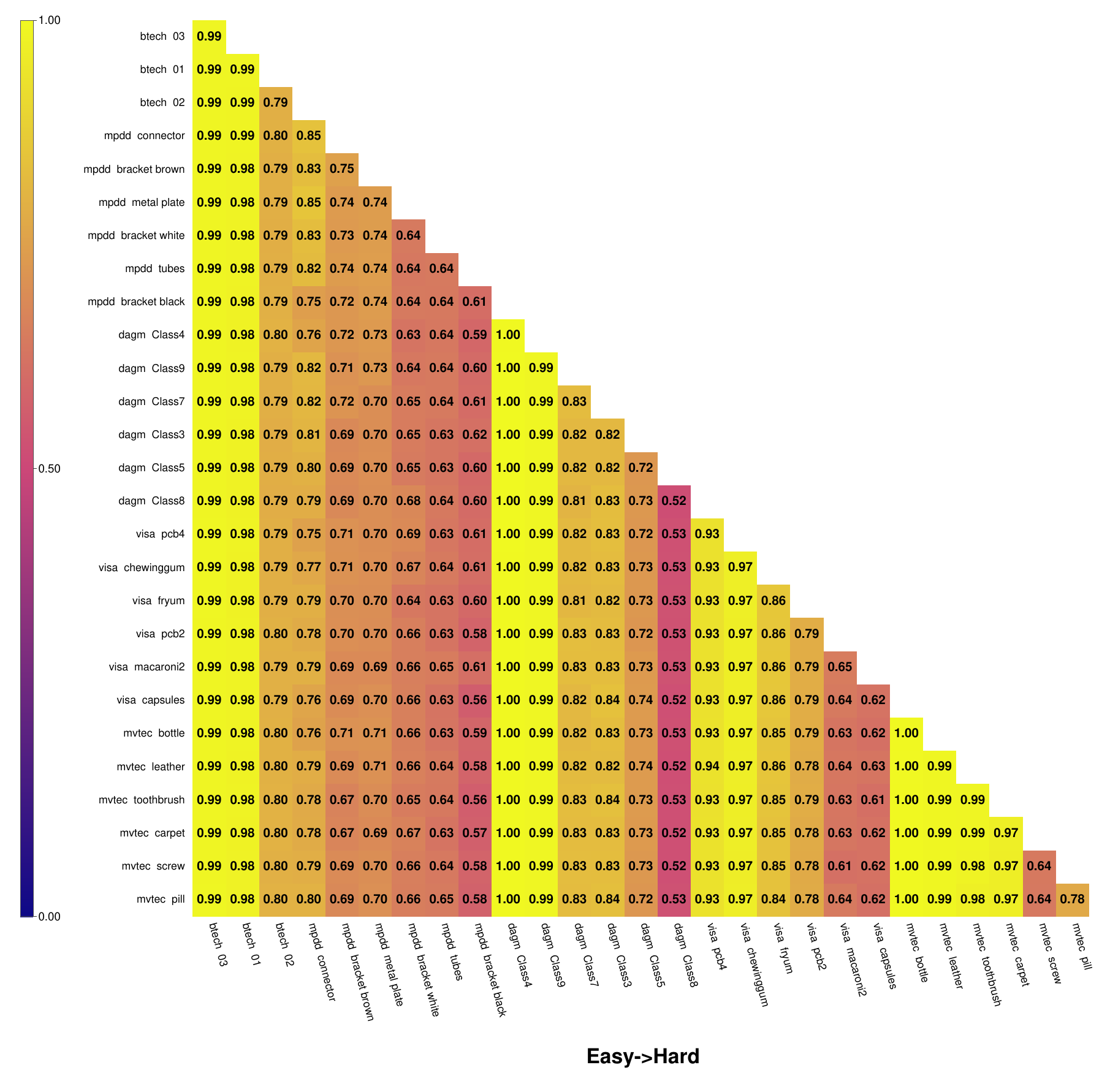}
  \caption{\textbf{MultiDataset}, ResNet-18, Easy-to-Hard: CL, image-level concept transition matrix.}
  \label{fig:ordering-multidataset-r18-e2h-cl-image}
\end{figure*}

\begin{figure*}[p]
  \centering
  \includegraphics[width=\textwidth]{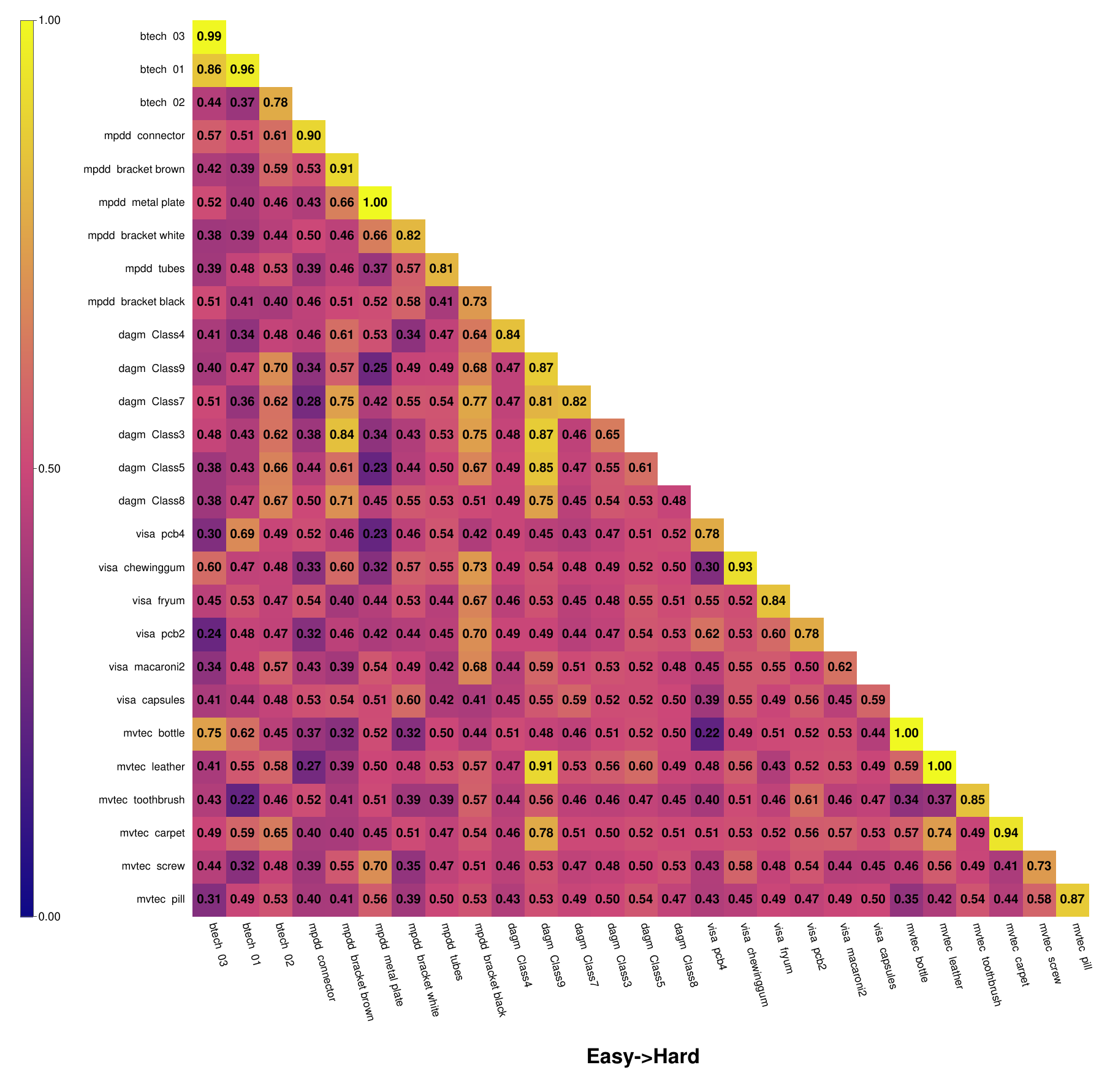}
  \caption{\textbf{MultiDataset}, ResNet-18, Easy-to-Hard: NCL, image-level concept transition matrix.}
  \label{fig:ordering-multidataset-r18-e2h-ncl-image}
\end{figure*}

\begin{figure*}[p]
  \centering
  \includegraphics[width=\textwidth]{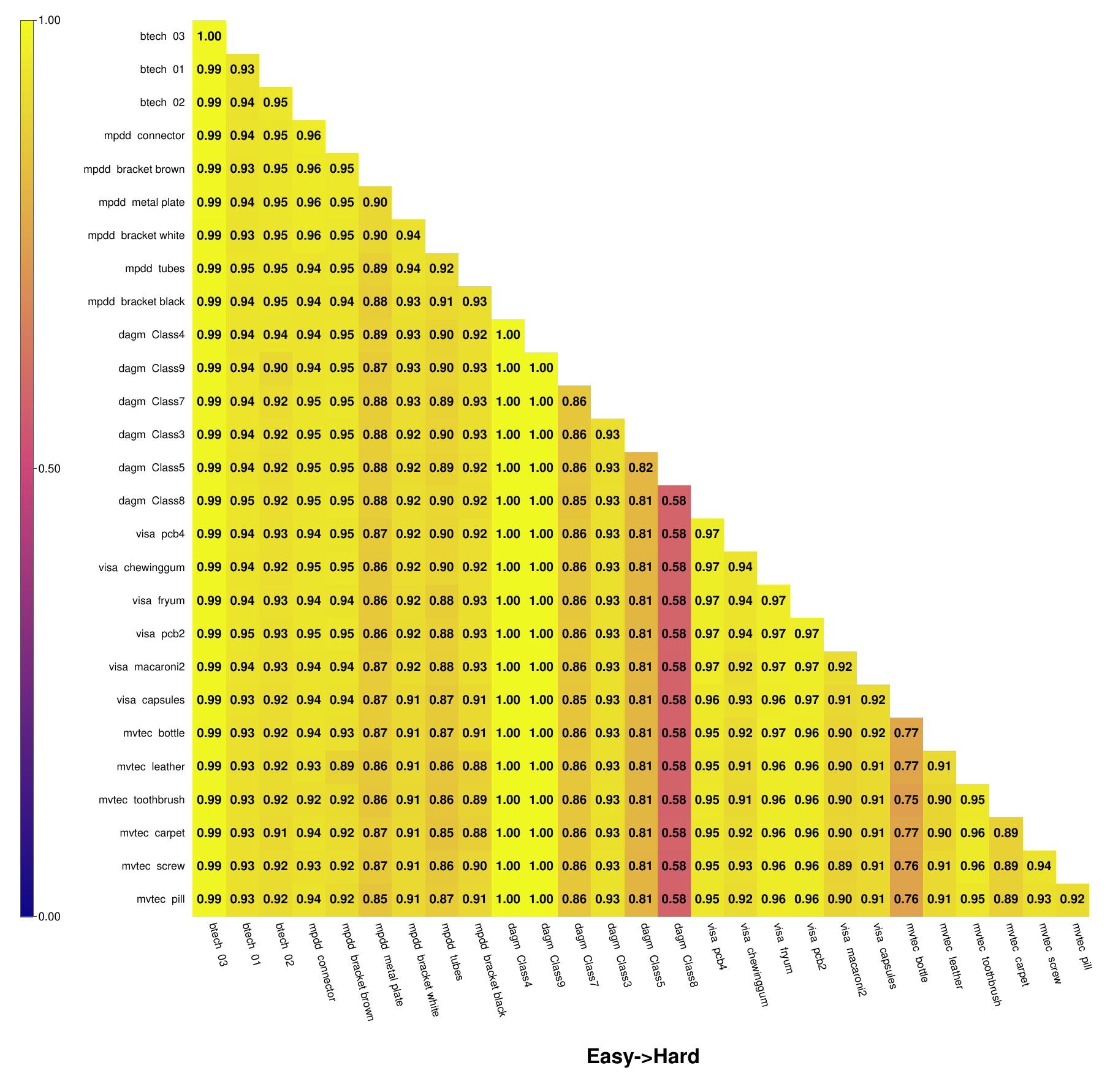}
  \caption{\textbf{MultiDataset}, ResNet-18, Easy-to-Hard: CL, pixel-level concept transition matrix.}
  \label{fig:ordering-multidataset-r18-e2h-cl-pixel}
\end{figure*}

\begin{figure*}[p]
  \centering
  \includegraphics[width=\textwidth]{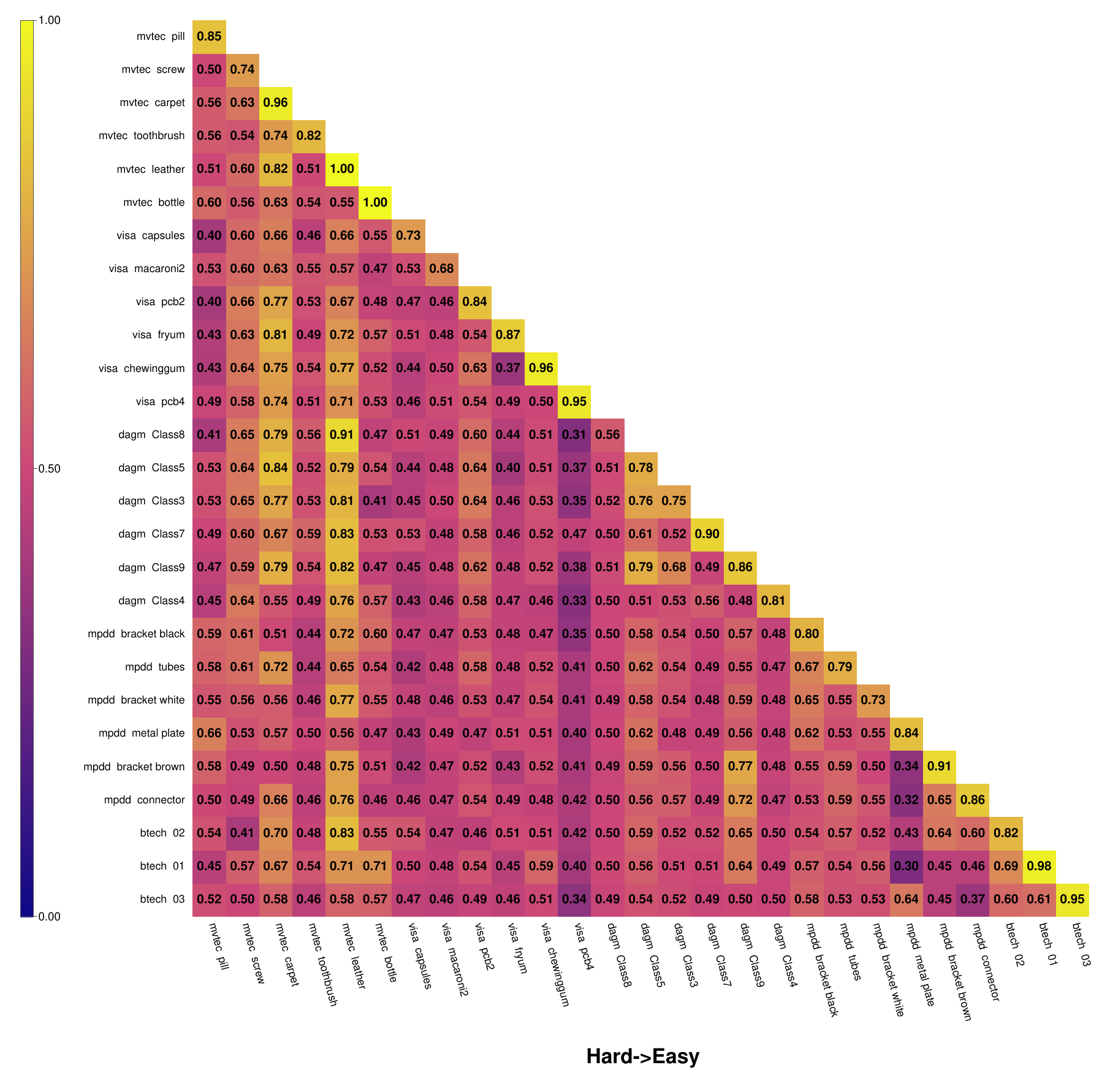}
  \caption{\textbf{MultiDataset}, Wide ResNet-50-2, Hard-to-Easy: NCL, image-level concept transition matrix.}
  \label{fig:ordering-multidataset-wrn50-h2e-ncl-image}
\end{figure*}

\begin{figure*}[p]
  \centering
  \includegraphics[width=\textwidth]{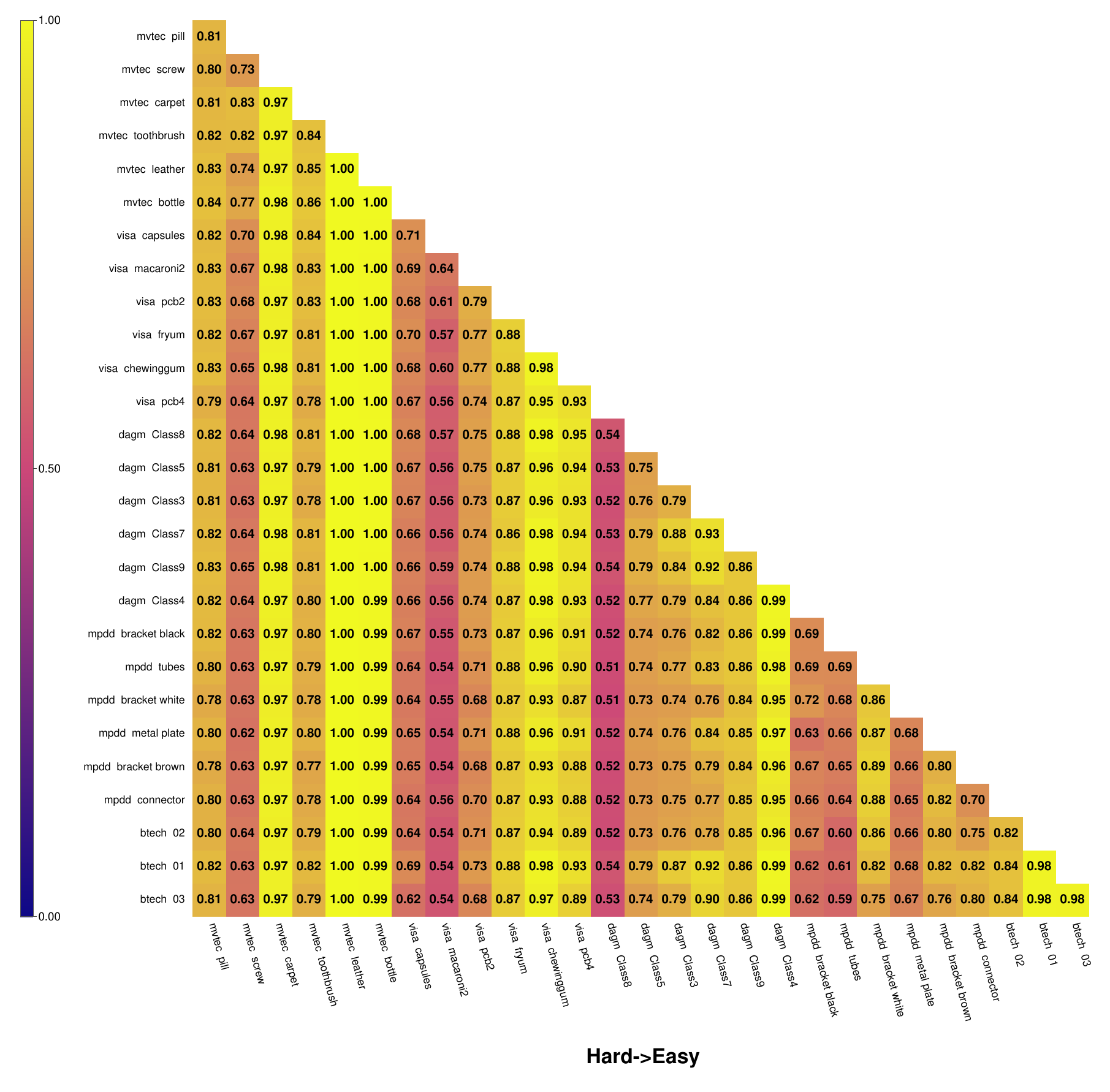}
  \caption{\textbf{MultiDataset}, Wide ResNet-50-2, Hard-to-Easy: CL, image-level concept transition matrix.}
  \label{fig:ordering-multidataset-wrn50-h2e-cl-image}
\end{figure*}

\begin{figure*}[p]
  \centering
  \includegraphics[width=\textwidth]{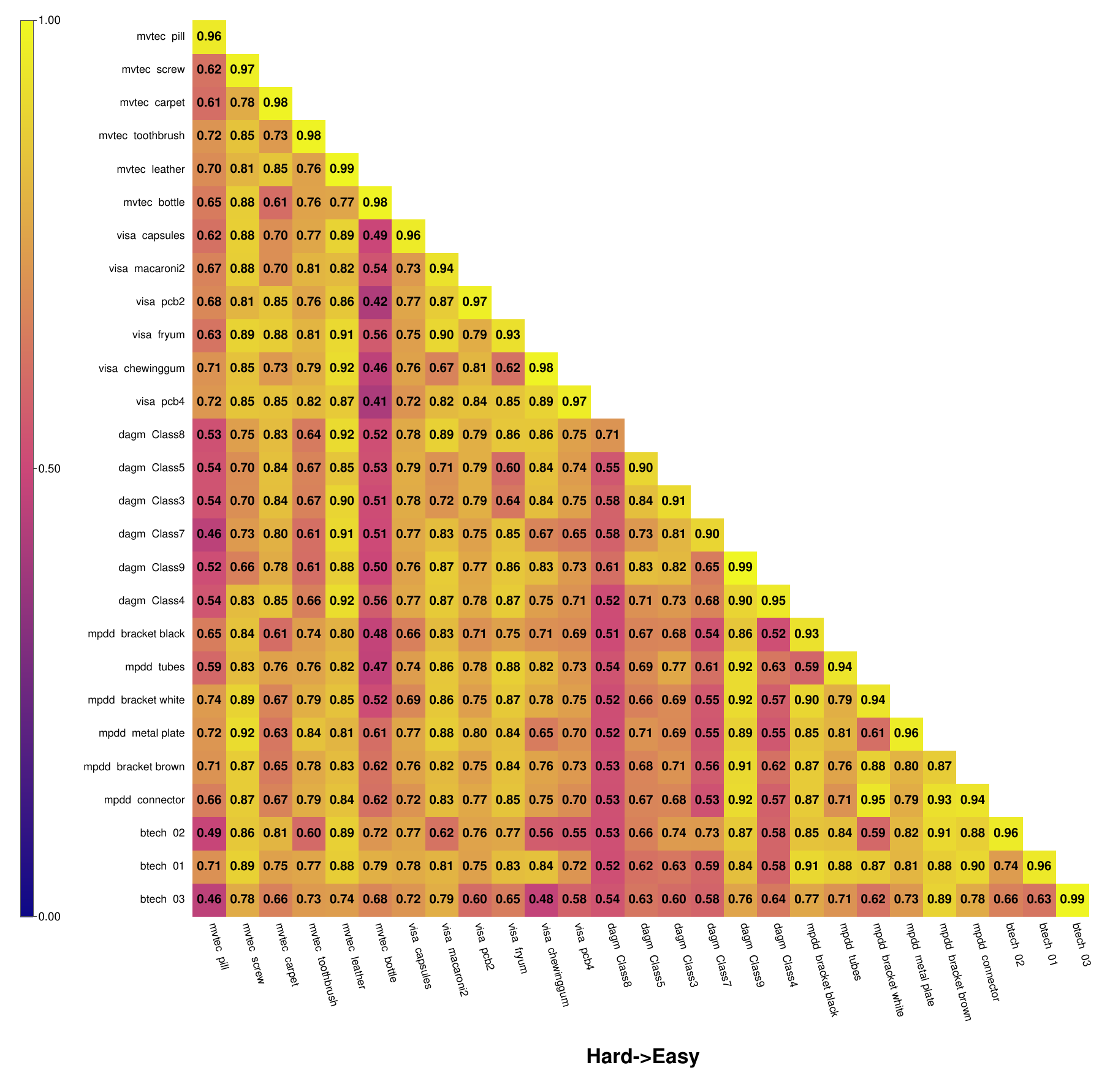}
  \caption{\textbf{MultiDataset}, Wide ResNet-50-2, Hard-to-Easy: NCL, pixel-level concept transition matrix.}
  \label{fig:ordering-multidataset-wrn50-h2e-ncl-pixel}
\end{figure*}

\begin{figure*}[p]
  \centering
  \includegraphics[width=\textwidth]{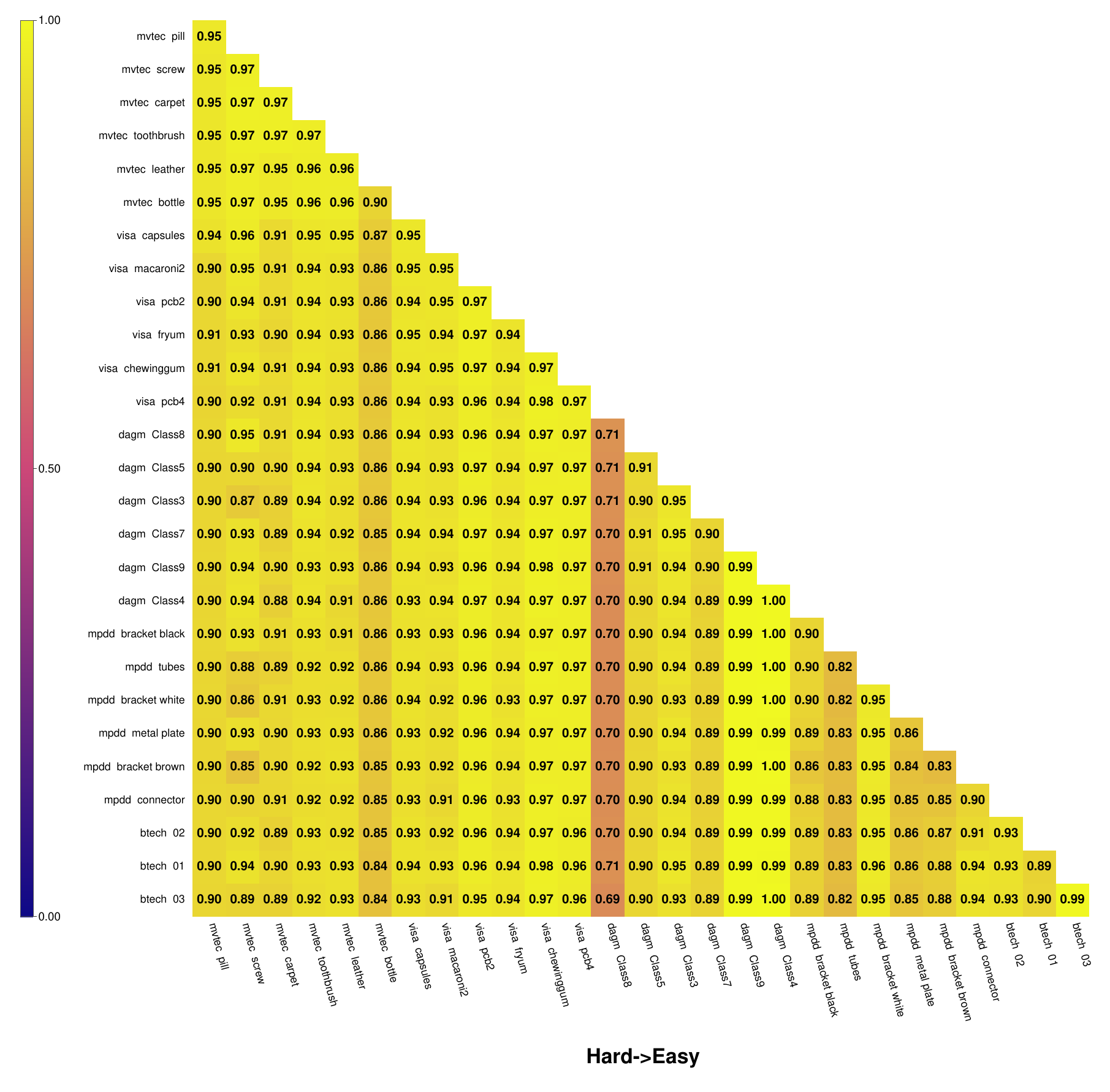}
  \caption{\textbf{MultiDataset}, Wide ResNet-50-2, Hard-to-Easy: CL, pixel-level concept transition matrix.}
  \label{fig:ordering-multidataset-wrn50-h2e-cl-pixel}
\end{figure*}

\begin{figure*}[p]
  \centering
  \includegraphics[width=\textwidth]{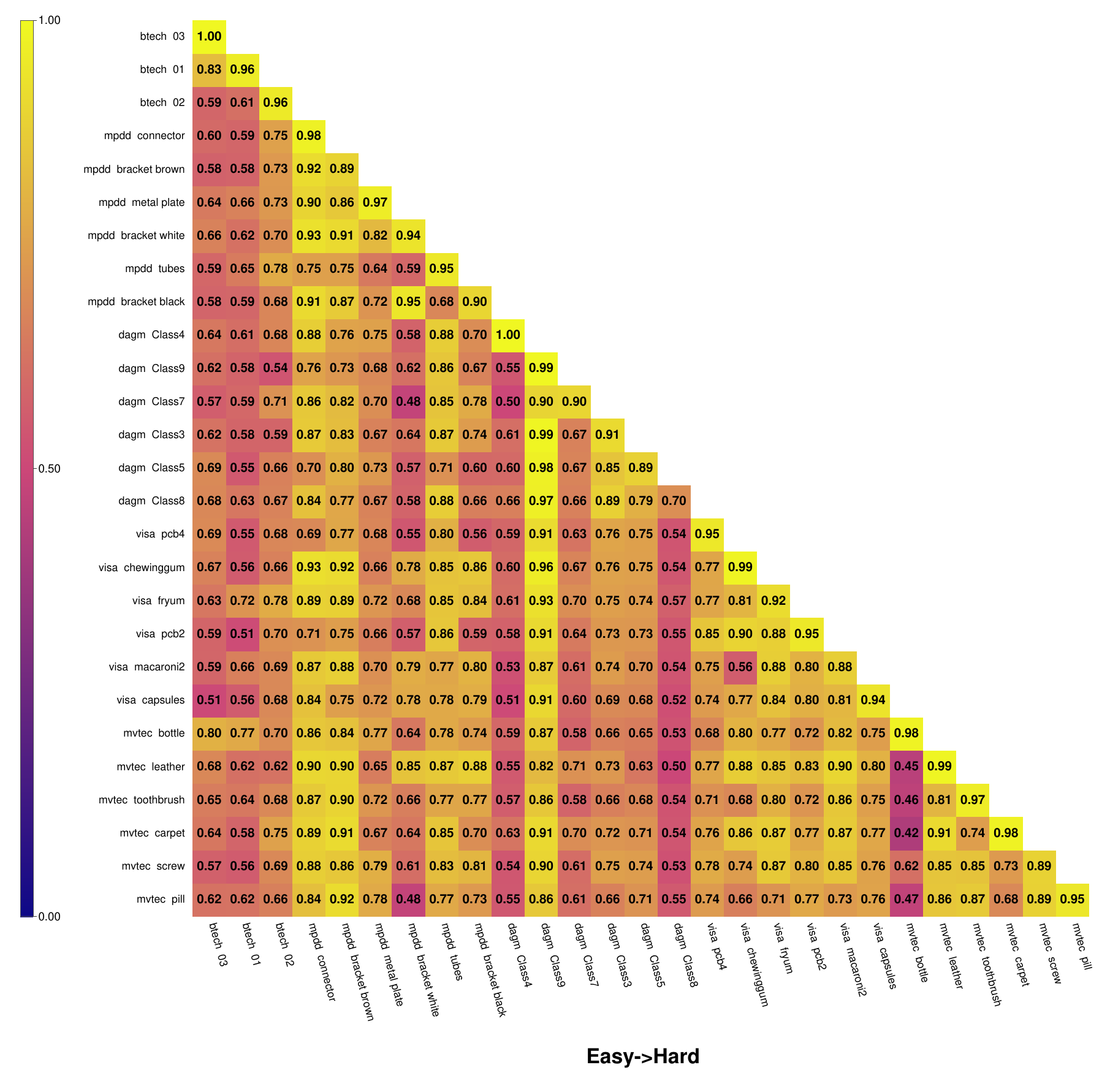}
  \caption{\textbf{MultiDataset}, Wide ResNet-50-2, Easy-to-Hard: NCL, pixel-level concept transition matrix.}
  \label{fig:ordering-multidataset-wrn50-e2h-ncl-pixel}
\end{figure*}

\begin{figure*}[p]
  \centering
  \includegraphics[width=\textwidth]{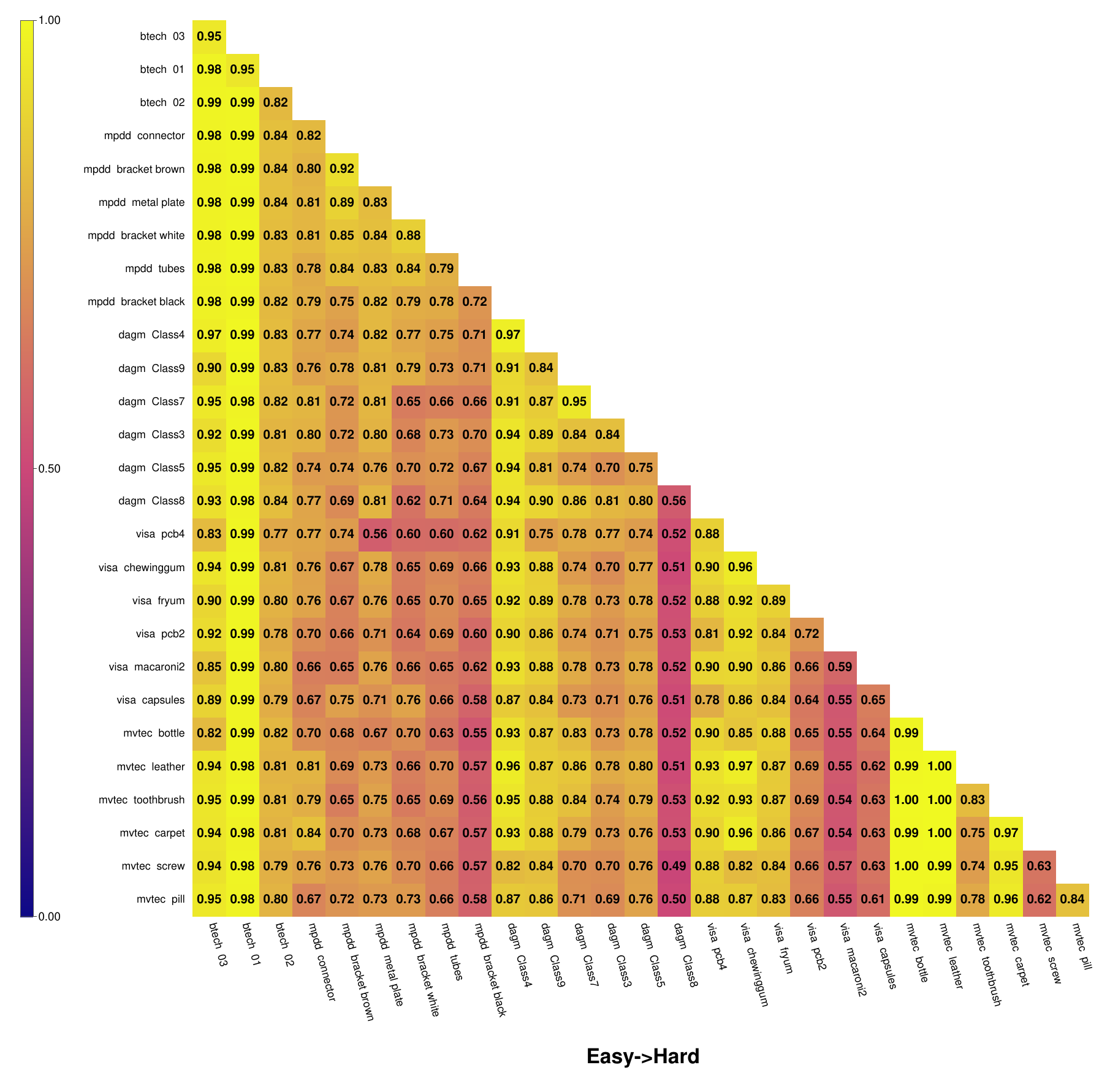}
  \caption{\textbf{MultiDataset}, Wide ResNet-50-2, Easy-to-Hard: CL, image-level concept transition matrix.}
  \label{fig:ordering-multidataset-wrn50-e2h-cl-image}
\end{figure*}

\begin{figure*}[p]
  \centering
  \includegraphics[width=\textwidth]{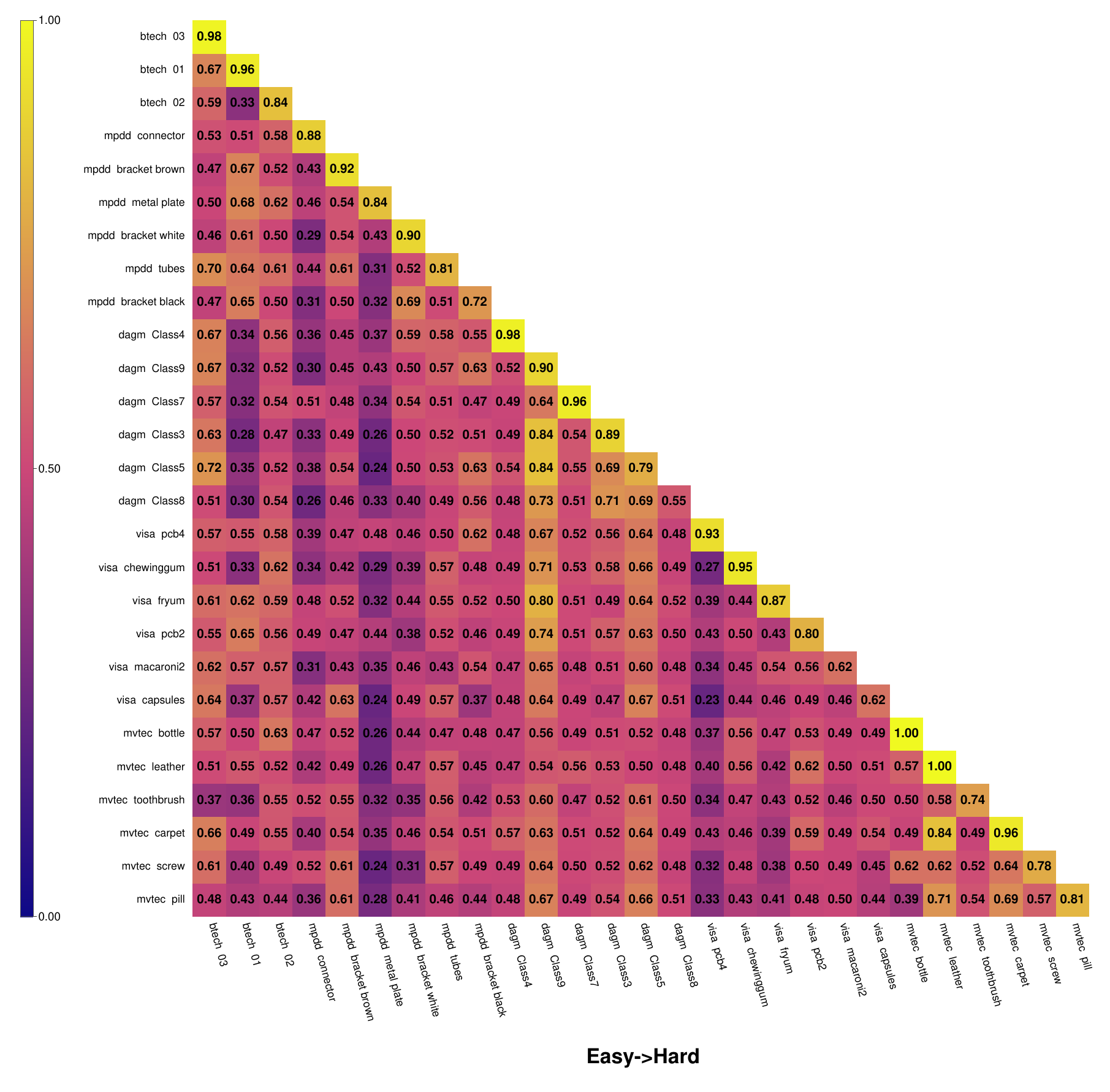}
  \caption{\textbf{MultiDataset}, Wide ResNet-50-2, Easy-to-Hard: NCL, image-level concept transition matrix.}
  \label{fig:ordering-multidataset-wrn50-e2h-ncl-image}
\end{figure*}

\begin{figure*}[p]
  \centering
  \includegraphics[width=\textwidth]{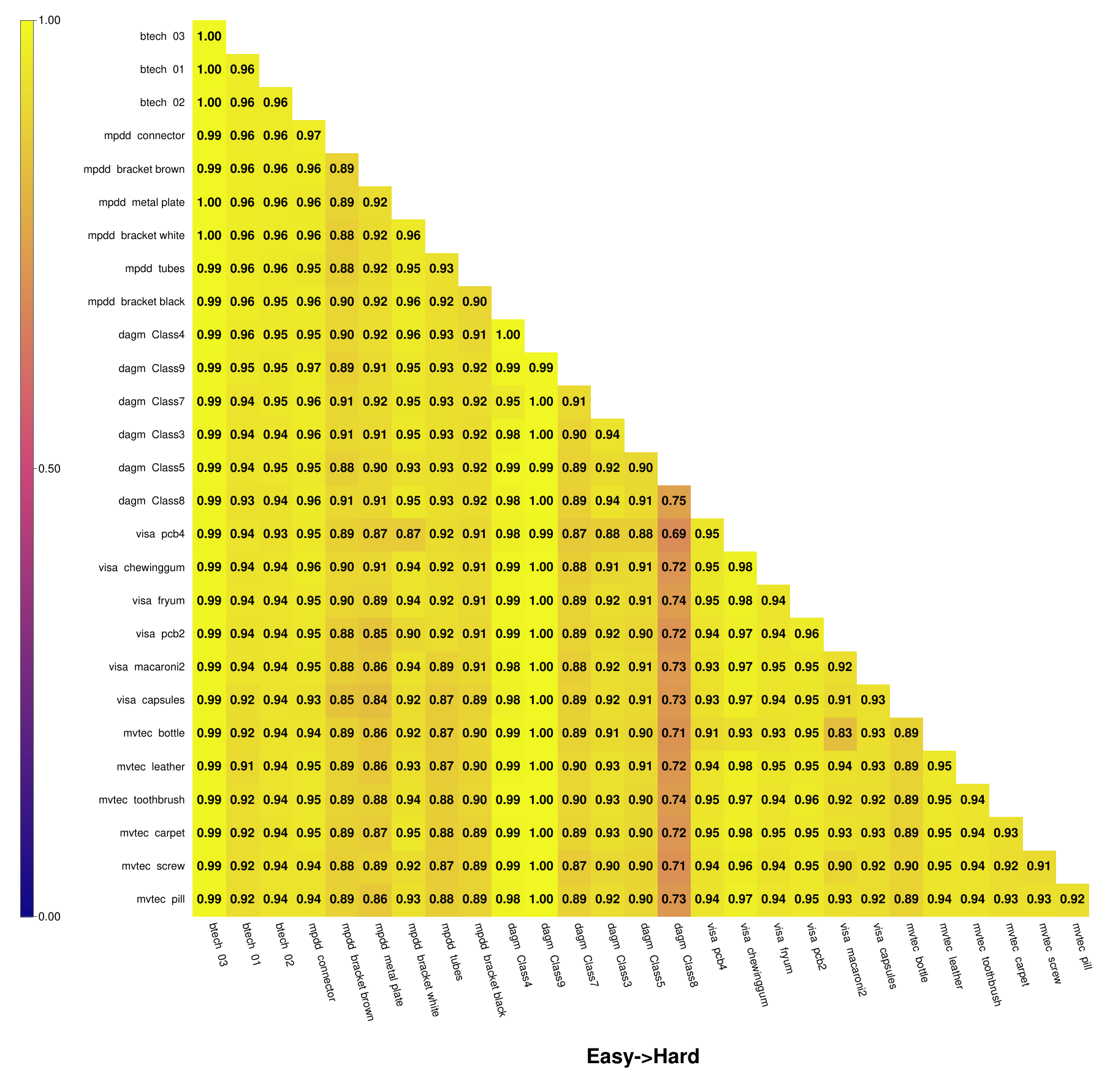}
  \caption{\textbf{MultiDataset}, Wide ResNet-50-2, Easy-to-Hard: CL, pixel-level concept transition matrix.}
  \label{fig:ordering-multidataset-wrn50-e2h-cl-pixel}
\end{figure*}

\paragraph{Multi-dataset streams.}
Figures~\ref{fig:ordering-multidataset-r18-h2e-ncl-image}--\ref{fig:ordering-multidataset-r18-e2h-cl-pixel}
and~\ref{fig:ordering-multidataset-wrn50-h2e-ncl-image}--\ref{fig:ordering-multidataset-wrn50-e2h-cl-pixel}
extend the transition-matrix analysis to the 27-task multi-dataset stream for
ResNet-18 and Wide ResNet-50-2, respectively. These heatmaps are averaged across
the three strongest methods selected from the single-dataset study, CFA, PaDiM,
and PaSTe, so they summarize behavior across multiple detector families rather
than a single model. Compared with the within-dataset matrices, the
multi-dataset heatmaps expose broader domain-shift effects: low-intensity bands
are less tied to a single product category and more often appear around
transitions between source datasets. This pattern is visible for both backbones,
which indicates that the multi-dataset scenario introduces interference beyond
ordinary within-dataset task ordering rather than an artifact of a particular
feature extractor.
The NCL--CL contrast remains clear in this harder setting. For both
Hard-to-Easy and Easy-to-Hard streams,
Figures~\ref{fig:ordering-multidataset-r18-h2e-ncl-image} and
\ref{fig:ordering-multidataset-r18-e2h-ncl-image} show that image-level NCL
performance is fragile after cross-dataset transitions, whereas the corresponding
CL heatmaps in Figures~\ref{fig:ordering-multidataset-r18-h2e-cl-image} and
\ref{fig:ordering-multidataset-r18-e2h-cl-image} recover more of the
off-diagonal mass. This supports the conclusion that continual adaptation is
especially important when new tasks change not only the inspected component but
also the dataset domain. Pixel-level results show the same qualitative pattern:
CL produces warmer and more coherent retained-task regions than NCL in
Figures~\ref{fig:ordering-multidataset-r18-h2e-cl-pixel},
\ref{fig:ordering-multidataset-r18-e2h-cl-pixel},
\ref{fig:ordering-multidataset-wrn50-h2e-cl-pixel}, and
\ref{fig:ordering-multidataset-wrn50-e2h-cl-pixel}, suggesting that localization
is more robust to cross-dataset heterogeneity once replay or method-specific
continual mechanisms are used. The Wide ResNet-50-2 heatmaps reinforce this
interpretation: despite changes in representation capacity, the same qualitative
structure persists---NCL is most brittle in off-diagonal retained-task regions,
whereas CL yields smoother lower-triangular blocks and better retention across
dataset boundaries. Thus, the multi-dataset findings are not confined to the
ResNet-18 backbone; they reflect a consistent benchmark-level stress pattern.

\clearpage

\clearpage

\clearpage

\clearpage

\clearpage

\clearpage

\clearpage

\clearpage


\clearpage

\clearpage

\clearpage

\clearpage

\clearpage

\clearpage

\clearpage

\clearpage

\end{document}